%% file: RO_HNN.tex
\documentclass{article}

\usepackage{microtype}
\usepackage{graphicx}
\usepackage{subcaption}
\usepackage{booktabs} 
\usepackage{multicol}
\usepackage{amssymb}
\usepackage{bm}
\usepackage{amsmath}
\usepackage{amsthm}
\usepackage{url}
\usepackage{graphicx}
\usepackage{xcolor}
\usepackage{import}
\usepackage{mathtools}
\usepackage{macros_kathi}
\usepackage{svg}
\usepackage{soul}
\usepackage{makecell}
\usepackage{multirow}
\usepackage{array}
\usepackage{colortbl}
\usepackage{siunitx}
\usepackage{adjustbox}
\usepackage{subcaption}
\usepackage{adjustbox}
\usepackage{booktabs}
\usepackage{wrapfig}
\usepackage[nolist]{acronym} 
\usepackage{hyperref}

\usepackage[accepted]{icml2026}

\usepackage{amsmath}
\usepackage{amssymb}
\usepackage{mathtools}
\usepackage{amsthm}

\usepackage[capitalize,noabbrev]{cleveref}

\theoremstyle{plain}
\newtheorem{theorem}{Theorem}[section]
\newtheorem{proposition}[theorem]{Proposition}

\theoremstyle{definition}

\theoremstyle{remark}

\icmltitlerunning{Learning Hamiltonian Dynamics at Scale}

\begin{document}
\input{acronyms}
\twocolumn[
  \icmltitle{Learning Hamiltonian Dynamics at Scale: A Differential-Geometric Approach}



  \icmlsetsymbol{equal}{*}

  \begin{icmlauthorlist}
    \icmlauthor{Katharina Friedl}{yyy}
    \icmlauthor{No\'emie Jaquier}{yyy}
    \icmlauthor{Alyx Liao}{yyy,comp}
    \icmlauthor{Danica Kragic}{yyy}
  \end{icmlauthorlist}

  \icmlaffiliation{yyy}{Department of Robotics, Perception, and Learning, KTH Royal Institute of Technology, Sweden}
  \icmlaffiliation{comp}{Ecole Normale Sup\'erieure Paris, France}

  \icmlcorrespondingauthor{Katharina Friedl}{kfriedl@kth.se}

  \icmlkeywords{Hamiltonian dynamics, symplectic geometry, Riemannian geometry, autoencoder, structure-preserving ROM, model order reduction}

  \vskip 0.3in
]



\printAffiliationsAndNotice{}  

\begin{abstract}
\input{texfiles/abstract.tex}
\end{abstract}

\input{texfiles/01introduction.tex}
\input{texfiles/02preliminaries.tex}

\input{texfiles/03method.tex}

\input{texfiles/04results.tex}

\input{texfiles/05conclusion.tex}

\clearpage

\section*{Impact Statement}
This paper presents work whose goal is to advance the field of Machine Learning. There are many potential societal consequences of our work, none which we feel must be specifically highlighted here.

\section*{Acknowledgements}
This work was supported by ERC AdV grant BIRD 884807, Swedish Research Council, and the Knut and Alice Wallenberg Foundation, including the Wallenberg AI, Autonomous Systems and Software Program (WASP).
The computations were enabled by the Berzelius resource provided by the Knut and Alice Wallenberg Foundation at the National Supercomputer Centre.

\bibliography{references}
\bibliographystyle{icml2026}

\newpage
\appendix
\onecolumn
\input{texfiles/technical_appendix.tex}
\input{texfiles/experimental_descriptive_appendix.tex}

\input{texfiles/experiments_in_appendix.tex}

\end{document}

%% file: acronyms.tex
\newacro{ae}[AE]{Autoencoder}
\newacro{rom}[ROM]{reduced-order model}
\newacro{fom}[FOM]{full-order model}
\newacro{mor}[MOR]{model order reduction}
\newacro{lnn}[LNN]{Lagrangian neural network}
\newacro{hnn}[HNN]{Hamiltonian neural network}
\newacro{dof}[DoF]{degrees-of-freedom}
\newacro{ivp}[IVP]{initial value problem}
\newacro{dnn}[DNN]{Deep Neural Network}
\newacro{spd}[SPD]{symmetric positive-definite}
\newacro{mlp}[MLP]{multilayer perceptron}
\newacro{fc}[FC]{fully-connected}
\newacro{gyrospd}[GyroSpd$_{\ty{++}}$]{gyrospace hyperplane-based}
\newacro{gyroai}[GyroAI]{gyrocalculus-based}
\newacro{rolnn}[RO-LNN]{reduced-order LNN}
\newacro{rohnn}[RO-HNN]{reduced-order Hamiltonian neural network}
\newacro{pde}[PDE]{partial differential equation}
\newacro{hnko}[HNKO]{Hamiltonian neural Koopman operator}

%% file: texfiles/abstract.tex
\label{abstract}

Embedding physical intuition into network architectures allows the learning of dynamics that enforce fundamental properties, such as energy conservation laws, thereby leading to physically-plausible predictions.
Yet, scaling these models to high-dimensional dynamical systems remains a significant challenge.  
This paper introduces Reduced-order Hamiltonian Neural Network (RO-HNN), a novel physics-inspired neural network that combines the conservation laws of Hamiltonian mechanics with the scalability of model order reduction. 
RO-HNN is built on two core components: a novel geometrically-constrained symplectic autoencoder that learns a low-dimensional, structure-preserving symplectic submanifold, and a geometric Hamiltonian neural network that models the dynamics on the submanifold. 
Our experiments demonstrate that RO-HNN provides physically-consistent, stable, and generalizable predictions of complex high-dimensional dynamics, thereby effectively extending the scope of Hamiltonian neural networks to high-dimensional physical systems.
 

%% file: texfiles/01introduction.tex
\section{Introduction}
\label{sec:introduction}

Learning the unknown governing equations of dynamical systems is of fundamental importance to model physical processes. In this context, generic neural models lack built-in physical intuition, thus resulting in limited explainability and poor generalization beyond the data support.
Embedding fundamental physical properties, such as conservation laws and boundary conditions, into neural networks has been shown to drastically improve their performance.
Various models incorporate physical intuition as soft constraints via penalty terms in the loss function~\citep{Saqlain23:PINNs}. This often leads to suboptimal enforcement of physical properties and to stiff optimization~\citep{wang2021understanding}, motivating the embedding of domain-specific priors as hard constraints in specialized neural architectures. This allowed recent methods to learn dynamics that preserve energy~\citep{Greydanus2019HNN,CranmerGreydanus2020LNN,Lutter2023DeLaN}, conserve mass and momentum~\citep{jnini2025riemann}, and strictly enforce general conservation laws~\citep{liu2024harnessing}, thereby improving performances, generalization, and stability while yielding physically-consistent predictions.

Hamiltonian mechanics, introduced by~\citet{Hamilton34} as a reformulation of Lagrangian mechanics, describe the evolution of a broad range of dynamical systems in robotics~\citep{Duong21Hamiltonian}, fluid dynamics~\citep{Salmon88:FluidMech}, quantum mechanics~\citep{Schrodinger26}, and biology~\citep{Duarte98:LotkaVolterra}, 
among others.
Hamiltonian systems evolve on a phase space with symplectic structure, naturally enforcing energy conservation~\citep{AbrahamMarsden87:Foundations}. 
Compared to Lagrangian mechanics, Hamiltonian mechanics provide a first-order formulation of dynamics that describes a broader range of physical systems. 
\acp{hnn} are gray-box models that embed the Hamiltonian structure as hard constraints in specialized deep learning architectures. 
\acp{hnn} either directly learn the Hamiltonian function, ensuring conservation laws by construction~\citep{Greydanus2019HNN, Lutter2023DeLaN}, or learn symplectomorphisms that preserve the invariants of interest via symplectic flows~\citep{Jin2020:SympNets}. \acp{hnn} were enhanced by including dissipation~\citep{zhong2020dissipative} and contact~\citep{Zhong2021Contact} models, and utilized for model-based control~\citep{Duong21Hamiltonian,zhong2020symplectic}. While most \acp{hnn} consider Hamiltonians characterized by a canonical symplectic form --- exhibited at least locally for all Hamiltonian systems --- few works proposed architectures handling non-canonical forms~\citep{Chen2021:LearnSympl} and more general Poisson systems~\citep{jin2023poissonNN, sipka2023poissonNN}.
Although \acp{hnn} yielded drastic performance improvements over generic black-box models, their application remains limited to low-dimensional systems with $2$-$5$ dimensions. 

Learning the dynamics of high-dimensional physical systems, such as robots, continua, or fluids, is arguably a difficult problem due to the increasing complexity and nonlinearity of their governing equations. 
Several approaches combine data-driven sparse identification of nonlinear dynamics (SINDy) and dimensionality reduction to discover high-dimensional governing equations~\citep{brunton16sindy,Champion19:SindyAE}. However, they disregard the \emph{apriori}-known structures of physical systems.
Another body of works~\citep{SanchezGonzalez2019Hamiltonian, thangamuthu22, rahma2025rapidghnn} represents the state of a physical system with particles identified with nodes in a graph and learn a Hamiltonian function via a graph neural network, thus leveraging message-passing among nodes to approximate complex dynamics. While effective for homogeneous systems, the generalization of these approaches to high-dimensional systems is limited by the assumption that all particles share similar dynamics and interaction patterns. 
In contrast,~\citet{SharmaKramer2024:LOpInf,friedl2025reduced} took inspiration from \ac{mor} to learn high-dimensional Lagrangian dynamics. 
\ac{mor} addresses the complexity of nonlinear high-dimensional governing equations, so-called \ac{fom}, by finding a \ac{rom}, i.e., a computationally-cheaper yet accurate low-dimensional surrogate model~\citep{Schilders2008MOR}. While \ac{mor} techniques are typically intrusive, i.e., they assume entirely-known \ac{fom} dynamics,~\citet{Sharma2024:LOpInf-NN} presented a novel non-intrusive \ac{mor}-based approach that learns the parameters of a high-dimensional Lagrangian system in a linear structure-preserving subspace. In a similar line,~\citet{friedl2025reduced} adopted a Riemannian perspective on the problem and introduced a physics-inspired neural architecture that jointly learns a non-linear embedded submanifold via a biorthogonal \ac{ae} and its associated low-dimensional conservative dynamics via a geometric \ac{lnn}. 
A different line of works leverage Koopman operator theory to model nonlinear, e.g., Hamiltonian, dynamics via a learned surrogate linear dynamic model embedded in a higher-dimensional latent space~\citep{lusch2018deep,wrongkoopman2024}.

\textbf{This paper} proposes a novel physics-inspired geometric deep neural network to learn the dynamics of high-dimensional Hamiltonian systems. In contrast to previous works that learn dynamics from high-dimensional observations such as images~\citep{Greydanus2019HNN,Chen2021:LearnSympl,Botev24:WhichPriorMatter}, we consider systems with \emph{intrinsically high-dimensional} phase spaces.  Taking inspiration from~\citep{SharmaKramer2024:LOpInf,friedl2025reduced}, we build on recent advances in Hamiltonian \ac{mor}~\citep{Peng2015:SympMOR,Buchfink2024MOR} and adopt a differential geometric perspective to embed the high-dimensional Hamiltonian structure as hard constraints in our architecture. \textbf{Our first contribution} is a geometrically-constrained symplectic \ac{ae} that learns a low-dimensional symplectic submanifold from trajectories of a high-dimensional Hamiltonian system. Unlike soft-constrained symplectic  networks~\citep{Buchfink23:SymplecticMOR}, our \ac{ae} guarantees the preservation of the symplectic structure of the \ac{fom}, including its conservation laws and stability properties~\citep{Lepri2024MOR}, with increased expressivity compared to linear and quadratic symplectic projections~\citep{Bendokat2022:SpStMOR,Sharma2023symplMOR}.
\textbf{Our second contribution} is a geometric \ac{hnn} that models conservative and dissipative Hamiltonian dynamics while accounting for the Riemannian geometry of its parameters, and resorts to symplectic integration~\citep{Tao2016:SymplInt} for accurate long-term dynamics simulation.
\textbf{Our third contribution} is a \ac{rohnn} that jointly learns a low-dimensional symplectic submanifold with a geometrically-constrained symplectic \ac{ae} and the dynamics parameters of the associated Hamiltonian function with a geometric \ac{hnn}. We validate our approach on three high-dimensional Hamiltonian systems: a pendulum, a thin cloth, and a particle vortex. Our experiments demonstrate that, due to its embedded geometries, \ac{rohnn} predicts accurate, stable, and physically-consistent trajectories, outperforming traditional \acp{hnn} and state-of-the-art reduction approaches. Source code is available at \url{https://github.com/katfriedl/reduced_hamiltonians}.

%% file: texfiles/02preliminaries.tex
\section{Background}
\label{sec2:preliminaries}
We briefly review Hamiltonian dynamics, Hamiltonian \ac{mor}, and related neural networks. App.~\ref{appendix:sec:geometry} provides preliminaries on Riemannian and symplectic geometry, and a summary of the geometric terminology and notations.

\subsection{Hamiltonian Dynamics on Symplectic Manifolds}
\label{sec2:sub1:hamiltonian_systems}
A symplectic manifold $(\mathcalm, \omega)$ is a $2n$-dimensional smooth manifold $\mathcalm$ equipped with a symplectic form $\omega$, i.e., a closed ($d\omega=0$), non-degenerate, differential $2$-form represented by a skew-symmetric matrix $\bm\omega$ in coordinates. We slightly abuse notation, equivalently denoting symplectic manifolds as $(\mathcalm, \bm\omega)$.
A Hamiltonian system $(\mathcalm, \bm\omega, \hamiltonian)$ is a dynamical system evolving on a symplectic manifold $(\mathcalm, \bm \omega)$ according to a smooth Hamiltonian function ${\hamiltonian :\mathcalm\to \mathbb R}$. The Hamiltonian vector field ${\bm X_{\hamiltonian} = \bm\omega^{-1}d\hamiltonian}$ is uniquely defined and preserves $\hamiltonian$. 
Trajectories ${\bm \gamma:\mathcal I \to \mathcalm}$ of the system over a time-interval $\mathcal I = \left[t_0, t_{\text{f}}\right]$ are solutions of the \ac{ivp}
\begin{equation}
    \tfrac{d}{dt}\bm\gamma\vert_t = \bm X_{\hamiltonian}\vert_{\bm \gamma(t)} \in T_{\bm \gamma(t)}\mathcalm, \: \text{with}\: \bm \gamma(t_0) = \bm \gamma_0 \in \mathcalm.
    \label{eq:sec2:hamiltonian_IVP}
\end{equation}

A diffeomorphism $f:(\mathcalm, \bm \omega)\to(\mathcal N, \bm \eta)$ between symplectic manifolds is a symplectomorphism if it preserves the symplectic form, i.e., $f^* \eta = \omega$ with $f^* \eta$ denoting the pullback of $\eta$ by $f$. 

Following Darboux theorem, there exists a canonical chart $(U, \phi)$, $\bm x\in U$ for each point $\bm x\in\mathcalm$ in which the symplectic form is represented as $\bm \omega=\mathbb J_{2n}^\intercal$ via the canonical Poisson tensor 
\begin{equation}
\mathbb J_{2n} = \begin{pmatrix}\bm 0 & \bm I_n\\ -\bm I_n & \bm 0\end{pmatrix}, \:\text{for which} \:\mathbb J_{2n}^\intercal=\mathbb J_{2n}^{-1}=-\mathbb J_{2n}.
    \label{eq:sec2:poisson_tensor}
\end{equation}
In other words, every symplectic manifold is locally symplectomorphic to $(\mathbb{R}^{2n},\mathbb J_{2n}^\intercal)$. 
A system $(\mathbb{R}^{2n}, \mathbb J_{2n}^\intercal, \hamiltonian)$ is called a canonical Hamiltonian system.

In this paper, we consider Hamiltonian systems $(\mathcalm,\bm \omega, \hamiltonian)$, on $\mathcalm$ globally valid canonical symplectic form $\bm\omega = \mathbb J_{2n}^\intercal$. In this case, the phase space $\mathcalm$ can be modeled on the cotangent bundle $\cotangentq{}$ of a smooth $n$-dimensional manifold $\mathcalq$~\citep{Weinstein71:LSubmanifolds} 
with canonical coordinates $(\bmq, \bm p)$ with position $\bm q\in\mathcal{Q}$ and conjugate momenta $\bm p\in\cotangentq{\bmq}$.
The Hamiltonian vector field simplifies to 
\begin{equation}
    \begin{pmatrix}
        \dot{\bmq} \\ \dot{\bmp}
    \end{pmatrix}=\bm X_{\hamiltonian}=\mathbb{J}_{2n}\,d\hamiltonian^\intercal = \begin{pmatrix}\frac{\partial \hamiltonian}{\partial \bmp} \\ -\frac{\partial \hamiltonian}{\partial \bmq}\end{pmatrix}.
\end{equation}
Moreover, the Hamiltonian system $(\cotangentq{},\; \mathbb J_{2n}^\intercal, \hamiltonian)$ relates to a Lagrangian function $\lagrangian:\tangentbundleq \to \euclideanspace$ via the Legendre transform, which takes $\lagrangian$ to $\hamiltonian = \dbmq^\intercal \bmp -  \lagrangian$ with $\bmp = \frac{\partial \lagrangian}{\partial \dbmq}$ and $\dbmq\in\tangentq{\bmq}$.
Mechanical systems often display a quadratic kinetic energy structure, where the configuration manifold $\mathcalq$ is a Riemannian manifold endowed with the kinetic-energy metric equal to the system's mass–inertia matrix $\bmM(\bmq)$. In this case, the Hamiltonian function is given by the sum of the system's kinetic $ T(\bm q, \bm p)$ and potential  $V(\bmq)$ energies as $\hamiltonian = T(\bmq, \bmp) + V(\bmq) = \frac{1}{2}\bmp^\intercal \bmM^{-1}(\bmq)\bmp + V(\bmq)$ and the momenta is $\bmp = \bmM(\bmq)\dbmq$. 

\subsection{Structure-Preserving Hamiltonian MOR}
\label{sec2:sub2:mor}
Given the known parametrized dynamic equations of a high-dimensional system, i.e., a \ac{fom}, \ac{mor} aims to construct a low-dimensional surrogate dynamic model, i.e., a \ac{rom}, that accurately and efficiently approximates the \ac{fom} trajectories. Structure-preserving \ac{mor} preserves the underlying geometric structure of the \ac{fom}, ensuring that its properties, e.g. stability and energy conservation, are maintained in the \ac{rom}. For Hamiltonian systems $(\mathcalm, \bm\omega, \hamiltonian)$, the symplectic structure is preserved by constructing a reduced Hamiltonian $(\check{\mathcalm}, \check{\bm \omega}, \check{\hamiltonian})$ with $\text{dim}(\checkmathcalm) \!=\! d \ll \text{dim}(\mathcalm) \!=\! n$, whose vector field $\check{\bm X}_{\hamiltonian}$ approximates the set of solutions $S = \left\{\bmgamma(t) \in \mathcalm \:\vert\: t \in \mathcal{I} \right\}\subseteq\mathcalm$ of~\eqref{eq:sec2:hamiltonian_IVP}.

Following the geometric perspective of~\citet{Buchfink2024MOR}, the reduced Hamiltonian $(\check{\mathcalm}, \check{\bm \omega}, \check{\hamiltonian})$ is derived by identifying the submanifold $\check{\mathcalm}$ via a smooth embedding 
$\varphi:\check{\mathcalm} \to \mathcalm$ such that 
\begin{equation}
    \check{\bm \omega} = \varphi^*\bm\omega = d\varphi^\intercal \bm \omega d\varphi,
    \label{eq:symplectic_embedding}
\end{equation} 
is non-degenerate. This implies that $(\check{\mathcalm}, \check{\bm \omega})$ is a symplectic manifold and $\varphi$ is a symplectomorphism~\citep[Lemma 5.13]{Buchfink2024MOR}. 
Note that structure-preserving Hamiltonian \ac{mor} typically considers a canonical \ac{fom} $(\mathbb{R}^{2n}, \mathbb J_{2n}^\intercal, \hamiltonian)$ reduced to a canonical \ac{rom} $(\mathbb{R}^{2d}, \mathbb J_{2d}^\intercal, \check\hamiltonian)$~\citep{Peng2015:SympMOR,Sharma2023symplMOR,Buchfink23:SymplecticMOR}. 
The Hamiltonian structure is preserved by constructing $\check{\hamiltonian}$ via the pullback of the embedding as $\check{\hamiltonian} = \varphi^*\hamiltonian = \hamiltonian \circ \varphi$.
Trajectories $\check{\bmgamma}(t)$ of the reduced-order system are then obtained from the \ac{rom} $\ddt\check{\bmgamma}\big\vert_{t} = \check{\bm{X}}_{\check{\hamiltonian}}\big\vert_{\check{\bmgamma}(t)} \in \mathcalt_{\check{\bmgamma}(t)}\check{\mathcalm}$, with $\check{\bm X}_{\check{\hamiltonian}} = \check{\bm \omega}^{-1} d\check{\hamiltonian}$. The reduced initial value $\check{\bmgamma}_0 = \rho(\bmgamma_0) \in \check{\mathcalm}$ is computed via the point reduction map $\rho: \mathcalm \to \checkmathcalm$ associated with $\varphi$, which must satisfy the projection properties 
\begin{equation}
\label{eq:projectionproperties}
    \rho \circ \varphi = \identity{\checkmathcalm} \quad \text{ and } \quad
    d\rho\vert_{\varphi(\check\bmx)} \circ d\varphi\vert_{\check\bmx} = \identity{\mathcal{T}_{\check\bmx}\checkmathcalm},
\end{equation}
$\forall \check \bmx \in \checkmathcalm$.
Trajectories of the original system are finally obtained as the approximation $\bmgamma(t)\approx\varphi(\check{\bmgamma}(t))$.

The embedding $\varphi$ and point reduction $\rho$ are key as they determine the \ac{rom} trajectories. Accurately approximating the \ac{fom} requires minimizing the reconstruction error\looseness-1
\begin{equation}
    \ell_{\text{rec}} = \frac{1}{N} \sum_{i=1}^N \| \varphi \circ \rho (\bm{x}_i) - \bm{x}_i \|^2.
    \label{eq:rec_error}
\end{equation}
Exact reconstruction requires $d\rho$ to be the symplectic inverse of $d\varphi$, i.e., $d\rho = d\varphi^+ = \check{\bm \omega}^{-1} d\varphi^\intercal \bm\omega$. In this paper, we introduce a geometrically-constrained \ac{ae} that fulfills~\eqref{eq:symplectic_embedding} and~\eqref{eq:projectionproperties} by design. 

\subsection{Hamiltonian Neural Networks}
While \ac{mor} reduces the dimensionality of systems with known dynamics, \acp{hnn} aim to learn the unknown dynamics of typically low-dimensional systems while ensuring energy conservation. Most \acp{hnn} assume canonical Hamiltonian systems or Hamiltonian systems with canonical symplectic form~\eqref{eq:sec2:poisson_tensor}. In this paper, we build on two \ac{hnn} variants that \emph{(1)} learn the Hamiltonian function as a single network $\hamiltonian_{\bm{\theta}}(\bmq,\bmp)$~\citep{Greydanus2019HNN}, or \emph{(2)} learn the kinetic and potential energy as two distinct networks, i.e., ${\hamiltonian_{\bm{\theta}}(\bmq,\bmp) = T_{\bm{\theta}_{\text{T}}}(\bmq,\bmp) + V_{\bm{\theta}_{\text{V}}}(\bmq)}$~\citep{zhong2020symplectic,Lutter2023DeLaN}. Given a set of $N$ observations $\left\{\bmq_i, \bmp_i, \dbmq_i, \dbmp_i\right\}_{i=1}^{N}$, the networks are trained to minimize the prediction error of the Hamiltonian vector field via the loss
\begin{equation}
    \ell_{\text{HNN}} = \| \frac{\partial \hamiltonian_{\bm{\theta}}}{\partial \bmp} (\bmq_i, \bmp_i) - \dbmq_i \|^2 + \| \frac{\partial \hamiltonian_{\bm{\theta}}}{\partial \bmq}(\bmq_i, \bmp_i) - \dbmp_i \|^2.
\end{equation}

%% file: texfiles/03method.tex
\section{Geometric Reduced-order Hamiltonian Neural Networks}
\label{sec3:method}

\begin{figure}
\centering\adjustbox{trim=0.4cm 0cm 0.4cm 0cm}{\includesvg[width=1.05\linewidth]{figures/fig_ro_hnn.svg}}
   \caption{Flowchart of the forward dynamics of the geometric \ac{rohnn}. The geometrically-constrained symplectic \ac{ae} (in blue) is built as the cotangent lift of a constrained \ac{ae} (top right). The geometric \ac{hnn} (in brown) is composed of
   two SPD networks, one MLP, and a Strang symplectic integrator.
   }
   \label{fig:figure_ro_hnn}
\end{figure}

We present the geometric reduced-order Hamiltonian neural network (\ac{rohnn}) that learns the unknown dynamics of high-dimensional Hamiltonian systems. 

We focus on systems $(\mathcalm, \mathbb{J}^\intercal_{2n}, \hamiltonian)$ evolving on a phase space $\manifold$ with canonical symplectic form $\mathbb{J}^\intercal_{2n}$ for which the solutions $\bm\gamma(t)$ of the \ac{fom}~\eqref{eq:sec2:hamiltonian_IVP} can be accurately approximated by a substantially lower dimensional surrogate model. Our goal is to learn a reduced Hamiltonian system $(\check{\mathcalm}, \check{\bm \omega}, \check{\hamiltonian})$ via non-intrusive structure-preserving \ac{mor}, where we set $\check{\mathcal M}$ as a phase space with $\check{\bm \omega} = \mathbb{J}^\intercal_{2d}$. Given a set of high-dimensional observations $\{\bmq_i, \bmp_i\}_{i=1}^N$, we identify low-dimensional dynamics by jointly learning a reduced symplectic manifold $(\check{\mathcal M}, \mathbb{J}^\intercal_{2d})$ via a smooth embedding $\varphi$ and a reduction $\rho$, and a latent Hamiltonian function $\check\hamiltonian$.\looseness-1

The proposed \ac{rohnn} ensures the preservation of the Hamiltonian structure by fulfilling three necessary conditions by design: \emph{(1)} the embedding $\varphi$ is a symplectomorphism, or equivalently 
\begin{equation}
    \check{\bm \omega} = \mathbb J_{2d} = d\varphi^\intercal \mathbb J_{2n} d\varphi;
    \label{eq:reduced_canonical_symplectic_form}
\end{equation} 
\emph{(2)} the embedding $\varphi$ and reduction map $\rho$ satisfy the projection properties~\eqref{eq:projectionproperties}; and \emph{(3)} $\check{\mathcal H}$ is a valid Hamiltonian function, thus preserving the reduced energy $\check{\mathcal E} = \mathcal E \circ \varphi$. The \ac{rohnn} fulfill \emph{(1)}-\emph{(2)} via a novel geometrically-constrained symplectic \ac{ae} (Sec.~\ref{sec3:sub2:symplectic_ae}), while \emph{(3)} is guaranteed by a reduced-order geometric \ac{hnn} (Sec.~\ref{sec3:sub1:hamiltonian_rom}), whose trajectories are obtained via symplectic integration (Sec.~\ref{sec3:sub4:symplectic_integration}). Accurate modeling of the high-dimensional dynamics is achieved by jointly training the \ac{ae} and the \ac{hnn} (Sec.~\ref{sec3:sub5:training}). The proposed \ac{rohnn} is illustrated in Fig.~\ref{fig:figure_ro_hnn}.

\subsection{Geometrically-constrained Symplectic Autoencoder}
\label{sec3:sub2:symplectic_ae}
Preserving the geometric structure of the original Hamiltonian \ac{fom} is crucial for the learned \ac{rom} to display similar dynamics. We introduce a geometrically-constrained symplectic \ac{ae} that projects a high-dimensional Hamiltonian system $(\mathcalm, \mathbb{J}^\intercal_{2n}, \hamiltonian)$ onto a low-dimensional nonlinear symplectic manifold $(\check{\mathcalm}, \mathbb{J}^\intercal_{2d})$ such that the reduced system strictly retains the Hamiltonian structure of the \ac{fom}. We parametrize the point reduction $\rho:\mathcalm\!\to\!\check{\mathcalm}$ and embedding $\varphi:\check{\mathcalm}\!\to\!\mathcalm$ as the encoder and decoder of an \ac{ae} designed to satisfy symplecticity~\eqref{eq:reduced_canonical_symplectic_form} and projection properties~\eqref{eq:projectionproperties} by construction. 
To do so, we leverage the cotangent bundle structure of the phase space $\manifold=\cotangentbundleq$.

Given a smooth embedding $\varphiq:\check{\mathcalq}\to\mathcalq$ and associated point reduction $\rhoq:\mathcalq\to\check{\mathcalq}$ satisfying~\eqref{eq:projectionproperties}, we define the cotangent-lifted embedding $\varphi$ and point reduction $\rho$ in canonical coordinates as
\begin{equation}
    \varphi(\checkbmq, \checkbmp) = \left(\begin{matrix}
        \varphiq (\checkbmq)\\ d\rhoq\vert_{\varphi_{\mathcal Q}(\checkbmq)}^\intercal\checkbmp
    \end{matrix}\right) 
    \text{ and } 
    \rho(\bmq, \bmp) = \left(\begin{matrix}
        \rhoq (\bmq) \\ d\varphiq\vert_{\rhoq(\bmq)}^\intercal \bmp
    \end{matrix}\right),
    \label{eq:cotangent_lifted_maps}
\end{equation}
where the pullbacks $d\rhoq\vert_{\varphi_{\mathcal Q}(\checkbmq)}^\intercal \checkbmp$ and $d\varphiq\vert_{\rhoq(\bmq)}^\intercal \bmp$ are computed using the Jacobian of $\rhoq$ and $\varphiq$.
\begin{proposition}
    The reduction map $\rho(\bmq, \bmp)$~\eqref{eq:cotangent_lifted_maps} satisfies the projection properties~\eqref{eq:projectionproperties}.
\end{proposition}
\vspace{-0.5cm}
\begin{proof}
    See App.~\ref{app:proofs}.
\end{proof}
\vspace{-0.2cm}
\begin{proposition}
    The embedding $\varphi(\checkbmq, \checkbmp)$~\eqref{eq:cotangent_lifted_maps} satisfies the symplecticity property~\eqref{eq:reduced_canonical_symplectic_form}.
\end{proposition}
\vspace{-0.5cm}
\begin{proof}
    See App.~\ref{app:proofs}.
\end{proof}
\vspace{-0.3cm}
Note that $d\rho\mathbb J_{2n}d\rho^\intercal =\mathbb J_{2d}$ is shown to hold on $\varphi(\check {\mathcal M})$ with similar arguments.

In practice, we learn the embedding $\varphiq$ and point reduction $\rhoq$ via the constrained AE architecture from~\citet{Otto2023MOR}, and compute their differentials analytically to construct the cotangent-lifted maps~\eqref{eq:cotangent_lifted_maps} (see App.~\ref{appendix1:constrained_ae} for details). The encoder and decoder are given as a composition of feedforward layers ${\rhoq\!=\!\rhoq^{(1)}\!\circ \!\ldots\! \circ \!\rhoq^{(L)}}$ and $\varphiq\!=\!\varphiq^{(L)}\!\circ \!\ldots \!\circ\! \varphiq^{(1)}$ with $\rhoq^{(l)}\!:\! \mathbbr{n_l} \!\to\! \mathbbr{n_{l-1}}$, $\varphiq^{(l)}\!:\! \mathbbr{n_{l-1}} \!\to\! \mathbbr{n_{l}}$, and $n_{l-1} \!\leq\! n_{l}$. 
The key to fulfill the projection properties~\eqref{eq:projectionproperties} is the construction of the layer pairs as
\begin{equation}
\begin{aligned}
    \rhoq^{(l)}(\bmq^{(l)}) &= \sigma_{-}\left(\bm{\Psi}_l^{\trsp}(\bmq^{(l)} -\bm{b}_{l})\right), \\ 
    \varphiq^{(l)}(\checkbmq^{(l-1)}) &= \bm{\Phi}_{l} \sigma_{+}(\checkbmq^{(l-1)}) +\bm{b}_{l},
    \label{eq:orthogonal-layers}
\end{aligned}
\end{equation}
where $(\bm{\Phi}_l, \bm{\Psi}_{l})$ and $(\sigma_{+}, \sigma_{-})$ are pairs of weight matrices and smooth activation functions such that $\bm{\Psi}_{l}^{\trsp} \bm{\Phi}_{l} \!=\! \bm{I}_{n_{l-1}}$ and $\sigma_{-}\circ\sigma_{+} \!=\! \identity{}$, respectively, and $\bm{b}_l$ are bias vectors. Therefore, each layer pair~\eqref{eq:orthogonal-layers} satisfies ${\rhoq^{(l)} \circ \varphiq^{(l)} = \identity{\mathbbr{n_{l-1}}}}$ and the constrained AE fulfills~\eqref{eq:projectionproperties}. Following~\citep{friedl2025reduced}, we ensure that the pairs of weight matrices adhere to the biorthogonality constraint $\bm{\Psi}_{l}^{\trsp} \bm{\Phi}_{l} \!=\! \bm{I}_d$ by accounting for the Riemannian geometry of biorthogonal matrices (see App.~\ref{app1:biorthogonal_manifold} for a background).
Specifically, we consider each pair $(\bm{\Phi}_{l}, \bm{\Psi}_{l})$ as an element of the biorthogonal manifold ${\mathcal{B}_{n_l,n_{l-1}}\!=\!\{ (\bm{\Phi}, \bm{\Psi})\!\in\!\mathbb{R}^{n_l \times n_{l-1}}\!\times\!\mathbb{R}^{n_l \times n_{l-1}} \!:\! \bm{\Psi}^{\trsp}\bm{\Phi}\!=\!\bm{I}_{n_{l-1}} \}}$ and optimize them to minimize the reconstruction error~\eqref{eq:rec_error} via Riemannian optimization~\citep{Absil07:RiemannOpt,Boumal22:RiemannOpt} (see  App.~\ref{app1:Riemannian_opt}). Note that this Riemannian approach was shown to consistently outperform the overparametrization proposed by~\citet{Otto2023MOR}, achieving lower reconstruction errors~\citep{friedl2025reduced}.
The constraint $\sigma_{-}\!\circ\!\sigma_{+}\!=\!\identity{}$ is met by utilizing the smooth, invertible activation functions defined in~\citep[Eq. 12]{Otto2023MOR}, see also App.~\ref{appendix1:constrained_ae}. 

As will be shown in Sec.~\ref{sec4:results}, the resulting geometrically-constrained symplectic \ac{ae} provides increased expressivity compared to linear and quadratic symplectic projection approaches~\citep{Peng2015:SympMOR,Sharma2023symplMOR}, while guaranteeing the symplectic structure of the latent space in constrast to weakly-symplectic \acp{ae} based on soft constraints~\citep{Buchfink23:SymplecticMOR}.
In the intrusive case, i.e., if the \ac{fom} is known, we can construct the reduced Hamiltonian function via the pullback of the cotangent-lifted embedding as $\check{\hamiltonian}=\varphi^*\hamiltonian$, which yields the Hamiltonian \ac{rom}.  Instead, in this paper, we consider the case where the high-dimensional dynamics are unknown, and learn the reduced-order Hamiltonian function $\check{\hamiltonian}$ with a geometric \ac{hnn}, as explained next.

\subsection{Conservative and Dissipative Hamiltonian ROMs}
\label{sec3:sub1:hamiltonian_rom}
\textbf{Conservative dynamics.} We propose to learn the reduced Hamiltonian dynamics in the embedded symplectic submanifold $(\check{\mathcalm}, \mathbb{J}^\intercal_{2d})$ via a \ac{hnn}. 
For general systems, we encode the reduced-order Hamiltonian function as a single neural network $\check{\hamiltonian}_{\bm{\theta}}(\checkbmq,\checkbmp)$ with parameters $\theta$, akin to~\citep{Greydanus2019HNN}. However, additional prior knowledge on the structure of the Hamiltonian is often available. For instance, the Hamiltonian function of mechanical systems sums a quadratic kinetic energy and a potential term. Leveraging that the learned symplectic submanifold preserves the original system structure, we propose to model the reduced Hamiltonian function as $\check\hamiltonian_{\bm{\theta}}(\checkbmq,\checkbmp) = \frac{1}{2}\checkbmp^\intercal \check{\bmM}^{-1}_{\bm{\theta}_{\check{T}}}(\checkbmq)\checkbmp + \check{V}_{\bm{\theta}_{\check{V}}}(\checkbmq)$ via two neural networks $\check{\bmM}^{-1}_{\bm{\theta}_{\check{T}}}$ and $\check{V}_{\bm{\theta}_{\check{V}}}$ with parameters $\bm{\theta}=\{\bm{\theta}_{\check{T}}, \bm{\theta}_{\check{V}}\}$. Existing \acp{hnn} ~\citep{Lutter2023DeLaN, zhong2020dissipative} enforce the symmetric positive-definiteness of the inverse mass-inertia matrix via a Euclidean network encoding its Cholesky decomposition $\bm{L}$, i.e., $\bmM^{-1}=\bm{L}\bm{L}^\intercal$. However, as for \acp{lnn}~\citep{friedl2025reduced}, this parametrization leads to flawed measures of distances in the space of \ac{spd} matrices and ultimately results in inaccurate dynamics predictions. To overcome this issue, we parametrize $\check{\bmM}^{-1}_{\bm{\theta}_\text{T}}$ via the \ac{spd} network from~\citet{friedl2025reduced} that accounts for the Riemannian geometry of the \ac{spd} manifold $\SPDd$ (see Apps.~\ref{appendix:sec:geometry},~\ref{app1:spd_networks}). The network $\check{\bmM}^{-1}_{\bm{\theta}_\text{T}}(\bmq) = (g_{\text{Exp}}\circ g_{\mathbb{R}}) (\bmq)$ is composed of \emph{(1)} a standard Euclidean \ac{mlp} $g_{\mathbb{R}}:\mathbb{R}^d\to\mathbb{R}^{d(d+1)/2}$ that maps the input configuration to the elements of a symmetric matrix $\bm{U}\in\text{Sym}^d$, and \emph{(2)} an exponential map layer $g_{\text{Exp}}$ that interprets $\bm{U}$ as an element of the tangent space $\mathcal{T}_{\bm{P}}\SPDd$, and maps it onto $\SPDd$.

\textbf{Dissipative dynamics.} While classical Hamiltonian dynamics conserve energy, dissipation and external inputs often appear in real-world systems. 
Both can be modeled in \acp{hnn} by complementing the Hamiltonian vector field with a force field $\bm X_{\mathcal F}$, so that the total vector field is $\bm X = \bm X_{\hamiltonian} + \bm X_{\mathcal F}$~\citep{sosanya2022dissipative,zhong2020dissipative}. 
We propose to leverage the structure-preserving symplectic submanifold and model dissipation and external inputs as a reduced-order force field $\bm \check{X}_{\mathcal F}$ on $(\check{\mathcalm}, \mathbb{J}^\intercal_{2d})$. Specifically, we consider high-dimensional systems with observed external inputs $\bmtau_{\text{ext}}$ and viscous damping following a Rayleigh dissipative function $\mathcal{D}(\bmq, \dot{\bmq}) = \frac{1}{2} \dot{\bmq}^\intercal \bm{D}(\bmq)\dot{\bmq}$ with unknown positive semi-definite dissipation matrix $\bm{D}(\bmq)$. The resulting force field is $\bm X_{\mathcal F} = \left(\begin{smallmatrix} \bm{0} \\ \bmtau_{\text{ext}} + \bmtau_{\text{d}} \end{smallmatrix}\right)$ with damping force $\bmtau_{\text{d}} = \frac{\partial \mathcal{D}(\bmq, \dot{\bmq})}{\partial \dot{\bmq}} = \bm{D}(\bmq)\dot{\bmq}$.

\begin{proposition}
    The reduced vector field $\check{\bm X}\!=\!\varphi^*\bm{X}$ obtained via the pullback of the cotangent-lifted embedding $\varphi$~\eqref{eq:cotangent_lifted_maps} preserves the structure of the total vector field $\bm X \!= \!\bm X_{\hamiltonian} \!+\! \bm X_{\mathcal F}$ with $\bm X_{\mathcal F}\! =\! \left(\begin{smallmatrix} \bm{0} \\ \bmtau_{\text{ext}} + \bmtau_{\text{d}}
    \end{smallmatrix}\right)$. 
    \label{prop:reduced_dissipative_vector_field}
\end{proposition}
\vspace{-0.5cm}
\begin{proof}
See App.~\ref{app:proofs}.
\end{proof}
\vspace{-0.3cm}
We propose to model the reduced Rayleigh dissipative function $\check{\mathcal D}_{\bm{\theta}_{\check{D}}}(\checkbmq, \dot{\checkbmq}) \!=\! \frac{1}{2} \dot{\checkbmq}^\intercal \check{\bm D}_{\bm{\theta}_{\check{D}}}(\checkbmq)\dot{\checkbmq}$ via a neural network $\check{\bm D}_{\bm{\theta}_{\check{D}}}$. 
Dissipative \acp{hnn}~\citep{zhong2020dissipative} constrain the dissipation matrix to be positive semi-definite through its Cholesky decomposition, i.e., $\bm{D}\!=\!\bm{L}\bm{L}^\intercal$, thus overlooking its intrinsic geometric structure. Instead, we utilize a second \ac{spd} network $\check{\bm D}_{\bm{\theta}_{\check{D}}}(\checkbmq) = (g_{\text{Exp}}\circ g_{\mathbb{R}}) (\bmq)$. 

Note that the dissipative dynamics no longer preserve a symplectic structure. However, Proposition~\ref{prop:reduced_dissipative_vector_field} shows that the \ac{rohnn} latent space still preserves the structure of high-dimensional dissipative dynamical systems characterized by a Rayleigh dissipative function. It is worth noting that these systems are equivalent to Port-Hamiltonian systems with energy dissipation matrix $\bm{R}(\bmq,\bmp)\!=\!\left(\begin{smallmatrix}\bm{0} & \bm{0} \\ \bm{0} & \bm{D}(\bmq)\end{smallmatrix}\right)$ and input $\bm{G}(\bmq,\bmp)\bm{u}\!=\!\bmtau_{\text{ext}}$, and to contact Hamiltonians $\hamiltonian_c(\bmq,\bmp,s)$ with $\dot{s}\!=\!\mathcal{D}$.

\textbf{Predicting dynamics.} The geometric \ac{hnn} predicts conservative and dissipative dynamics as
\begin{equation}
    \dot{\checkbmq}_{\text{p}} = \frac{\partial \check{\hamiltonian}_{\bm{\theta}}}{\partial \checkbmp} 
        \text{ and } 
    \dot{\checkbmp}_{\text{p}} = -\frac{\partial \check{\hamiltonian}_{\bm{\theta}}}{\partial \checkbmq}  - \check{\bm D}_{\bm{\theta}_D}(\check{\bmq}) \frac{\partial \check{\hamiltonian}}{\partial \check{\bmp}} 
    + \check{\bmtau}_{\text{ext}},
        \label{eq:predicted-hnn-vf}
\end{equation}
where the predictions of the dynamic model are denoted via the subscript $\text{p}$. Note that $\check{\mathcal{D}}_{\bm{\theta}_D}\!=\!0$ and $\check{\bmtau}_{\text{ext}}\!=\!\bm{0}$ in the conservative case. The architecture is illustrated in Fig.~\ref{fig:figure_ro_hnn}. 

Predicting system trajectories according to the learned reduced-order Hamiltonian dynamics involves \emph{(1)} integrating the latent predictions $(\dot{\checkbmq}_{\text{p}},\dot{\checkbmp}_{\text{p}})$~\eqref{eq:predicted-hnn-vf}, and \emph{(2)} decoding the obtained reduced-order position and momentum $(\checkbmq_{\text{p}},\checkbmp_{\text{p}})$ into the high-dimensional coordinates of the original system with the lifted point embedding $\varphi$~\eqref{eq:cotangent_lifted_maps}, i.e.,  $(\tilde{\bmq}_{\text{p}},\tilde{\bmp}_{\text{p}}) =\varphi(\checkbmq_{\text{p}},\checkbmp_{\text{p}})$. In this paper, we propose to integrate the learned reduced-order Hamiltonian flow via symplectic integration, as explained next.

\subsection{Trajectory Prediction via Symplectic Integration}
\label{sec3:sub4:symplectic_integration}
Symplectic integrators are particularly well suited to integrate Hamiltonian dynamics as they preserve the geometric structure and invariants of the Hamiltonian flow~\citep{LeimkuhlerReich05:SimulatingHamiltonian}. They were shown to be key to accurately integrate learned \acp{hnn} dynamics, thus preventing long-term drifting of numerical solutions~\citep{Chen2019symplecticRNNs,Xiong2020:NonsepSymplNN}.

The Hamiltonian dynamics learned in Sec.~\ref{sec3:sub1:hamiltonian_rom} are nonseparable, thus prohibiting the usage of standard explicit integration schemes, e.g., leapfrog~\citep{LeimkuhlerReich05:SimulatingHamiltonian}.
Instead, we integrate the reduced-order Hamiltonian flow~\eqref{eq:predicted-hnn-vf} using the second-order symplectic integrator of~\citep{Tao2016:SymplInt} based on Strang splitting, akin to~\citep{Xiong2020:NonsepSymplNN}. In a nutshell, the integrator considers an augmented Hamiltonian function
\begin{equation*}
    \bar{\hamiltonian}(\bmq,\bmp,\bm x, \bm y) = \hamiltonian(\bmq, \bm y) + \hamiltonian(\bmp, \bm x) + \frac{1}{2}w(\|\bmq, \bm x \|^2 + \| \bmp, \bm y\|^2),
\end{equation*} 
with extended phase space, for which high-order separable symplectic integrators with explicit updates can be constructed. A numerical integrator approximating $\bar{\hamiltonian}$ is obtained by composing the obtained explicit flows, which we refer to as Strang-symplectic integrator. Additional details are provided in App.~\ref{appendix1:symplectic_integrator}.

\subsection{Model Training}
\label{sec3:sub5:training}
Finally, we propose to jointly learn the parameters $\{\bm{\Phi}_{l}, \bm{\Psi}_{l}, \bm{b}_{l}\}_{l=1}^{L}$ of the \ac{ae} and $\{\bm{\theta}_{\check{T}}, \bm{\theta}_{\check{V}},\bm{\theta}_{\check{D}}\}$ of the latent geometric \ac{hnn}. 
As the learned dynamics are expected to predict multiple steps, we consider a loss that numerically integrates the latent predictions $(\dot{\checkbmq}_{\text{p}},\dot{\checkbmp}_{\text{p}})$~\eqref{eq:predicted-hnn-vf} via $H$ forward Strang-symplectic integration steps before decoding.
Given sets of observations $\left\{\bmq_i(\mathcal{I}_i), \bmp_i(\mathcal{I}_i), \bm{\tau}_i(\mathcal{I}_i)\right\}_{i=1}^N$ over intervals $\mathcal{I}_i = \left[t_{i}, t_{i}+H\Delta t\right]$ with constant integration time $\Delta t$, the resulting multi-step loss is

\small
\vspace{-0.8cm}
\begin{align} 
    \label{eq:loss_RO_HNN}
    \ell_{\text{RO-HNN}}
    &= \frac{1}{HN} \sum_{i=1}^N \sum_{j=1}^H 
    \underbrace{\|\tilde{\bmq}_{i}(t_{i,j}) - \bmq_i(t_{i,j}) \|^2}_{\ell_{\text{AE}}}  \\
    &+ \underbrace{\| \tilde{\bmp}_{i}(t_{i,j}) - \bmp(t_{i,j}) \|^2}_{\ell_{\text{AE}}}  + \gamma\; \|\bm{\theta}\|_2^2 \nonumber  \\
    &+  
    \underbrace{\lambda \|\checkbmq_{\text{p}, i}(t_{i,j}) - \checkbmq_i(t_{i,j}) \|^2 +  
    \lambda  \|\checkbmp_{\text{p}, i}(t_{i,j}) - \checkbmp_i(t_{i,j}) \|^2}_{\ell_{\text{HNN},d}} \nonumber \\
    &+ \underbrace{\|\tilde{\bmq}_{\text{p}, i}(t_{i,j}) - \bmq_{i}(t_{i,j})\|^2 + \|\tilde{\bmp}_{\text{p}, i}(t_{i,j}) - \bmp_{i}(t_{i,j})\|^2}_{\ell_{\text{HNN},n}}, \nonumber
\end{align}
\normalsize
where $(\tilde{\bm q}, \tilde{\bm p}) = \rho \circ \varphi (\bmq, \bmp)$, $(\tilde{\bm q}_{\text{p}, i}, \tilde{\bm p}_{\text{p}, i}) = \varphi (\bmq_{\text{p}, i}, \bmp_{\text{p}, i})$, 
$\checkbmq_{\text{p}, i}(t_{i,j}) \!=\! \int_{t_{i}}^{t_{i,j}}  \dot{\checkbmq}_{\text{p},i}  \mathrm{dt}$ and $\checkbmp_{\text{p}, i}(t_{i,j}) \!=\! \int_{t_{i}}^{t_{i,j}}  \dot{\checkbmp}_{\text{p},i} \mathrm{dt}$ with $t_{i,j}=t_{i} + j\Delta t$, and $\lambda, \gamma \in \mathbb{R}_{>0}$ are scaling factors. Note that initial conditions to the latent IVPs are given by the encoded observations at the initial timestep $\check{\bm q}_{\text{p}, i}(t_{i}) = \check{\bm q}_{i}$ and $\check{\bm p}_{\text{p}, i}(t_{i}) = \check{\bm p}_{i}$.
We optimize the network parameters via Riemannian Adam~\citep{becigneul2018riemannianoptimization}, see  App.~\ref{app1:Riemannian_opt}.

%% file: texfiles/04results.tex
\section{Experiments}
\label{sec4:results}

We evaluate the proposed \ac{rohnn} to learn the dynamics of three simulated high-dimensional Hamiltonian systems: a $15$-\ac{dof} pendulum, a $90$-\ac{dof} particle vortex, and a $600$-\ac{dof} thin cloth. Our experiments showcase that \acp{rohnn} accurately predict long-term trajectories of high-dimensional Hamiltonian systems, highlighting the importance of embedding geometric inductive biases as hard constraints in the \ac{ae} and \ac{hnn}.
Details about datasets, network architectures, and model training are provided in App.~\ref{appendix2:experimental_descriptive_appendix}. Additional results and runtimes are provided in Apps.~\ref{appendix3:experiments_in_appendix} and~\ref{appendix3:runtimes}.

\begin{figure*}[tbp]
    \centering
    \makebox[\textwidth][c]{
    \begin{minipage}[t]{0.73\textwidth}
    \vbox to 0.6\textwidth{
        \begin{minipage}[t]{\linewidth}
        \centering
        \captionof{table}{Pendulum: Mean and standard deviation of prediction errors ($\downarrow$) over $10$ test trajectories.}
        \label{tab:bd_pend:pred_errors}
        \resizebox{\linewidth}{!}{
                    \begin{tabular}{cccccc}
         & $H\Delta t$ (s) & \textcolor{blue01}{\ac{rohnn}} & $15$-DoF \ac{hnn} & \textcolor{orange02}{$3$-DoF \ac{hnn}} & \textcolor{green03}{HNKO}
        \\
        \midrule
        
        \multirow{2}{*}{$\frac{\|\tilde{\bmq}_{\text{p}} - \bmq\| }{ \|\bmq\|}$} & $0.25$ &  $\it{(1.66\pm 1.38)\times 10^{-1}}$  &  $(5.33 \pm 6.02)\times 10^{-1}$  & $\bm{(1.22\pm 0.92)\times 10^{-1}}$  & $(5.64\pm 4.41)\times 10^{-1}$\\ 
        
        & $5$ & $\it{(7.08\pm 7.56)\times 10^{-1}}$   & ---   & $\bm{(5.44\pm 6.93)\times 10^{-1}}$  & $(1.32\pm 0.94)\times 10^{0}$\\  
        
        \midrule
        
        \multirow{2}{*}{$\frac{\|\tilde{\bmp}_{\text{p}} - \bmp\| }{ \|\bmp\|}$}  & $0.25$ & $\it{(5.33\pm 5.23)\times 10^{-2}}$   & $(1.76 \pm 2.40)\times 10^{-1}$   & $\bm{(2.50\pm 2.96)\times 10^{-2}}$   & $(5.93\pm 10.73)\times 10^{-1}$\\ 
        
        & $5$ & $\it{(1.98\pm 2.67)\times 10^{-1}}$   & ---   & $\bm{(1.85\pm 3.94)\times 10^{-1}}$   & $(1.23\pm 2.08)\times 10^{0}$\\  
        \end{tabular}
        }
\end{minipage}
        }
        \vspace{-7.5cm}
    \captionof{figure}{Pendulum: Median and quartiles of relative error and reconstructed trajectories of the \ac{rohnn} (\solidblueoneline), $3$-\ac{dof} \ac{hnn} (\solidorangetwoline), and HNKO (\solidgreenthreeline) vs. ground truth (\blackline) for a horizon $H\Delta t\!=\!5$s. The $15$-\ac{dof} \ac{hnn} diverges and is not shown.}
    \label{fig:bd_pend:combination_figure_traj_errors}
    
    \end{minipage}
    
    \hspace{0.02\textwidth}

        \begin{minipage}[t]{0.25\textwidth} 
        \vbox to 1.2\textwidth{
        \centering
        \vspace{-1.8cm}
        \adjustbox{width=1.0\linewidth}{
            \includesvg{figures/bd_pend/fulltraj_bd1_q1_and_rel_q_err_median.svg}
        }
        }
    \end{minipage}
    }
    \vspace{-4.5cm}
\end{figure*}

\begin{table*}[tbp]
\centering

\caption{Pendulum: Mean and standard deviation of reconstruction, prediction, and symplecticity errors ($\downarrow$) of intrusive symplectic dimensionality reduction approaches over $10$ test trajectories.}
\label{tab:bd_pend:ae_rec_and_proj_based_pred_errors}
\resizebox{.75\linewidth}{!}{
\begin{tabular}{ccccc}
 & Linear SMG & Quadr. SMG & Weakly-sympl. \ac{ae} & Geom. Sympl. \ac{ae} (ours) 
\\
\midrule
$\|\tilde{\bmq} - \bmq\| / \|\bmq\|$ & $(2.21 \pm 1.17)\times 10^{-1}$  &  $(2.84 \pm 4.23)\times 10^{0}$  & $(1.43 \pm 0.68)\times 10^{-1}$  & \bm{$(8.84 \pm 6.22)\times 10^{-2}$} \\ 

$\|\tilde{\bmp} - \bmp\| / \|\bmp\|$ & $(4.43 \pm 3.99)\times 10^{-1}$  &  $(2.75 \pm 1.60)\times 10^{-1}$  & $(1.57 \pm 1.55)\times 10^{-1}$  & \bm{$(4.09 \pm 3.99)\times 10^{-2}$} \\ 
\midrule
$\|\tilde{\bmq}_{\text{p}} - \bmq\| / \|\bmq\|$ &  $(2.58\pm 2.33)\times 10^{-1}$  &  $(3.53 \pm 5.18)\times 10^{0}$  & $(7.10\pm 7.02)\times 10^{-1}$  & $\bm{(1.13 \pm 0.92)\times 10^{-1}}$  \\ 

$\|\tilde{\bmp}_{\text{p}} - \bmp\| / \|\bmp\|$ & $(2.16\pm 1.89)\times 10^{-1}$   & $(8.68 \pm 1.04)\times 10^{-1}$   & $(2.16\pm 1.89)\times 10^{-1}$  &  $\bm{(4.68 \pm 4.32)\times 10^{-2}}$ \\ 
\midrule
$\|\mathbb J_{2d} - d\varphi^\intercal \mathbb J_{2n} d\varphi\|$ & $\bm{0.0 \pm 0.0}$ & $\bm{0.0 \pm 0.0}$ & $(1.67 \pm 0.35) \times 10^{-2}$ & $\bm{0.0 \pm 0.0}$ \\
$\|d\rho - d\varphi^+\|$ & $\bm{0.0 \pm 0.0}$ & $\bm{0.0 \pm 0.0}$ & $(5.32 \pm 1.45)\times 10^{0}$ & $(9.53 \pm 5.35)\times 10^{-1}$ \\
\end{tabular}
}
\end{table*}

\subsection{Coupled Pendulum (15 \ac{dof})}
\label{sec4:sub1:bd_coupled_pendulum}

We consider a $15$-\ac{dof} augmented pendulum whose nonlinear dynamics are specified from the symplectomorphism of a latent $3$-\ac{dof} pendulum augmented with a $12$-\ac{dof} mass-spring mesh. As the mesh oscillations are small, the system dynamics are intentionally constructed to showcase a system that is approximately reducible to $3$ dimensions.

\textbf{Learning high-dimensional dynamics.} We train a \ac{rohnn} with latent dimension $d\!=\!3$ and a conservative geometric \ac{hnn} with Strang-symplectic integration on $3000$ observations $\{\bmq_i, \bmp_i\}$ (see App.~\ref{appendix2:pendulum} for details). We compare our \ac{rohnn} with a $15$-\ac{dof} geometric \ac{hnn} that directly learns high-dimensional dynamic parameters, and with a \ac{hnko}~\citep{wrongkoopman2024} that learns a discrete linear predictor embedded in a $100$-dimensional lifted space. For completeness, we also consider a $3$-\ac{dof} geometric \ac{hnn} trained on observations of the latent system. 
Notice that this model would not be deployable in practice as it requires ground truth information, i.e., latent observations, that would not be available.

Table~\ref{tab:bd_pend:pred_errors} reports short- and long-term relative prediction errors over $H\Delta t\!=\!\{0.25,5\}$s. 
The \ac{rohnn} outperforms the $15$-\ac{dof} \ac{hnn} and the \ac{hnko}, leading to significantly lower prediction errors.
Due to the high dimensionality, the $15$-\ac{dof} \ac{hnn} was difficult to train and did not lead to stable long-term predictions. As also shown in Fig.~\ref{fig:bd_pend:combination_figure_traj_errors}, the \ac{hnko} learns stable, but inaccurate long-term predictions. In contrast, the \ac{rohnn} achieves similar long-term predictions as the $3$-\ac{dof} \ac{hnn}, which is expected to perform best as trained directly on the low-dimensional system (see also Figs.~\ref{fig:bd_pend:full_traj_plot_pendulum_3dofs_over_time}-\ref{fig:bd_pend:full_traj_latent_space_rohnn_vs_3dpend} in App.~\ref{appendix3:experiments_pendulum}). This validates the \ac{rohnn} ability to jointly learn a latent symplectic submanifold and associated dynamics. Table~\ref{tab:app_bd_pend:noisy_prediciton_results} in App.~\ref{appendix3:experiments_pendulum} shows that the \ac{rohnn} is also robust to observation noise, consistently outperforming the \ac{hnko}.\looseness-1

\textbf{\ac{ae} architecture.}
The quality of the learned symplectic submanifold is crucial for learning accurate dynamics, as they may systematically deviate from the ground truth if the submanifold does not accurately capture the solution space of the high-dimensional system. We analyze the influence of the reduction method in the \ac{rohnn} and compare the proposed geometrically-constrained symplectic \ac{ae} with linear and quadratic symplectic manifold Galerkin (SMG) projections~\citep{Peng2015:SympMOR,Sharma2023symplMOR} which preserve the symplectic structure by construction, and a weakly-symplectic \ac{ae}~\citep{Buchfink23:SymplecticMOR} which encourages structure preservation via a penalty term in the loss (see App.~\ref{appendix2:pendulum} for details). We train each approach on $3000$ observations of the $15$-\ac{dof} pendulum. Here, we consider an intrusive \ac{mor} setup and project the known \ac{fom} dynamics onto the learned submanifold to predict new trajectories ($H\Delta t=0.25$s). Table~\ref{tab:bd_pend:ae_rec_and_proj_based_pred_errors} shows that, due to their increased expressivity, the \acp{ae} outperform the linear and quadratic projections, with the geometrically-constrained symplectic \acp{ae} achieving the lowest reconstruction and prediction errors (see also Fig.~\ref{fig:bd_pend:intrusive_k_pred_comparison_boxplots} in App.~\ref{appendix3:experiments_pendulum}).
Note that only the geometrically-constrained \ac{ae} yielded stable longer-term predictions (see Fig.~\ref{fig:bd_pend:combination_figure_traj_errors}).
Moreover, only the weakly-symplectic \ac{ae} results in an error on the symplecticity condition~\eqref{eq:reduced_canonical_symplectic_form}, which is expected as both SMG projections and geometrically-constrained symplectic \ac{ae} fulfill it by construction.
Both SMG projections also ensure that the differential $d\rho$ is the symplectic inverse of $d\varphi$ by design, while the geometrically-constrained \ac{ae} leads to a lower error than the weakly-constrained one. Note that jointly training the \ac{ae} with the geometric \ac{hnn} in the \ac{rohnn} is beneficial as it further reduces this error to $(7.42 \pm 1.21)\times 10^{-1}$.

\textbf{Latent \ac{hnn} architecture. } 
We compare the performance of our geometric \ac{hnn} to learn the low-dimensional dynamics of the latent $3$-\ac{dof} pendulum against \emph{(1)} a non-geometric variant that parametrizes the inverse mass-inertia matrix via a Cholesky network, and \emph{(2)} two \acp{hnn} encoded as a single black-box network ${\hamiltonian}_{\bm{\theta}}$, where we consider two \acp{mlp} of $64$- and $256$-neurons width. As shown in Fig.~\ref{fig:3d_pend:full_traj_arch__and_solver_ablations_at_D3000}-\emph{left}, the geometric \ac{hnn} achieves the lowest reconstruction error, followed by the Cholesky \ac{hnn} (see also App.~\ref{appendix3:experiments_pendulum}). This showcases the importance of considering both the quadratic energy structure of mechanical systems, and the geometry of their mass-inertia matrices. 

\textbf{Latent integrator.}
We compare the Strang symplectic integrator against a symplectic leapfrog that disregards that the Hamiltonian is non-separable, and a Runge-Kutta of order $4$ that overlooks its symplectic structure. Fig.~\ref{fig:3d_pend:full_traj_arch__and_solver_ablations_at_D3000}-\emph{middle}, \emph{right} show that the Strang symplectic integrator achieves the lowest reconstruction error and conserves energy best during integration (see also App.~\ref{appendix3:experiments_pendulum}).

\begin{figure*}[tbp]
    \centering
    \adjustbox{trim=0cm 0.2cm 0cm 0.2cm}{
            \includesvg[width=.75\linewidth]{figures/pendulum_3d/full_traj_arch_and_solver_abl_at_D3000.svg}  
        }
    \caption{Pendulum: Ablation of the latent \ac{hnn} architecture (\emph{left}) and latent integrator (\emph{middle}, \emph{right}) of the \ac{rohnn}.
    }
    \label{fig:3d_pend:full_traj_arch__and_solver_ablations_at_D3000}
\end{figure*}

\begin{figure*}
    \centering
    \adjustbox{trim=0cm 0.15cm 0cm 0.0cm}{
            \includesvg[width=\linewidth]{figures/pv_dyns/6d_fulltraj_pv.svg}  
        }
    \caption{Particle vortex: Predicted (\orangecircle, \darkorangecircle, \marooncircle) vs ground truth (\greycircle, \darkgreycircle, \blackcircle) positions. Times beyond $10$s are out of the training data distribution.}
    \label{fig:pv:6dvortex_over_time}
\end{figure*}

\begin{table*}[tbp]
\centering
\caption{Particle vortex: Mean and standard deviation of \ac{rohnn} prediction errors ($\downarrow$) for different \acp{ae} and latent \acp{hnn}, and of \ac{hnko} over $10$ test trajectories.}
\label{tab:pv_dyns:h_vs_spd_table}
\resizebox{0.54\linewidth}{!}{
\begin{tabular}{ccccc}
    \ac{ae}& \ac{hnn} & $d$ & $\|\tilde{\bmq}_{\text{p}} - \bmq\| / \|\bmq\|$ & $\|\tilde{\bmp}_{\text{p}} - \bmp\| / \|\bmp\|$ \\
\toprule
\multirow{4}{*}{Symplectic} & \multirow{2}{*}{Black-box} & 
$6$ & $(6.73\pm2.83)\times 10^{-1}$ & $(6.28\pm2.18)\times 10^{-1}$ \\
& & $10$ & $(7.34\pm3.03)\times 10^{-1}$ & $(7.04\pm2.27)\times 10^{-1}$ \\
\cmidrule(lr){2-5}
& \multirow{2}{*}{Geometric} & $6$ & $\bm{(4.00\pm 2.01)\times 10^{-1}}$ & $\bm{(4.33\pm0.14)\times 10^{-1}}$ \\
& & $10$ & $\bm{(4.44\pm0.60)\times 10^{-1}}$ & $\bm{(4.60\pm 0.32)\times 10^{-1}}$ \\
\midrule
\multirow{4}{*}{Non-Symplectic} & \multirow{2}{*}{Black-box} & 
$6$ & $(6.49\pm3.19)\times 10^{-1}$ & $(6.72\pm3.70)\times 10^{-1}$ \\
& & $10$ & $(7.24\pm1.09)\times 10^{-1}$ & $(7.48\pm2.58)\times 10^{-1}$ \\
\cmidrule(lr){2-5}
& \multirow{2}{*}{Geometric} & $6$ & $(6.14\pm 2.26)\times 10^{-1}$ & $(7.18\pm2.84)\times 10^{-1}$ \\
& & $10$ & $(7.31\pm0.88)\times 10^{-1}$ & $(6.99\pm0.65)\times 10^{-1}$ \\
\midrule
\multicolumn{2}{c}{\ac{hnko}} & $800$ & $(1.60\pm 0.56)\times 10^{0}$ & $(1.52\pm0.85)\times 10^{0}$  \\
\end{tabular}
}
\end{table*}

\subsection{Particle Vortex ($90$-\ac{dof})}
\label{sec4:sub3:pv_sys}

Next, we learn the dynamics of a particle vortex composed of $n=90$ particles with uniform interaction strengths. As the particle vortex dynamics are purely determined via the logarithmic interaction, its Hamiltonian function does not separate into kinetic and potential energies. 
We train \acp{rohnn} with $d\!=\!\{6,10\}$ and (\emph{1}) a geometric \ac{hnn} and \emph{(2)} a black-box \ac{hnn} $\check{\hamiltonian}_{\bm{\theta}}$, both with Strang-symplectic integration (see App.~\ref{appendix2:pv_sys} for details). Fig.~\ref{fig:pv:6dvortex_over_time} depicts the predicted particle positions and momenta for a prediction horizon of $H\!=\!100$, showing that the \ac{rohnn} accurately predicts the particle vortex dynamics and generalizes beyond the data support ($t>10$s). As shown in Table~\ref{tab:pv_dyns:h_vs_spd_table} and Fig.~\ref{fig:pv:k_step_pos_mom_pred_box_err_over_latdim_and_architectures} in App.~\ref{appendix3:experiments_pv}, 
the geometric \acp{hnn} outperform the black-box \acp{hnn} despite the lack of structure of the ground truth Hamiltonian. This suggests that the \ac{ae} learns a symplectomorphism to a latent space where the Hamiltonian can be decomposed into two energy terms, thereby taking advantage of the additional structure of the geometric \ac{hnn}.
To further validate the role of symplecticity-preserving components, we compare variants of the \ac{rohnn} with non-symplectic \ac{ae} for $d\!=\!\{6,10\}$-dimensional geometric and black-box \acp{hnn} $\check{\hamiltonian}_{\bm{\theta}}$.
Compared to the symplectic \ac{ae} variants, the \acp{rohnn} with non-symplectic \ac{ae} degrade the predictive performance of models using a latent geometric \ac{hnn}. This suggests that a symplectic latent space simplifies the convergence of structure-preserving geometric \ac{hnn}. This hypothesis is supported by the fact that the predictive performance remain comparable between both \ac{ae} variant when combined with a latent black-box \acp{hnn}.
Moreover, the $6$-dimensional models slightly outperform the $10$-dimensional ones, showing that the choice of latent dimension trades off between the latent space expressivity and the limitations of \acp{hnn} in higher dimensions (see App.~\ref{appendix3:experiments_pv} for more results).
Finally, Table~\ref{tab:pv_dyns:h_vs_spd_table} also reports the prediction errors of a \ac{hnko}, which is significantly outperformed by all \ac{rohnn} variants, confirming the expressivity of nonlinear latent dynamic models.

\begin{table*}[h]
\centering
\caption{Cloth: Mean and standard deviation of \ac{rohnn} reconstruction and prediction errors ($\downarrow$) for different parametrization of the latent dissipation matrix $\check{\bm{D}}$, and different architecture variants and baselines over $10$ test trajectories. Note that the \ac{rolnn} does not predict momenta $\bm p$.}
\label{tab:cloth:long_traj_rel_errors_different_damping_architectures}
\resizebox{.95\linewidth}{!}{
\begin{tabular}{c c c c c c c c}

    \ac{ae} & \ac{hnn} & $d$ & $\check{\bm{D}}$   & $\|\tilde{\bm q}_{\text p} - \bm q\|/\|\bm q\|$ & $\|\tilde{\bm p}_{\text p} - \bm p\|/\|\bm p\|$ & $\|\check{\bm q}_{\text p} - \check{\bm q}\|/\|\check{\bm q}\|$ & $\|\check{\bm p}_{\text p} - \check{\bm p}\|/\|\check{\bm p}\|$ \\
\toprule
\multirow{6}{*}{Symplectic} &\multirow{6}{*}{Geometric}  & \multirow{3}{*}{$6$} & Cholesky & $(4.21 \pm 1.07)\times 10^{-2}$ &  $(4.11 \pm 2.93)\times 10^{-1}$  & $(3.45 \pm 1.13)\times 10^{-2}$ & $(6.90 \pm 10.90)\times 10^{-2}$ \\
                   & & & \ac{spd} & $(4.15 \pm 2.10)\times 10^{-2}$ & $(3.58 \pm 3.03)\times 10^{-1}$ & $(3.11 \pm 1.86)\times 10^{-2}$ & $(8.42 \pm 10.40)\times 10^{-2}$ \\
                   & & & Ground truth & $(3.18 \pm 0.74)\times 10^{-2}$ & $(3.45 \pm 3.99)\times 10^{-1}$ & $(2.58 \pm 0.94)\times 10^{-2}$ & $(4.77 \pm 5.36)\times 10^{-2}$ \\
\cmidrule(lr){3-8}
& & \multirow{3}{*}{$10$} & Cholesky & $(2.62 \pm 0.74)\times 10^{-2}$ & $(3.39 \pm 3.01)\times 10^{-1}$ & $(1.86 \pm 0.69)\times 10^{-2}$ & $(3.90 \pm 4.36)\times 10^{-2}$ \\
                   & & & \ac{spd} & $(3.21 \pm 1.25)\times 10^{-2}$ & $\bm{(3.33 \pm 2.88)\times 10^{-1}}$ & $(3.45 \pm 1.13)\times 10^{-2}$ & $\bm{(1.96 \pm 1.10)\times 10^{-2}}$ \\
& & & Ground truth & $\bm{(2.31 \pm 0.70)\times 10^{-2}}$ & $(3.44 \pm 3.05)\times 10^{-1}$ & $\bm{(1.37 \pm 0.63)\times 10^{-2}}$ & $(4.26 \pm 4.58)\times 10^{-2}$ \\
\midrule
Symplectic & Black-box & \multirow{2}{*}{$10$} & \multirow{2}{*}{Ground truth} & $(1.95 \pm 3.72)\times 10^{0}$ & $(2.01 \pm 7.59)\times 10^{2}$ & $(1.76 \pm 3.15)\times 10^{0}$ & $(6.83 \pm 8.26)\times 10^{1}$ \\
Non-Symplectic & Geometric &  &   & $(3.14 \pm 3.48)\times 10^{0}$ & $(7.89 \pm 19.04)\times 10^{1}$ & $(2.63 \pm 2.59)\times 10^{0}$ & $(1.37 \pm 3.21)\times 10^{2}$ \\
\midrule
\multicolumn{2}{c}{\ac{rolnn}} & $10$ & Ground truth & $(1.05 \pm 0.78)\times 10^{1}$ & --- & $(1.87 \pm 1.35)\times 10^{-1}$ & --- \\
\multicolumn{2}{c}{\ac{hnko}} & $80$ & Ground truth & $(4.56 \pm 14.13)\times 10^{2}$ & $(3.86 \pm 1.52)\times 10^{2}$ & $(1.76 \pm 0.52)\times 10^{-1}$ & $(4.83 \pm 7.44)\times 10^{-1}$ \\
\end{tabular}
}
\vspace{-0.35cm}
\end{table*}

\subsection{Cloth ($600$-\ac{dof})}
\label{sec4:sub2:cloth}
Next, we learn the dynamics of a high-deformable damped system, namely a simulated $600$-\ac{dof} thin cloth falling onto spheres of different radius, akin to~\citep{friedl2025reduced}. The system is intrinsically damped due to external dissipation forces $\bmtau_{\text{d}}$.
We train two \acp{rohnn} with $d=\{6,10\}$ and a dissipative geometric \ac{hnn} with Strang-symplectic integration on $20$ trajectories of $3000$ observations $\{\bmq_i, \bmp_{i}, \bmtau_{i}\}$ each, where $\bmtau=\bmtau_{\text{c}}$ are measured external constraint forces (see App.~\ref{appendix2:cloth}). Fig.~\ref{fig:cloth:mj_vis_damped_traj_pred_10D_6D_rolnn} depicts the predicted cloth configurations for a horizon $H \Delta t = \SI{0.5}{\second}$, showing that the \ac{rohnn} accurately predicts the high-dimensional dissipative dynamics of the cloth, generalizing beyond the data support ($t>0.3$s).
We observe that the $10-$dimensional model preserves more details of the cloth than the $6$-dimensional one, as also validated by the prediction errors reported in Table~\ref{tab:cloth:long_traj_rel_errors_different_damping_architectures}.
We compare the \ac{rohnn} against a \ac{rolnn}~\citep{friedl2025reduced} with latent dimension $d=10$, for which dissipation forces are not learned but provided as ground truth. The last row of Fig.~\ref{fig:cloth:mj_vis_damped_traj_pred_10D_6D_rolnn} and Table~\ref{tab:cloth:long_traj_rel_errors_different_damping_architectures} show that the \ac{rohnn} predictions are more accurate than the \ac{rolnn}'s, despite that the former learns the dissipation forces via the latent damping matrix. 
Table~\ref{tab:cloth:long_traj_rel_errors_different_damping_architectures} also compares the \ac{rohnn} with a \ac{hnko}, which leads to noticeably worse predictions compared to all reduced-order models. 
We hypothesize this due to the fact that \ac{hnko} does not explicitly enforce the symplectic structure of its high-dimensional($d=80$) latent dynamics.
App.~\ref{appendix3:experiments_cloth} reports additional results, ablations, and comparisons with \acp{rolnn} (see Fig.~\ref{fig:cloth:fulltraj_abl_lnn_arch_10d_cloth_medians_positions}).

\textbf{Latent damping.}
We compare the performance of the dissipative \ac{rohnn} against \emph{(1)} a conservative \ac{rohnn}, where the dissipation forces $\bmtau_{\text{d}}$ are not learned but provided as ground truth in the external input $\bmtau\! =\! \bmtau_{\text{c}}+ \bmtau_{\text{d}}$, and \emph{(2)} a dissipative \ac{rohnn} where the dissipation matrix is parametrized via Cholesky decomposition. Note that the mass-inertia matrix is parametrized via \ac{spd} networks in all cases. 
Table~\ref{tab:cloth:long_traj_rel_errors_different_damping_architectures} shows that both dissipative \acp{rohnn} successfully learn the dissipation forces, achieving similar prediction errors as the conservative \ac{rohnn} (see also Fig.~\ref{fig:cloth:fulltraj_10d_arch_abl_median} in App.~\ref{appendix3:experiments_cloth}). The geometric \ac{hnn} slightly outperforms its Cholesky counterpart, showing the importance of considering geometry. However, the effect is less pronounced as when learning the inverse mass-inertia matrix, which we attribute to the reduced influence of damping compared to inertia in the overall dynamics.
Table~\ref{tab:cloth:long_traj_rel_errors_different_damping_architectures} reports additional results for a non-symplectic \ac{ae} combined with a geometric latent \ac{hnn}, and a symplectic \ac{ae} combined with a black-box latent \ac{hnn}. Both variants operate in a $d=10$-dimensional latent space and receive damping forces as ground truth values. Both variants are significantly outperformed by the \acp{rohnn}, indicating that preserving symplectic structure in both the latent representation and dynamic model contribute to accurate long-horizon prediction.
As the dissipative dynamics do not preserve the symplectic structure, we compare the Strang symplectic integrator, which assumes a symplectic structure, against a non-symplectic Runge-Kutta integrator of order $4$. 
Fig.~\ref{fig:cloth:symplectic_architecture_ablation} in App.~\ref{appendix3:experiments_cloth} shows that, despite the dissipative structure, the Strang symplectic integrator outperforms the Runge-Kutta one. We hypothesize that this is due to the fact that the evolution of this dissipative system is mostly governed by its Hamiltonian function, especially over the short timesteps taken by the integrators.

\begin{figure}
    \centering
    \adjustbox{trim=0.4cm 0.0cm 0.2cm 0cm}{
            \includesvg[width=\linewidth]{figures/cloth/damped_cloth_longterm_pred_lat_dim_comparison_mj_vis_complete.svg}  
        }
    \caption{Predicted positions of the damped cloth with \acp{rohnn} with $d=\{6,10\}$ and \ac{rolnn} with $d=10$ for a $625\times$ longer horizon than during training. Times beyond $0.3$s are out of the training data distribution.}
    \label{fig:cloth:mj_vis_damped_traj_pred_10D_6D_rolnn}
    \vspace{-0.5cm}
\end{figure}

%% file: texfiles/05conclusion.tex
\section{Conclusions} 
\label{sec:conclusion}

This paper proposed a novel physics-inspired neural network, \ac{rohnn}, for learning the unknown dynamics of high-dimensional Hamiltonian systems from data. Our model provides physically-consistent, accurate, and stable predictions that generalize beyond the data support. 
To achieve this, it systematically integrates geometric inductive bias by defining structure-preserving symplectic embeddings, considering the geometry of the dynamics parameters within the model and for optimization, and leveraging structure-preserving symplectic integrators. 
We showed that the structural incorporation of these priors in the architecture is essential to learn high-dimensional dynamics, as Euclidean and soft-constrained approaches consistently underperformed.
Our experiments on systems ranging from $15$ to $600$ \acp{dof} demonstrate that \acp{rohnn} efficiently scale to high-dimensional systems through their reduced descriptions. 
While these results highlight the scalability of \ac{rohnn}, a remaining challenge is managing the trade-off inherent to the latent dimension: Although larger latent spaces can capture more complex dynamics, they increase the difficulty of training of the latent \ac{hnn}. 
Future work will extend \ac{rohnn} to Hamiltonian systems with non-canonical symplectic forms. To do so, we plan to leverage Darboux theorem to explore the development of local \acp{rohnn}. We will also generalize the \ac{rohnn} to learn more general dynamics of Port-Hamiltonian and contact Hamiltonian systems. 
Finally, we will investigate model-based control strategies within the \ac{rohnn} latent space, akin to the controllers developed by~\citet{friedl2026:control} for Lagrangian systems.

%% file: texfiles/technical_appendix.tex
\section{Riemannian and Symplectic Geometry}
\label{appendix:sec:geometry}
In this section, we provide a short background on Riemannian and symplectic geometry, which compose the theoretical backbone of the \ac{rohnn}. We refer the interested reader to~\citep{AbrahamMarsden87:Foundations,Lee13:SmoothManifolds} for more details. 
Table~\ref{tab:geometric_concepts} briefly summarizes the main geometric concepts and their notation.

\begin{table}[h]
\centering
\caption{Main geometric terminology and notation.}
\label{tab:geometric_concepts}
\resizebox{\linewidth}{!}{
\begin{tabular}{c c c}
\textbf{Symbol} & \textbf{Meaning}\\
$\mathcal M$ & smooth manifold, $\text{dim}(\mathcal M)=2n$\\
$\omega$ & symplectic form \\
$(\mathcal M, \bm \omega)$ & symplectic manifold, i.e. smooth manifold $\mathcal M$ equipped with symplectic form $\omega$\\
$\mathcal H:\mathcal M \to \mathbb R$ & Hamiltonian functional defined on a symplectic manifold\\
$(\mathcal M, \bm \omega, \mathcal H)$ & Hamiltonian system, defined by a Hamiltonian $\mathcal H$ evolving on a symplectic manifold $(\mathcal M, \bm \omega)$\\
$\bm {\mathcal X}_{\mathcal H}$ & Hamiltonian vector field, governing system evolution of a Hamiltonian system\\
$\mathbb J_{2n}$ & canonical symplectic form in coordinates\\
$(\mathcal M, \mathbb J_{2n}, \mathcal H)$ & canonical Hamiltonian system, i.e., Hamiltonian system on symplectic manifold with canonical sysmplectic form\\
$\bm q \in \mathcal Q$ & position on configuration manifold, $\text{dim}(\mathcal Q) = n$\\
$\mathcal T_{\bm q} \mathcal Q$ & tangent space of configuration manifold $\mathcal Q$ at position $\bm q$\\
$\mathcal T \mathcal Q$ & tangent bundle of configuration manifold $\mathcal Q$\\
$\mathcal T_{\bm q}^* \mathcal Q$ & cotangent space of configuration manifold $\mathcal Q$ at position $\bm q$\\
$\mathcal T^* \mathcal Q$ & cotangent bundle of configuration manifold $\mathcal Q$\\
$\bm p \in \mathcal T^*\mathcal Q$ & momenta, elements of the cotangent bundle\\
$\varphi_{\mathcal Q}$  & embedding of a low-dimensional submanifold\\
$\rho_{\mathcal Q}$ & reduction onto a low-dimensional submanifold, left inverse of $\varphi_{\mathcal Q}$ on its domain.\\
$\varphi$ & embedding of a low-dimensional symplectic submanifold\\
$\rho$ & reduction onto a low-dimensional symplectic submanifold, left inverse of $\varphi$ on its domain.\\
$\varphi ^*\mathcal H = \mathcal H \circ \varphi$  & pullback of the Hamiltonian $\mathcal H$ by function $\varphi$ via pre-composition\\
$\varphi ^*\bm \omega = d\varphi^\intercal \bm \omega d\varphi $ & pullback of the symplectic form $\bm \omega$ by function $\varphi$\\
$d f$ & differential of a smooth map $f$, corresponding in coordinates to the Jacobian, or for scalar-valued functions, to the transpose gradient\\
$\check{\cdot }$ & reduced, low-dimensional, submanifold quantities\\
symplectomorphism  & map between two symplectic manifolds that preserves the symplectic form\\
cotangent lift & extending a map from a configuration manifold to a cotangent bundle
\end{tabular}
}
\end{table}

Riemannian and symplectic manifolds are smooth manifolds with special structures. A smooth manifold $\mathcalm$ of dimension $n$ can be intuitively conceptualized as a manifold that is locally, but not globally, similar to the Euclidean space $\euclideanspace^n$. The smooth structure of $\mathcalm$ allows the definition of derivative of curves on the manifold, which are tangent vectors. The set of all tangent vectors at a point $\bm{x}\in\mathcalm$ defines the tangent space $\tangentm{\bm{x}}$ which is a $n$-dimensional vector space. Tangent vectors can be represented on an ordered basis of $\tangentm{\bm{x}}$ as $\bm{v}=v^i\frac{\partial}{\partial x^i}|_{\bm{x}}$. The tangent bundle $\tangentbundlem$ is the disjoint union of all tangent spaces on $\mathcalm$ and is $2n$-dimensional smooth manifold.

The cotangent space $\cotangentm{\bm{x}}$ at $\bm{x}\in\mathcalm$ is the dual of the tangent space $\tangentm{\bm{x}}$, i.e., ${\cotangentm{\bm{x}}=\{\lambda | \lambda:\tangentm{\bm{x}}\to\euclideanspace \;\text{linear}\}}$. Cotangent vectors can be represented on an ordered basis of $\cotangentm{\bm{x}}$ as $\bm{\lambda}=\lambda_i\mathrm{d}x^i|_{\bm{x}}$. The cotangent bundle $\cotangentbundlem$ is the disjoint union of all cotangent spaces on $\mathcalm$ and is $2n$-dimensional smooth manifold, similarly as the tangent bundle.

A smooth mapping $f$ between two smooth manifolds $\checkmathcalm$ and $\mathcalm$ with $\text{dim}(\checkmathcalm) \!=\! d \ll \text{dim}(\mathcalm) \!=\! n$ is an immersion if the differential $\mathrm{d}f|_{\check{\bm{x}}}:  \mathcal{T}_{\check{\bm{x}}}\mathcalm\to\tangentm{f(\check{\bm{x}})}$ is injective at each $\check{\bm{x}}\in\checkmathcalm$. An embedding is an immersion that is also a homeomorphism onto its image, i.e., it is an injective and structure-preserving map. In this case, $\checkmathcalm$ is an embedded submanifold of $\mathcalm$.
The pullback of a function $h:\mathcalm\to\euclideanspace$ by a smooth mapping $f:\mathcal{N}\to\mathcalm$ between two smooth manifolds $\mathcal{N}$ and $\mathcalm$ is a smooth function $f^*h$ with
\begin{equation}
    f^*h(\bm{x}) = h(f(\bm{x})) = (h \circ f)(\bm{x}).
\end{equation}

\subsection{Riemannian Geometry}
A Riemannian manifold $(\mathcalm,g)$ is a smooth manifold $\mathcalm$ endowed with a Riemannian metric $g$, i.e., a smoothly-varying inner product $g_{\bm{x}}:\tangentm{\bm{x}}\times\tangentm{\bm{x}}\to\euclideanspace$. In coordinates, a Riemannian metric is represented by a \ac{spd} matrix. The Riemannian metric defines the notion of distance on the manifold, as well as the so-called geodesics, which are length-minimizing curves on the manifold. 

Learning and optimization methods involving Riemannian data typically take advantage of their Euclidean tangent spaces to operate. Specifically, the exponential map $\expmapblank{\bm x} :\tangentm{\bm{x}}\to\mathcalm$ and logarithmic map $\logmapblank{\bm x}:\mathcalm\to\tangentm{\bm{x}}$, derived from the Riemannian metric, allows us to map back and forth between the Euclidean tangent space and the manifold. Moreover, the parallel transport $\prltrspblank{\bm{x}}{\bm{y}}: \tangentm{\bm{x}}\to\tangentm{\bm{y}}$ move tangent vectors across tangent spaces such that their inner product is conserved.

A Lagrangian system $(\mathcalm,g,\lagrangian)$ is a dynamical system evolving on a Riemannian manifold $(\mathcalm,g)$ according to a smooth Lagrangian function $\lagrangian:\tangentbundlem\to\euclideanspace$.

\subsection{Symplectic Geometry}
A symplectic manifold $(\mathcalm, \omega)$ is a $2n$-dimensional smooth manifold $\mathcalm$ equipped with a symplectic form $\omega$, i.e., a closed, non-degenerate, differential $2$-form $g_{\bm{x}}:\tangentm{\bm{x}}\times\tangentm{\bm{x}}\to\euclideanspace$, which satisfies 
\begin{equation}
\omega(\bm{u},\bm{v})=-\omega(\bm{v}, \bm{u}), \quad \omega(\bm{u},\bm{v})\;\forall \bm{v} \Rightarrow \bm{u}=\bm{0}, \quad \text{and} \quad d\omega=0  
\end{equation}
for all $\bm{u},\bm{v}\in\tangentm{\bm{x}}$. In coordinates, a symplectic form is represented by a skew-symmetric matrix $\bm\omega$. We slightly abuse notation, equivalently denoting symplectic manifolds as $(\mathcalm, \bm\omega)$. Notice that the non-degeneracy of $\omega$ implies that all symplectic manifolds are of even dimension.

A diffeomorphism $f:(\mathcalm, \bm \omega)\to(\mathcal N, \bm \eta)$ between symplectic manifolds is a symplectomorphism if it preserves the symplectic form, i.e., $f^* \eta = \omega$ with $f^* \eta$ denoting the pullback of $\eta$ by $f$. 
The Hamiltonian flow $\phi_{t}:(\mathcalm, \bm \omega) \to (\mathcalm, \bm \omega)$ induced by $\bm X_{\hamiltonian}$ is a symplectomorphism, as it maps points $\bm{x}\in \mathcalm$ along the integral curves of the manifold thus preserving the symplectic form.

Following Darboux' theorem, there exists a canonical chart $(U, \phi)$, $\bm x\in U$ for each point $\bm x\in\mathcalm$ in which the symplectic form is represented as $\bm \omega=\mathbb J_{2n}^\intercal$ via the canonical Poisson tensor 
\begin{equation*}
\mathbb J_{2n} = \left(\begin{matrix}\bm 0 & \bm I_n\\ -\bm I_n & \bm 0\end{matrix}\right), \quad \text{for which} \quad \mathbb J_{2n}^\intercal=\mathbb J_{2n}^{-1}=-\mathbb J_{2n}.
\end{equation*}
In other words, every symplectic manifold is locally symplectomorphic to $(\mathbb{R}^{2n},\mathbb J_{2n}^\intercal)$. 
A system $(\mathbb{R}^{2n}, \mathbb J_{2n}^\intercal, \hamiltonian)$ is called a canonical Hamiltonian system.
Moreover, the cotangent bundle $\mathcal T^*\mathcalm$ any $n$-dimensional smooth manifold $\mathcalm$ carries a canonical symplectic structure, making it a symplectic manifold $(\mathcal T^*\mathcalm, \mathbb{J}_{2n})$.

A Hamiltonian system $(\mathcalm, \bm\omega, \hamiltonian)$ is a dynamical system evolving on a symplectic manifold $(\mathcalm, \bm \omega)$ according to a smooth Hamiltonian function $\hamiltonian :\mathcalm\to \mathbb R$.

\section{Riemannian Manifolds of Interest}
\label{app1:riemannian_manifolds_of_interest}
This section provides a brief overview of the Riemannian manifolds of interest for this paper, namely the manifold of \ac{spd} matrices $\SPD$ (App.~\ref{app1:spd_manifold}), and the biorthogonal manifold $\mathcal{B}_{n,d}$ (App.~\ref{app1:biorthogonal_manifold}).

\subsection{The Manifold of SPD Matrices}
\label{app1:spd_manifold}

We denote the set of $n\times n$ symmetric matrices as $\text{Sym}^n=\{\bm{S}\in \mathbbr{n \times n} | \bm{S} = \bm{S}^\intercal\}$.
The set of SPD matrices ${\SPD = \left\{\bm{\Sigma} \in \text{Sym}^n \:\rvert\:\bm{\Sigma} \succ \mathbf{0}\right\}}$ forms a smooth manifold of dimension $\text{dim}(\SPD) = \frac{n(n+1)}{2}$, which can be represented as the interior of a convex cone embedded in $\text{Sym}^n$. The tangent space $\mathcal{T}_{\bm{\Sigma}}\SPD$ at a point ${\bm{\Sigma}\in\mathcal{S}_{\ty{++}}^n}$ is identified with $\text{Sym}^n$.

The \ac{spd} manifold can be endowed with various Riemannian metrics, resulting in different theoretical properties and closed-form operations. We utilize the widely-used affine-invariant metric~\citep{pennecAIM}, which places symmetric matrices with non-positive eigenvalues at infinite distance from any \ac{spd} matrix and prevents the well-known swelling effect~\citep{feragenfuster2017}.
The affine-invariant metric defines the inner product $g:\mathcal{T}_{\bm{\Sigma}} \SPD \times \mathcal{T}_{\bm{\Sigma}} \SPD \to \mathbb{R}$ given two matrices $\bm{T}_1$, $\bm{T}_2\in\mathcal{T}_{\bm{\Sigma}} \SPD$, as
\begin{equation}
\langle \bm{T}_1,\bm{T}_2 \rangle_{\bm{\Sigma}} \;\;=\;\; \tr(\bm{\Sigma}^{-\frac{1}{2}}\bm{T}_1\bm{\Sigma}^{-1}\bm{T}_2\bm{\Sigma}^{-\frac{1}{2}}).
\label{Eq:SPDinnerprod}
\end{equation} 
The corresponding geodesic distance, exponential map, logarithmic maps, and parallel transport are computed in closed form as
\begin{align}
\manifolddist{\bm{\Lambda}}{\bm{\Sigma}} &= \|\log(\bm{\Sigma}^{-\frac{1}{2}}\bm{\Lambda}\bm{\Sigma}^{-\frac{1}{2}})\|_\text{F}, \\
\expmap{\bm{\Sigma}}{\bm{S}} &= \bm{\Sigma}^{\frac{1}{2}}\exp(\bm{\Sigma}^{-\frac{1}{2}}\bm{S}\bm{\Sigma}^{-\frac{1}{2}})\bm{\Sigma}^{\frac{1}{2}}, \\
\logmap{\bm{\Sigma}}{\bm{\Lambda}} &= \bm{\Sigma}^{\frac{1}{2}}\log(\bm{\Sigma}^{-\frac{1}{2}}\bm{\Lambda}\bm{\Sigma}^{-\frac{1}{2}})\bm{\Sigma}^{\frac{1}{2}}, \\
\prltrsp{\bm{\Sigma}}{\bm{\Lambda}}{\bm{T}} &= \bm{A}_{\bm{\Sigma}\to\bm{\Lambda}} \; \bm{T} \; \bm{A}_{\bm{\Sigma}\to\bm{\Lambda}}^\trsp,
\label{Eq:SPDmaps}
\end{align}
where $\exp(\cdot)$ and $\log(\cdot)$ denote the matrix exponential and logarithm functions, and $\bm{A}_{\bm{\Sigma}\to\bm{\Lambda}} = \bm{\Lambda}^{\frac{1}{2}}\bm{\Sigma}^{-\frac{1}{2}}$.
These operations are key for the \ac{spd} networks encoding the mass-inertia and damping matrices in geometric \acp{hnn} (see Sec.~\ref{sec3:sub1:hamiltonian_rom}), and for the on-manifold parameter optimization of \ac{spd} parameters of the network when training the model (see Sec.~\ref{sec3:sub5:training}).

\subsection{The Biorthogonal Manifold}
\label{app1:biorthogonal_manifold}
The biorthogonal manifold is the smooth manifold ${\mathcal{B}_{n,d}=\{ (\bm{\Phi}, \bm{\Psi})\in\mathbb{R}^{n \times d}\times\mathbb{R}^{n \times d}\:\rvert\:\bm{\Psi}^{\trsp}\bm{\Phi}=\bm{I}_d \}}$ formed by pairs of full-row-rank matrices $\bm{\Phi}, \bm{\Psi}\in\mathbbr{n\times d}$, with $n\geq d \geq 1$ satisfying the biorthogonality condition $\bm{\Psi}^\trsp\bm{\Phi} = \mathbf{I}$~\citep{Otto2023MOR}. The biorthogonal matrix manifold $\mathcal{B}_{n,d}$ is an embedded submanifold of the Euclidean product space $\mathbbr{n\times d} \times \mathbbr{n\times d}$ with dimension $\text{dim}\left(\mathcal{B}_{n,d}\right) = 2nd - d^2$.
The tangent space at a point $\biopair \in \bio$ is given by
\begin{equation}
    \label{eq:bio_tangentspace}
    \Tbio = \left\{\left(\bm{V}, \bm{W}\right) \in \mathbbr{n\times d}\times \mathbbr{n \times d}\;:\; \bm{W}^\trsp\bm{\Phi} + \bm{\Psi}^\trsp\bm{V} = \mathbf{0}\right\}.
\end{equation}
A pair of matrices $(\bm{X},\bm{Y})\in \mathbbr{n\times d} \times \mathbbr{n\times d}$ can be projected onto the tangent space $\Tbio$ via the projection operation $\text{Proj}_{\biopair} :\mathbb{R}^{n \times d}\times\mathbb{R}^{n \times d} \to \Tbio$ defined as 
\begin{equation}
    \label{eq:bio_man_proj}
    \projblank{\biopair}{\bm{X},\bm{Y}} = \left(\bm{X} - \bm{\Psi}\bm{A}, \bm{Y} - \bm{\Phi}\bm{A}^\trsp\right),
\end{equation}
where $\bm{A}$ is a solution to the Sylvester equation
$\bm{A}(\bm{\Phi}^\trsp\bm{\Phi}) + (\bm{\Psi}^\trsp\bm{\Psi})\bm{A} = \bm{Y}^\trsp \bm{\Phi} + \bm{\Psi}^\trsp\bm{X}$.

When optimizing the parameters of the geometrically-constrained symplectic \ac{ae} presented in Sec.~\ref{sec3:sub2:symplectic_ae}, it is crucial to account for the biorthogonal geometry of the pairs of weight matrices~\citep{friedl2025reduced}. Therefore, we train the model by optimizing pairs of weight matrices via Riemannian optimization on the biorthogonal manifold (see Sec.~\ref{sec3:sub5:training}). Riemannian optimization algorithms utilize the exponential map and the parallel transport operations, which are difficult to obtain in closed form for the biorthogonal manifold. Therefore, we leverage a first-order approximation of the exponential map, i.e., a retraction map $\mathrm{R}_{\biopair}:\Tbio \to \bio$, defined as
\begin{equation}
    \label{eq:retraction_bio}
    \retrblank{\biopair}{\bm{V}, \bm{W}} = \left((\bm{\Phi} + \bm{V})\left((\bm{\Psi + \bm{W}})^\trsp(\bm{\Phi}+\bm{V})\right)^{-1}, \left(\bm{\Psi} + \bm{W}\right)\right).
\end{equation}
Moreover, we use a first-order approximation of the parallel transport operation defined via the successive application of retraction and projection as 
\begin{equation}
    \prltrspblank{\left(\bm{\Phi}_1, \bm{\Psi}_1\right)}{\left(\bm{\Phi}_2, \bm{\Psi}_2\right)} = \text{Proj}_{\left(\bm{\Phi}_2, \bm{\Psi}_2\right)} \circ \mathrm{R}_{\left(\bm{\Phi}_1, \bm{\Psi}_1\right)}.
\end{equation}

\section{SPD Network}
\label{app1:spd_networks}
As explained in Sec.~\ref{sec3:sub1:hamiltonian_rom}, we learn reduced Hamiltonian dynamics in the embedded symplectic submanifold via a latent geometric \ac{hnn} that parametrizes the inverse mass-inertia and damping matrices via \ac{spd} networks that account for their intrinsic geometry. We use a \ac{spd} network introduced in~\citep{friedl2025reduced} composed of \emph{(1)} Euclidean layers $g_{\mathbb{R}}$, and \emph{(2)} an exponential map layer $g_{\text{Exp}}$, which we detail next.

\textbf{Euclidean Layers} $g_{\mathbb{R}}$. The SPD network leverages classical fully-connected layers to model functions that return elements on the tangent space of a manifold. The output of the $l$-th Euclidean layer $\bm{x}^{(l)}$ is given by 
\begin{equation}
    \label{eq:ff_nn_euclideanlayer}
    \bm{x}^{(l)} = \sigma(\bm{A}_{l}\bm{x}^{(l-1)} + \bm{b}_{l}),
\end{equation}
with $\bm{A}_{l} \in\mathbbr{n_{l}\times n_{l-1}}$ and $\bm{b}_{l}\in\mathbbr{n_{(l)}}$ the weight matrix and bias of the layer $l$, and $\sigma$ a nonlinear activation function of choice. 

\textbf{Exponential Map Layer} $g_{\text{Exp}}$. The exponential map layer is used to map layer inputs $\bm{X}^{(l-1)} \in \text{Sym}^n$ from the tangent space onto the manifold $\SPD$. The layer output is given by
\begin{equation}
    \label{eq:exp_layer}
    \bm{X}^{(l)} = \text{Exp}_{\bm{P}}(\bm{X}^{(l-1)}),
\end{equation}
with $\bm{P}\in \SPD$ denoting the basepoint of the considered tangent space. Following the results of the ablation conducted in~\citep{friedl2025reduced}, we define $\bm{P}$ as equal to the identity matrix $\mathbf{I}$, so that the layer input is assumed to lie in the tangent space at the origin of the cone. 

Note that~\citet{friedl2025reduced} additionally consider \ac{spd} layers mapping \ac{spd} matrices to \ac{spd} matrices, analogous to fully-connected Euclidean layers. However, the \ac{spd} networks with additional \ac{spd} layers were shown to achieve similar performances as those employing solely Euclidean and exponential-map layers. Therefore, we do not integrate such layers in the \ac{spd} networks of the \ac{rohnn}. 

\section{Proofs of Propositions of Sec.~\ref{sec3:method}}
\label{app:proofs}

\begin{proposition}
    The reduction map $\rho(\bmq, \bmp)$~\eqref{eq:cotangent_lifted_maps} satisfies the projection properties~\eqref{eq:projectionproperties}.
\end{proposition}
\begin{proof}
    It is clear that the cotangent-lifted map $\rho$ fulfills~\eqref{eq:projectionproperties} as $\rhoq$ satisfies~\eqref{eq:projectionproperties} by assumption.
\end{proof}

\begin{proposition}
    The embedding $\varphi(\checkbmq, \checkbmp)$~\eqref{eq:cotangent_lifted_maps} satisfies the symplecticity property~\eqref{eq:reduced_canonical_symplectic_form}.
\end{proposition}
\begin{proof}
    Proving the statement is equivalent to show that the differential $d\varphi = \left(\begin{smallmatrix}
        d\varphiq & \bm 0 \\ \frac{\partial ( d\rhoq^\intercal \checkbmp)}{\partial \checkbmq} & d\rhoq^\intercal
    \end{smallmatrix}\right)$
    belongs to the symplectic Stiefel manifold $\sympstiefel(2n, 2d) = \left\{\bm U \in \euclideanspace^{2n\times 2d}\:\vert \: \bm U^\intercal \mathbb J_{2n}  \bm U = \mathbb J_{2d}\right\}$.
    A block matrix $\bm U = \left(\begin{smallmatrix}
        \bm A & \bm B \\ \bm C & \bm D
    \end{smallmatrix}\right)$ belongs to $\sympstiefel(2n, 2d)$ if its block elements satisfy the condition
    \small
    \begin{equation*}
    \bm U ^\intercal \mathbb J_{2n}^\intercal \bm U = 
        \left(\begin{matrix}
            \bm A^\intercal &\bm C^\intercal \\ \bm B^\intercal & \bm D^\intercal
        \end{matrix}\right)\left(\begin{matrix}
            \bm 0 & -\bm{I}_n\\ \bm I_n &\bm 0
        \end{matrix}\right)\left(\begin{matrix}
            \bm A & \bm B \\ \bm C & \bm D
        \end{matrix}\right) = \left(\begin{matrix}
            \bm C^\intercal \bm A - \bm A^\intercal\bm C & \bm C^\intercal \bm B - \bm A^\intercal \bm D \\ \bm D^\intercal \bm A - \bm B^\intercal \bm C& \bm D^\intercal \bm B - \bm B^\intercal \bm D 
        \end{matrix}\right) = \left(\begin{matrix}
            \bm 0 & -\bm{I}_d\\ \bm I_d &\bm 0
        \end{matrix}\right),
        \label{eq:equation9_general} 
    \end{equation*}
    \normalsize
i.e., the differential $d\varphi$ must satisfy 
\begin{equation}
d\varphi^\intercal \mathbb J^\intercal d\varphi = \left(\begin{matrix}
        \bm C^\intercal d\varphiq-d\varphiq^\intercal\bm C & -d\varphiq^\intercal d\rhoq^\intercal \\ d\rhoq d\varphiq & \bm 0 
    \end{matrix}\right) = \left(\begin{matrix}
            \bm 0 & -\bm{I}_d\\ \bm I_d &\bm 0
        \end{matrix}\right) \quad \text{with} \quad \bm C=\frac{\partial (d\rhoq\vert_{\varphi_{\mathcal Q}(\checkbmq)}^\intercal \checkbmp)}{\partial \checkbmq}.
    \label{eq:equation9} 
\end{equation}
By assumption, $\rhoq$ fulfills the projection properties~\eqref{eq:projectionproperties}, so that $d\rhoq d\varphiq=d\varphiq^\intercal d\rhoq^\intercal=\bm I_d$ holds by construction. It remains to prove $\bm C^\intercal d\varphiq-d\varphiq^\intercal\bm C = \bm 0$. We denote the elements of the canonical and reduced canonical coordinates as $q^{\custunderline{i}}$, $p_{\custunderline{i}}$ and $\check q^{\custunderline{\alpha}}$, $\check p_{\custunderline{\alpha}}$, respectively. By definition, we have ${(d\varphiq)^{\custunderline{i}}_{\custunderline{\alpha}}=\frac{\partial q^{\custunderline{i}}}{\partial \check q^{\custunderline{\alpha}}}}$ and $(d\rhoq)^{\custunderline{\alpha}}_{\custunderline{i}} = \frac{\partial \check q^{\custunderline{\alpha}}}{\partial q^{\custunderline{i}}}$ and the projection properties hold by assumption, i.e., $(d\rhoq)^{\custunderline{\alpha}}_{\custunderline{i}} (d\varphiq)^{\custunderline{i}}_{\custunderline{\beta}} = \delta^{\custunderline{\alpha}}_{\custunderline{\beta}}$ $\forall \check q\in\check\mathcalq$, with $\delta^{\custunderline{\alpha}}_{\custunderline{\beta}}\!=\!1$ if $\alpha\!=\!\beta$ and $\delta^{\custunderline{\alpha}}_{\custunderline{\beta}}\!=\!0$ otherwise. Therefore, we have $p_{\custunderline{i}} = (d\rhoq)^{\custunderline{\alpha}}_{\custunderline{i}} \check p_{\custunderline{\alpha}}$, and 
\begin{equation*}
    C_{\custunderline{i\gamma}} = \frac{\partial p_{\custunderline{i}}}{\partial \check q^{\custunderline{\gamma}}} = \frac{\partial}{\partial \check q^{\custunderline{\gamma}}}((d\rhoq)^{\custunderline{\alpha}}_{\custunderline{i}} \check p_{\custunderline{\alpha}}) = \frac{\partial (d\rhoq)^{\custunderline{\alpha}}_{\custunderline{i}}}{\partial q^{\custunderline{j}}}\frac{\partial q^{\custunderline{j}}}{\partial\check q^{\custunderline{\gamma}}}\check p_{\custunderline{\alpha}} = \frac{\partial(d\rhoq)^{\custunderline{\alpha}}_{\custunderline{i}}}{\partial q^{\custunderline{j}}} (d\varphiq)^{\custunderline{j}}_{\custunderline{\gamma}}\check p_{\custunderline{\alpha}}.
\end{equation*}
We aim to show that $\bm C^\intercal d\varphiq$ is symmetric, i.e., $(d\varphiq)^{\custunderline{i}}_{\custunderline{\beta}} C_{\custunderline{i\gamma}} = (d\varphiq)^{\custunderline{i}}_{\custunderline{\gamma}} C_{\custunderline{i\beta}}$.
Using the projection properties, we can write $\check p_{\custunderline{\beta}} = (d\varphiq)^{\custunderline{i}}_{\custunderline{\beta}} (d\rhoq)^{\custunderline{\alpha}}_{\custunderline{i}} \check p_{\custunderline{\alpha}}$. Differentiating with respect to $\check q^{\custunderline{\gamma}}$ yields 
\begin{align*}
\vspace{-0.2cm}
    0 = \frac{\partial}{\partial \check q^{\custunderline{\gamma}}}\big((d\varphiq)^{\custunderline{i}}_{\custunderline{\beta}} (d\rhoq)^{\custunderline{\alpha}}_{\custunderline{i}} \check p_{\custunderline{\alpha}}\big) &= \frac{(d\varphiq)^{\custunderline{i}}_{\custunderline{\beta}}}{\partial \check q^{\custunderline{\gamma}}} (d\rhoq)^{\custunderline{\alpha}}_{\custunderline{i}} \check p_{\custunderline{\alpha}} + (d\varphiq)^{\custunderline{i}}_{\custunderline{\beta}} \frac{(d\rhoq)^{\custunderline{\alpha}}_{\custunderline{i}}}{\partial q^{\custunderline{j}}}\frac{\partial q^{\custunderline{j}}}{\partial \check q^{\custunderline{\gamma}}} \check p _{\custunderline{\alpha}} \\ &=  \frac{(d\varphiq)^{\custunderline{i}}_{\custunderline{\beta}}}{\partial \check q^{\custunderline{\gamma}}} p_{\custunderline{i}} + (d\varphiq)^{\custunderline{i}}_{\custunderline{\beta}} C_{\custunderline{i\gamma}} = \frac{\partial q^{\custunderline{i}}}{\partial \check q^{\custunderline{\beta}} \partial \check q^{\custunderline{\gamma}}} p_{\custunderline{i}} + (d\varphiq)^{\custunderline{i}}_{\custunderline{\beta}} C_{\custunderline{i\gamma}}.
\end{align*}
As the Hessian in the first term is symmetric, the equality implies the symmetricity of the second term, i.e.,
\begin{equation*}
    (d\varphiq)^{i}_{\beta} C_{\custunderline{i\gamma}} = (d\varphiq)^{i}_{\gamma} C_{\custunderline{i\beta}},
\end{equation*} 
and thus~\eqref{eq:equation9} holds.
\end{proof}
Note that $d\rho\mathbb J_{2n}d\rho^\intercal =\mathbb J_{2d}$ is shown to hold on $\varphi(\check {\mathcal M})$ with similar arguments.
Moreover, a similar proof is presented by~\citet{Sharma2023symplMOR} in the context of quadratic symplectic projections.

\begin{proposition}
    The reduced vector field $\check{\bm X}\!=\!\varphi^*\bm{X}$ obtained via the pullback of the cotangent-lifted embedding $\varphi$~\eqref{eq:cotangent_lifted_maps} preserves the structure of the total vector field $\bm X \!= \!\bm X_{\hamiltonian} \!+\! \bm X_{\mathcal F}$ with $\bm X_{\mathcal F}\! =\! \left(\begin{smallmatrix} \bm{0} \\ \bmtau_{\text{ext}} + \bmtau_{\text{d}}
    \end{smallmatrix}\right)$. 
    \label{prop-app:reduced_dissipative_vector_field}
\end{proposition}
\begin{proof}
The reduced vector field decomposes as $\check{\bm X}\!=\!\varphi^*\bm{X} \!=\!\varphi^*\bm{X}_\hamiltonian + \varphi^*\bm{X}_\mathcal{F} \!=\! \check{\bm X}_{\check \hamiltonian} + \check{\bm X}_{\check{\mathcal F}}$ with $\check{\bm X}_{\check{\hamiltonian}} \!=\! \check{\bm \omega}^{-1} d\check{\hamiltonian}$ (see Sec.~\ref{sec2:sub2:mor}) and $\check{\bm X}_{\check{\mathcal F}}$ is obtained by pulling back the external and damping terms. Since generalized forces belong to the cotangent bundle $\cotangentbundleq$, they are embedded and reduced via the cotangent-lifted maps~\eqref{eq:cotangent_lifted_maps} as $\varphi(\check{\bmq},\check{\bmtau})$ and $\rho(\bmq, \bmtau)$, leading to the reduced external inputs ${\check{\bmtau}_{\text{ext}}=d\varphiq\vert_{\checkbmq}^\intercal \; \bmtau_{\text{ext}}}$. The reduced Rayleigh dissipative function is obtained via the pullback of the tangent-lifted embedding $\varphi_{\tangentbundleq}(\check{\bmq},\dot{\check{\bmq}}) \!=\! (\varphiq(\check{\bmq})^\intercal, (d\varphiq|_{\check{\bmq}}\dot{\check{\bmq}})^\intercal)^\intercal$ as $\check{\mathcal D} \!=\! \varphi_{\tangentbundleq}^* \mathcal D \!=\! \frac{1}{2}\dot{\check{\bm q}}^\intercal\check{\bm D}(\check{\bm q})\dot{\check{\bm q}}$ with positive semi-definite reduced damping matrix $\check{\bm D}(\check {\bm q}) \!=\! d\varphi_{\mathcal Q}^\intercal\bm D(\bm q) d\varphi_{\mathcal Q}$. The reduced damping force is then $\check{\bmtau}_{\text{d}} = \frac{\partial \check{\mathcal{D}}(\check{\bmq}, \dot{\check{\bmq}})}{\partial \dot{\check{\bmq}}} = \check{\bm{D}}(\check{\bmq})\dot{\check{\bmq}}$. Therefore, the reduced force field is $\check{\bm X}_{\check{\mathcal F}} = \left(\begin{smallmatrix} \bm{0} \\ \check{\bmtau}_{\text{ext}} + \check{\bmtau}_{\text{d}} \end{smallmatrix}\right)$.
\end{proof}

\section{Additional Details on the Geometrically-constrained Symplectic Autoencoder}
\label{appendix1:constrained_ae}
\subsection{Constrained Autoencoder}
The geometrically-constrained symplectic \ac{ae} presented in Sec.~\ref{sec3:sub2:symplectic_ae} builds on the constrained \ac{ae} architecture introduced in~\citep{Otto2023MOR}. Specifically, we learn the embedding $\varphiq$ and associated point reduction $\rhoq$ via a constrained \ac{ae} with layer pairs~\eqref{eq:orthogonal-layers}, and compute their differential to construct the tangent-lifted maps~\eqref{eq:cotangent_lifted_maps}, as explained in Sec.~\ref{sec3:sub2:symplectic_ae}. 
To guarantee the projection properties, the constrained \ac{ae} architecture from~\citep{Otto2023MOR} leverages pairs of biorthogonal weight matrices, which are described in Sec.~\ref{sec3:sub2:symplectic_ae}, and pairs of invertible activation functions, which we introduce next. 

The nonlinear activation functions $\sigma_{-}$ and $\sigma_{+}$ employed in the encoder and decoder network must satisfy $\sigma_{-} \circ \sigma_{+} = \text{id}$. To do so, they are defined as 
\begin{equation}
    \label{eq:AE_activation_functions}
    \sigma_{\pm}(x_i) = \frac{bx_i}{a}\mp \frac{\sqrt{2}}{a \sin(\alpha)} \pm \frac{1}{a}\sqrt{\left(\frac{2x_i}{\sin(\alpha)\cos(\alpha)}\mp\frac{\sqrt{2}}{\cos(\alpha)}\right)^2 + 2a},
\end{equation}
with 
\begin{equation}
    \begin{cases}
        a &= \csc^2(\alpha) - \sec^2(\alpha),\\
        b &= \csc^2(\alpha) + \sec^2(\alpha).
    \end{cases}
\end{equation}
The activations then resemble smooth, rotation-symmetric versions of the common leaky ReLu activations. The parameter $0<\alpha<\frac{\pi}{4}$ sets the slope of the activation functions. Throughout our experiments, we set $\alpha = \frac{\pi}{8}$. 

\citet{Otto2023MOR} proposed to incorporate the biorthogonality of the weight matrices by considering an overparametrization of the biorthogonal weights along with a soft constraint in the form of additional penalty losses. However, this approach does not guarantee the biorthogonality condition, in contrast to the Riemannian approach we use in this paper. 
Moreover, as shown in~\citep{friedl2025reduced}, the overparametrized model leads to higher reconstruction errors compared to constrained \ac{ae} trained on the biorthogonal manifold. 

\subsection{Computation of the Cotangent-lifted Maps}
We construct the cotangent-lifted maps~\eqref{eq:cotangent_lifted_maps} by differentiating the outputs of the encoder $\rhoq$ and decoder $\varphiq$ networks with respect to their inputs. To avoid the computational cost related to the automatically-differentiated transposed Jacobian-vector product, our implementation computes layer-wise analytical derivatives and obtains the full differentials via the chain rule. 
The derivatives of the nonlinear activations $\sigma_{\pm}$ are given by 
\begin{equation}
    \label{eq:AE_activation_functions_derivatives}
    \sigma_{\pm}'(x_i) = \frac{d}{dx_i}\sigma_{\pm}(x_i) = \frac{b}{a} \pm \frac{2}{a \sin(\alpha)\cos(\alpha)} \frac{\frac{2x_i}{\sin(\alpha)\cos(\alpha)}\mp\frac{\sqrt{2}}{\cos(\alpha)}}{\sqrt{\left(\frac{2x_i}{\sin(\alpha)\cos(\alpha)}\mp\frac{\sqrt{2}}{\cos(\alpha)}\right)^2 + 2a}},
\end{equation} 
thus fulfilling the inverse-derivative property $\sigma_{-}'(\sigma_{+}(x_i)) \, \sigma_{+}'(x_i) = 1$ by construction.

The pullbacks $d\rhoq\vert_{\varphi_{\mathcal Q}(\checkbmq)}^\intercal \checkbmp$ and $d\varphiq\vert_{\rhoq(\bmq)}^\intercal \bmp$ are computed analytically as a composition of transposed layer derivatives ${d\rhoq^\intercal = {d\rhoq^{(L)}}^\intercal \circ \ldots \circ {d\rhoq^{(1)}}^\intercal}$ and $d\varphiq={d\varphiq^{(1)}}^\intercal\circ \ldots \circ {d\varphiq^{(L)}}^\intercal$, with ${d\rhoq^{(l)}\in \mathbbr{n_{l-1}\times n_{l}}}$ and ${d\varphiq^{(l)}\in \mathbbr{n_{l}\times n_{l-1}}}$. 
From the definition of the layer pairs~\eqref{eq:orthogonal-layers}, the transpose of the layer derivatives are given as
\begin{equation}
    {d\rhoq^{(l)}}\vert_{\bmq^{(l-1)}}^\intercal = \bm{\Psi}_l \text{diag}(\sigma_{-}'(\bmq^{(l-1)})) \quad\quad \text{and} \quad\quad
    {d\varphiq^{(l)}}\vert_{\checkbmq^{(l-1)}}^\intercal = \text{diag}(\sigma_{+}'(\checkbmq^{(l-1)})) \bm{\Phi}_{l}^\intercal,
    \label{eq:orthogonal-layers-transpose-derivatives}
\end{equation} 
with $\text{diag}(\bm{v})=\left(\begin{smallmatrix}
    v_1 & \ldots & 0 \\ \vdots & \ddots & \vdots \\ 0 &\ldots & v_d 
\end{smallmatrix}\right)$.

The computation of the layer derivatives requires storing the intermediate reduced and reconstructed positions $\checkbmq^{(l-1)}$ and $\bmq^{(l-1)}$ for each layer, which are obtained during the forward pass through the position encoder $\rhoq$ and decoder $\varphiq$. 
During the momentum forward pass, we store each intermediate $\bmp^{(l-1)} = {d\rhoq^{(l)}}^\intercal \bmp^{(l)}$ and $\checkbmp^{(l)} = {d\varphiq^{(l)}}^\intercal \bmp^{(l-1)}$.
This allows the computational cost of one momentum forward pass to roughly be equal to that of one forward pass of the position projection, scaling constantly through the matrix-multiplication of weights with system dimensionality $\text{dim}(\mathcal Q).$ We provide wall-clock evaluation times of our geometrically-constrained symplectic AE on the $600$-\ac{dof} cloth dataset in App.~\ref{appendix2:cloth}.

\section{Strang-Symplectic Integrator}
\label{appendix1:symplectic_integrator}

As explained in Sec.~\ref{sec3:sub4:symplectic_integration}, we integrate the learned reduced-order Hamiltonian flow~\eqref{eq:predicted-hnn-vf} using the second-order symplectic integrator of~\citep{Tao2016:SymplInt}, which we refer to as Strang-symplectic integrator. 

The Strang-symplectic integrator approximates the flow of a non-separable Hamiltonian function $\hamiltonian(\bmq, \bmp)$ by considering an augmented Hamiltonian function
\begin{equation}
    \bar{\hamiltonian}(\bmq,\bmp,\bm x, \bm y) = \hamiltonian_A(\bmq, \bm y) +  \hamiltonian_B(\bmp, \bm x) + w\hamiltonian_C (\bmq,\bmp,\bm x, \bm y),
\end{equation}
in an extended phase space, where $\hamiltonian_A(\bmq, \bm y)$ and $\hamiltonian_B(\bmp, \bm x)$ are two copies of the original system with mixed-up positions and momenta, and $\hamiltonian_C = \frac{1}{2} ( \|\bmq, \bm x \|^2 + \| \bmp, \bm y\|^2)$ is an artificial restraint with parameter $w$ controlling the binding of $\hamiltonian_A(\bmq, \bm y)$ and $\hamiltonian_B(\bmp, \bm x)$.
The dynamics of the augmented Hamiltonian $\bar{\hamiltonian}$ are
\begin{align}
    \dot{\bm{q}} &= \frac{\partial}{\partial \bmp} \bar{\hamiltonian}(\bmq,\bmp,\bm x, \bm y) = \frac{\partial}{\partial \bmp} H(\bm{x}, \bmp) + w(\bmp - \bm y) \\
    \dot{\bm{p}} &= \frac{\partial}{\partial \bmq} \bar{\hamiltonian}(\bmq,\bmp,\bm x, \bm y) = \frac{\partial}{\partial \bmq} H(\bm{q}, \bm y) - w(\bmq - \bm x) \\
    \dot{\bm{x}} &= \frac{\partial}{\partial \bm y} \bar{\hamiltonian}(\bmq,\bmp,\bm x, \bm y) = \frac{\partial}{\partial \bm y} H(\bm{q}, \bm y) + w(\bm y - \bm p) \\ 
    \dot{\bm{y}} &= \frac{\partial}{\partial \bmp} \bar{\hamiltonian}(\bmq,\bmp,\bm x, \bm y) = \frac{\partial}{\partial \bmp} H(\bm{x}, \bmp) - w(\bm x - \bmq)
\end{align}
and leads to the same exact \ac{ivp} solutions as the original function $\hamiltonian(\bmq, \bmp)$.
High-order symplectic integrators can be construct for each of the component of the augmented Hamiltonian $\bar{\hamiltonian}$ as
\begin{align}
    \phi_{\hamiltonian_A}^\delta &= \left( \begin{matrix}
        \bmq \\ 
        \bmp - \delta \frac{\partial}{\partial \bmq} H(\bm{q}, \bm y) \\
        \bm x + \delta \frac{\partial}{\partial \bm y} H(\bm{q}, \bm y) \\
        \bm y
    \end{matrix} \right), \quad 
    \phi_{\hamiltonian_B}^\delta = \left( \begin{matrix}
        \bmq + \delta \frac{\partial}{\partial \bm p} H(\bm{x}, \bm p) \\ 
        \bmp  \\
        \bm x  \\
        \bm y - \delta \frac{\partial}{\partial \bmx} H(\bm{x}, \bmp)
    \end{matrix} \right), \\
    \phi_{w\hamiltonian_C}^\delta &= \frac{1}{2} \left( \begin{matrix}
        \left(\begin{matrix}
            \bmq + \bm x \\ \bmp + \bm y
        \end{matrix}\right) 
        + \bm{R}(\delta)
        \left(\begin{matrix}
            \bmq - \bm x \\ \bmp - \bm y
        \end{matrix}\right)
        \\
        \left(\begin{matrix}
            \bmq + \bm x \\ \bmp + \bm y
        \end{matrix}\right) 
        - \bm{R}(\delta)
        \left(\begin{matrix}
            \bmq - \bm x \\ \bmp - \bm y
        \end{matrix}\right)
    \end{matrix} \right), \quad \text{with} \quad
    \bm{R}(\delta) = \left( \begin{matrix}
        \cos(2w\delta)\bm{I} & \sin(2w\delta)\bm{I} \\
        -\sin(2w\delta)\bm{I} & \cos(2w\delta)\bm{I}
    \end{matrix}\right).
\end{align}
\citet{Tao2016:SymplInt} proposed to construct a numerical symplectic integrator that approximates the flow of $\bar{\hamiltonian}$ by composing these maps according to Strang splitting as
\begin{equation}
    \phi_{\bar{\hamiltonian}} = \phi_{\hamiltonian_A}^{\delta/2} \circ \phi_{\hamiltonian_B}^{\delta/2} \circ \phi_{w\hamiltonian_C}^{\delta/2} \circ \phi_{\hamiltonian_B}^{\delta/2} \circ \phi_{\hamiltonian_A}^{\delta/2}.
\end{equation}
The obtained Strang-symplectic integrator preserves the symplectic volume like the exact Hamiltonian flow.

The scalar parameter $w\in \mathbb R$, binding the two augmented Hamiltonians during the integration process, is obtained as optimization parameter during training. To enforce $w \leq 0$, we do not learn $w$ directly. Instead, we learn it using the SoftPlus function with a small numerical offset for stability as $\log(1+e^{\bm{\theta}_w}) + 10^{-4}$, as part of the \ac{hnn} network parameters $\bm \theta_w \in \bm \theta$.

\section{Network Training via Riemannian Optimization}
\label{app1:Riemannian_opt}
Training a neural network corresponds to finding a solution to an optimization problem 
\begin{equation}
    \min_{\bm{x}\in\manifold} \ell(\bm{x}),
\end{equation} 
where $\ell$ is the loss we aim to minimize, and $\bm{x}\in \mathcal M$ is the optimization variable, a.k.a the network parameters. For the \ac{rohnn}, we train the network by minimizing the loss $\ell_{\text{RO-HNN}}$~\eqref{eq:loss_RO_HNN}. In this case, $\manifold$ is defined as a product of Euclidean, \ac{spd}, and biorthogonal manifolds to jointly optimize the parameters $\{\bm{\Phi}_{l}, \bm{\Psi}_{l}, \bm{b}_{l}\}_{l=1}^{L}$ of the \ac{ae} and $\{\bm{\theta}_{\check{T}}, \bm{\theta}_{\check{V}},\bm{\theta}_{\check{D}}\}$ of the latent geometric \ac{hnn}. To account for the curvature of the non-Euclidean parameter spaces, we leverage Riemannian optimization~\citep{Absil07:RiemannOpt,Boumal22:RiemannOpt} to optimize the \ac{rohnn} loss $\ell_{\text{RO-HNN}}$~\eqref{eq:loss_RO_HNN}.

Conceptually, each iteration step in a first-order (stochastic) Riemannian optimization method consists of the three following successive operations: 
\begin{equation}
\bm{\eta}_t \gets h\big(\text{grad}\: \ell(\bm{x}_t), \bm{\tau}_{t-1}\big), \quad
\bm{x}_{t+1} \gets \expmap{\bm{x}_t}{-\alpha_t \bm{\eta}_t}, \quad
\bm{\tau}_t \gets \prltrsp{\bm{x}_t}{\bm{x}_{t+1}}{\bm{\eta}_t}.
\label{eq:RiemannianGD}
\end{equation}
where
\emph{(1)} given the current parameter estimate $\bm{x}_t$, a search direction $\bm{\eta}_t\in\tangentspace{\bm{x}_t}$ is computed based on a function $h$ (determined by the choice of the optimization method) of the Riemannian gradient $\text{grad}\:\ell$, and of $\bm{\tau}_{t-1}$, which corresponds to the parallel-transport of the previous search direction on to the new estimate's tangent space $\tangentspace{\bm{x}_t}$; \emph{(2)} the estimate $\bm{x}_t$ is updated by projecting the search direction $\bm{\eta}_t$ scaled by a learning rate learning rate $\alpha_t$ onto the manifold via the exponential map, \emph{(3)} the current search direction is parallel-transported to the tangent space of the updated estimate to prepare for the next iteration.  
In this paper, we use the Riemannian Adam~\citep{becigneul2018riemannianoptimization} implemented in Geoopt~\citep{geoopt} to optimize the \ac{rohnn} parameters. 
The relevant manifold operations for the optimization procedure are given in closed-form in App.~\ref{app1:riemannian_manifolds_of_interest}.

%% file: texfiles/experimental_descriptive_appendix.tex
\section{Additional Details on Experiments}
\label{appendix2:experimental_descriptive_appendix}
This section presents additional details on the experimental setup of Sec.~\ref{sec4:results}.

\subsection{Coupled Pendulum of Section \ref{sec4:sub1:bd_coupled_pendulum}}
\label{appendix2:pendulum}
\subsubsection{Dataset}
\label{appendix2:pendulum:sub:data_generation}

\begin{figure}[tbp]
    \centering
    \includegraphics[width=0.3\textwidth]{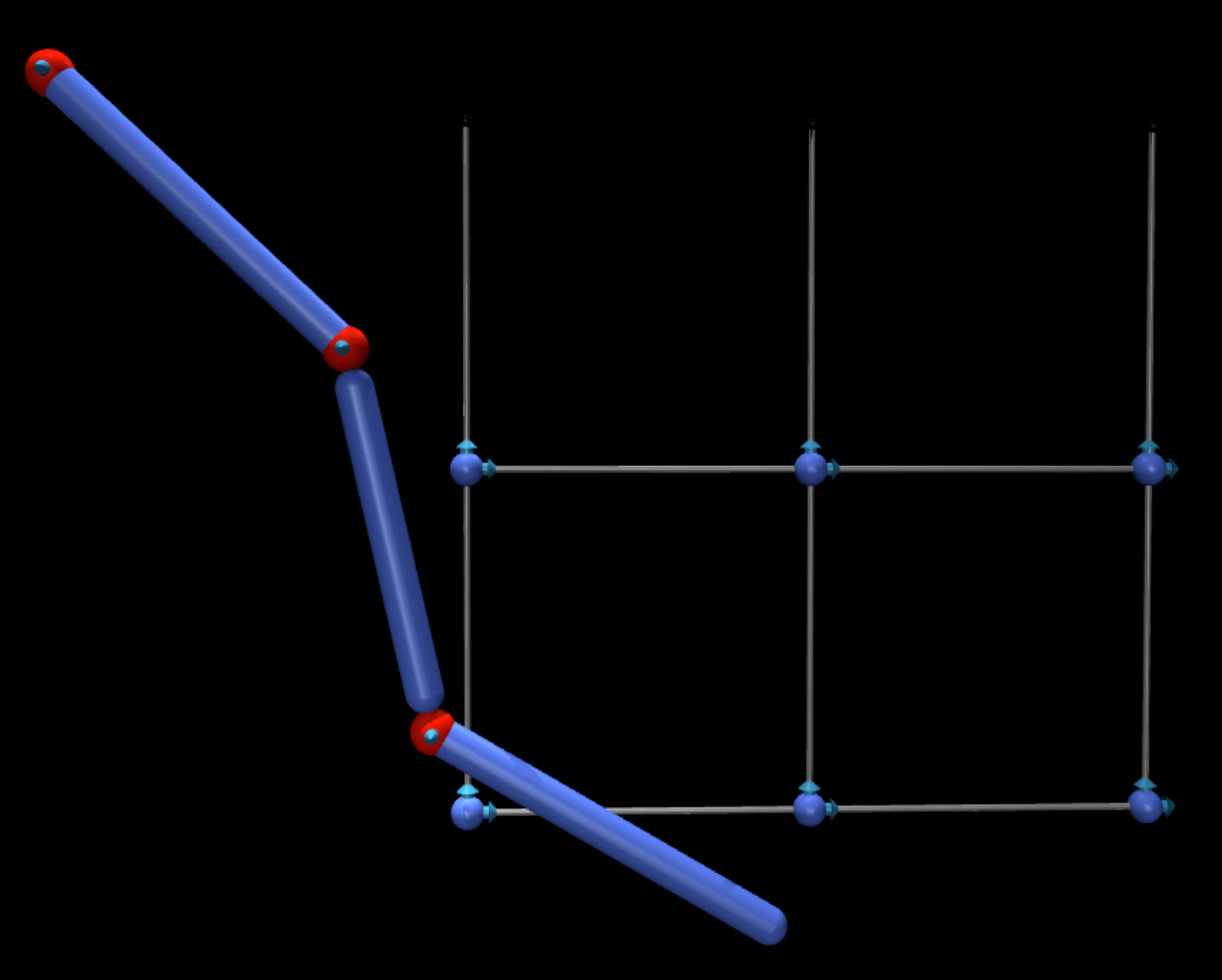}
    \caption{Illustration of the latent system used to obtain the dynamics of a $15$-\ac{dof} augmented pendulum via a symplectomorphism. The latent system consists of an independent $3$-\ac{dof} planar pendulum and a $12$-\ac{dof} planar mass-spring mesh.}
    \label{fig:bd_pend:mj_model_screenshot}
\end{figure}

\textbf{System.} Our first set of experiments is conducted on the augmented pendulum, a nonlinear conservative system with $n\!=\!15$-\ac{dof}. The pendulum dynamics are specified from the symplectomorphism of a latent Hamiltonian system composed of two independent subsystems: a $3$-\ac{dof} planar pendulum, and a $12$-\ac{dof} planar oscillating mass-spring mesh, see Fig.~\ref{fig:bd_pend:mj_model_screenshot}. The pendulum dynamics evolve on a slower timescale and with larger amplitude than the mesh oscillations. Consequently, a surrogate model based solely on the pendulum would capture the dominant behavior of the full system, i.e. the system is well-reducible with a Hamiltonian \ac{rom}. As we have access to the ground truth dynamics of the $15$-\ac{dof} pendulum, this scenario allows for various ablations on the network architecture.

We simulate both subsystems in \textsc{Mujoco}~\citep{Todorov12:mujoco}. The pendulum's links $i=\{1,2,3\}$ are modeled as capsules of radius $r_i = \SI{0.025}{\m}$, length $l_i = \SI{0.5}{\m}$, and mass $m_i = \SI{0.5}{\kg}$, connected via hinge joints. The initial configurations and velocities for each \ac{dof} are randomly sampled from the intervals $q_{\text{pend}, i}(t=0) \in \left[-30, 30\right]\SI{}{\degree}$ and ${\dot q_{\text{pend}, i}(t=0) \in \left[-23, 23\right]\SI{}{\degree\per\second}}$. 
The mass-spring mesh consists of $6$ masses $m_j = \SI{0.005}{\kg}$, equally spaced in a $3\times 2$ grid along the $x-$ and $z-$axes of the simulation environment. Each mass is connected to its immediate neighbors, and the top three masses are each additionally connected to a fixed anchor point above the grid, via springs of resting length $s_j=\SI{0.5}{\m}$ and linear stiffness constants $k_j=\SI{0.01}{\newton\per\meter}$. 
Initial displacements and velocities for each \ac{dof} are randomly sampled from the intervals $q_{\text{ms}, j}(t=0) \in \left[-1, 1\right]\times10^{-2}\SI{}{\meter}$ and $\dot q_{\text{ms}, j}(t=0) \in \left[-2, 2\right]\times 10^{-3}\SI{}{\meter\per\second}$.

\textbf{Data generation.} Each simulation is recorded for $T=\SI{5}{\s}$ at a timestep of $\Delta t = 10^{-2}\SI{}{\second}$, yielding $N=30$ training trajectories $\mathcal{D}_{\text{pend}} = \{\{\bmq_{\text{pend}, n,k}, \bmp_{\text{pend}, n,k}\}_{k=1}^{K}\}_{n=1}^{N}$ and $\mathcal{D}_{\text{ms}} = \{\{\bmq_{\text{ms}, n,k}, \bmp_{\text{ms}, n,k}\}_{k=1}^{K}\}_{n=1}^{N}$ with $K=500$ observations each.
To form the full $15$-dimensional dataset, the position and momentum vectors of the pendulum and mass-spring mesh are concatenated as $\bm q_{\text{aug}} = (\begin{matrix} \bmq_{\text{pend}}^\intercal & \bmq_{\text{ms}}^\intercal \end{matrix})^\intercal$ and $\bm p_{\text{aug}} = (\begin{matrix} \bmp_{\text{pend}}^\intercal, & \bmp_{\text{ms}}^\intercal \end{matrix})^\intercal$ to obtain a dataset $\mathcal{D}_{\text{aug}} = \{\{\bmq_{\text{aug}, n,k}, \bmp_{\text{aug}, n,k}\}_{k=1}^{K}\}_{n=1}^{N}$ with $(\bmq_{\text{aug}, n,k}, \bmp_{\text{aug},n,k})\in \mathcal T^*(\mathcal Q_{\text{pend}} \times \mathcal Q_{\text{ms}})$.

To ensure that the reducibility of the augmented dataset is not purely of numerical nature, we transform the observed dynamics of the latent system onto more complex ones via a symplectomorphism $h:(\mathcal T^*(\mathcal Q_{\text{pend}} \times \mathcal Q_{\text{ms}}), \mathbb{J}_{2n})\to (\mathcal T^*\mathcal Q, \mathbb{J}_{2n})$ and obtain the final dataset $\mathcal{D} =  \{\{h(\bmq_{\text{aug}, n,k}, \bmp_{\text{aug}, n,k})\}_{k=1}^{K}\}_{n=1}^{N}$.
Practically, the symplectomorphism $h$ is defined via the cotangent-lifted embedding $\varphi$ of a map $\rho_{\mathcal Q}: Q_{\text{pend}} \times \mathcal Q_{\text{ms}} \to \mathcal Q$, that we parametrize as a 3-layer encoder of the constrained AE from Sec.~\ref{sec3:sub2:symplectic_ae} and \ref{appendix1:constrained_ae}. With $l=\{1,2,3\}$ layers of constant layer and latent dimension $n_l=n_0=15$, weights of each layer $\bm\Psi_l$ initialized as random orthogonal matrices $\bm O \in \mathbb{R}^{n_l\times n_l}$ sampled from the Haar distribution, and zero biases $\bm b_l=\mathbf 0$. Notice that due to the constant dimension through the AE-layers, with decoder weights set to  $\bm\Phi_l=\bm\Psi_l = \bm O$, the position decoder of the constrained \ac{ae} returns an analytic inverse, and its cotangent lift $h^{-1}$.

The testing dataset is constructed in the same manner for $N=10$ trajectories.

\subsubsection{Model Training}
\label{appendix2:pendulum:sub:model_training}
For the experiments of Sec.~\ref{sec4:sub1:bd_coupled_pendulum}, we train a geometric \ac{rohnn} composed of a geometrically-constrained symplectic \ac{ae} and a latent geometric \ac{hnn}. As described in Sec.~\ref{sec3:sub2:symplectic_ae}, the geometrically-constrained symplectic \ac{ae} is built from the cotangent lift of a constrained \ac{ae} composed of layer pairs $\rhoq^{(l)}: \mathbbr{n_l} \to \mathbbr{n_{l-1}}$ and $\varphiq^{(l)}: \mathbbr{n_{l-1}} \to \mathbbr{n_{l}}$ as defined in~\eqref{eq:orthogonal-layers} (see Sec.~\ref{sec3:sub2:symplectic_ae} and App.~\ref{appendix1:constrained_ae}). We use $l=\{1,2,3\}$ pairwise biorthogonal encoder and decoder layers of sizes $n_l = \{6, 12, 15\}$ with latent space dimension $n_0=3$. The biorthogonal weight matrices are initialized by sampling a random orthogonal matrix $\bm O \in \mathbb{R}^{n_l\times n_l}$ from the Haar distribution and setting $\bm\Phi=\bm\Psi=\bm O_{[:,:n_{l-1}]}$, where $\bm O_{[:,:n_{l-1}]}$ are the first $n_{l-1}$ column entries of $\bm O$. Bias vectors are initialized as $\bm{b}_{l}=\bm{0}$.
For the latent geometric HNN, we parametrize the potential energy network $\check V_{\bm\theta_{\check{V}}}$ and the Euclidean part $g_{\mathbb{R}}$ of the inverse mass-inertia network $\check{\bmM}^{-1}_{\bm\theta_{\check{T}}}$ each with $L_{\check{V}}=L_{\check T, \mathbb R}=2$ hidden Euclidean layers of $32$ neurons and SoftPlus activation functions. We fix the basepoint of the exponential map layer $g_{\text{Exp}}$ to the origin $\bm P=\mathbf I$. Weights are initialized by sampling from a Xavier normal distribution with gain $\sqrt{2}$ and bias vector entries set to $1$.
We train the model on the joint loss~\eqref{eq:loss_RO_HNN} with scaling factor $\lambda = 1$ for the latent loss on $3000$ uniformly sampled random points from the dataset $\mathcal{D}$ with Strang-symplectic integration (see Sec.~\ref{sec3:sub4:symplectic_integration}) over a training horizon of $H_{\train}=12$ timesteps. We use a learning rate of $1.5\times 10^{-2}$ for the \ac{ae} parameters and $7\times 10^{-4}$ for the \ac{hnn} parameters. We train the model with Riemannian Adam~\citep{becigneul2018riemannianoptimization} until convergence at $3000$ epochs.

\textbf{\ac{ae} baselines.}
In Sec.~\ref{sec4:sub1:bd_coupled_pendulum}, we compare the geometrically-constrained symplectic \ac{ae} with linear and quadratic symplectic manifold Galerkin (SMG) projections~\citep{Peng2015:SympMOR,Sharma2023symplMOR}, and a weakly-symplectic \ac{ae}~\citep{Buchfink23:SymplecticMOR}.
We implement the linear and quadratic SMG projections onto a 3-dimensional symplectic submanifold following~\citep{Sharma2023symplMOR}.
We compute the reduction parameters based on a singular value decomposition computed from $3000$ randomly sampled training datapoints in $\mathcal{D}$.

For the weakly-symplectic \ac{ae}~\citep{Buchfink23:SymplecticMOR}, we train two independent constrained AEs for position and momentum reduction and embedding, i.e., $\rhoq^{(l)}: \mathbbr{n_l} \to \mathbbr{n_{l-1}}$, $\varphiq^{(l)}: \mathbbr{n_{l-1}} \to \mathbbr{n_{l}}$, and $\rho_{\mathcal{P}}^{(l)}: \mathbbr{n_l} \to \mathbbr{n_{l-1}}$, $\varphi_{\mathcal{P}}^{(l)}: \mathbbr{n_{l-1}} \to \mathbbr{n_{l}}$, and compute the embedding and reduction for the symplectic manifold as 
\begin{equation}
    \varphi(\checkbmq, \checkbmp) = \left(\begin{matrix}
        \varphiq \\ \varphi_{\mathcal{P}}
    \end{matrix}\right) 
    \quad\quad \text{and} \quad\quad 
    \rho(\bmq, \bmp) = \left(\begin{matrix}
        \rhoq \\ \rho_{\mathcal{P}}
    \end{matrix}\right).
    \label{app:nonsymplecticAE}
\end{equation}
Note that this architecture also fulfills the projection properties~\eqref{eq:projectionproperties} by construction, as the other reduction approaches. However, it does not satisfy the symplecticity property~\eqref{eq:reduced_canonical_symplectic_form}. To enforce this property,~\citet{Buchfink23:SymplecticMOR} introduces a symplecticity loss
\begin{equation}
    \ell_{\text{sympl}} = \frac{1}{N}\sum_{i=1}^N \| \mathbb J_{2d} - d\varphi^\intercal \mathbb J_{2n} d\varphi \|^2_{\text{F}}.
    \label{eq:symplecticity_loss}
\end{equation}
The weakly-symplectic \ac{ae} is trained by minimizing the sum of the reconstruction loss $\ell_{\text{AE}}$ from~\eqref{eq:loss_RO_HNN} and the symplecticity loss~\eqref{eq:symplecticity_loss}.

For the geometrically-constrained symplectic \ac{ae}, we consider the same architecture as in the \ac{rohnn} described above.

All \ac{ae} architectures consist of $l=\{1,2,3\}$ biorthogonal encoder and decoder layers with $n_l = \{6, 12, 15\}$ with latent space dimension $n_0=3$.
We train both \ac{ae} on $3000$ samples from the dataset $\mathcal{D}$ with Riemannian Adam with a learning rate of $1.5\times 10^{-2}$ until convergence at $3000$ epochs.

\textbf{\ac{hnn} baselines.}
In Sec.~\ref{sec4:sub1:bd_coupled_pendulum}, we also ablate the choice of latent \ac{hnn} and integrator. To isolate the \ac{hnn} performance, we consider the low-dimensional dataset $\mathcal{D}_{\text{pend}}$ of the $3$-\ac{dof} pendulum and no reduction.
For the Cholesky \ac{hnn} where the inverse mass-inertia matrix is parametrized via a Cholesky network, we implement shared parameters for the inverse mass-matrix and potential energy networks, i.e., $\bm{\theta}_{\check{\text{T}}} \cap \bm{\theta}_{\check{\text{V}}}$, following~\citep{Lutter2023DeLaN}. The MLP consists of $2$ hidden Euclidean SoftPlus layers of $64$ neurons, while separate output layers return the potential energy and the Cholesky decomposition.
For the black-box \acp{hnn}, we use a single fully-connected \ac{mlp} to model a Hamiltonian function ${\hamiltonian}_{\bm \theta}$. We conduct experiments with two black-box \acp{hnn} of $2$ hidden layers with a width of $64$, and $256$ neurons, respectively.
In all cases, the weights are initialized by sampling from a Xavier normal distribution with gain $\sqrt{2}$, and the bias vector entries are initialized to $1$.

We train all architectures on $3000$ datapoints of the dataset $\mathcal{D}_{\text{pend}}$ with Riemannian Adam optimizer on the \ac{hnn} term $\ell_{\text{HNN},d}$ of the loss~\eqref{eq:loss_RO_HNN} over a training horizon of $H_{\train}=12$ timesteps.
For the ablation of the \ac{hnn} architecture, we use the Strang-symplectic integrator. The geometric \ac{hnn} and Cholesky networks are trained until convergence at $2500$ epochs with learning rate set to $7\times 10^{-4}$.
The black-box \acp{hnn} are trained at a learning rate of $2\times 10^{-3}$ for $3000$ epochs.

For the ablation of the integrator, we use the geometric \ac{hnn} and compare the Strang-symplectic integrator with an explicit Euler integrator, a Runge-Kutta integrator of order $4$, and a symplectic leapfrog integrator. 

\textbf{\ac{hnko} baseline.} In Sec.~\ref{sec4:sub1:bd_coupled_pendulum}, we compare the \ac{rohnn} against the \ac{hnko} proposed by~\citet{wrongkoopman2024}. Moreover, in App.~\ref{appendix3:experiments_pendulum}, we evaluate the performance of the \ac{rohnn} against the \ac{hnko} under noisy observations. Hyperparameters are selected and refined empirically following the supplementary material and code provided by~\cite{wrongkoopman2024}. 

The \ac{hnko} first maps the $2n=30$-dimensional observations into a $d=100$-dimensional lifted latent space via a fully-connected neural network of $6$ hidden layers with Tanh activations. The latent dynamics are then propagated on a $50$-dimensional sphere via the special orthogonal Hamiltonian Koopman operator, implemented by a constrained linear, bias-free layer with $100$-dimensional input and output. The predicted states are mapped back onto the original space with a fully-connected neural network with $3$ hidden layers and Tanh activations. 
The overall model is trained on $3000$ randomly sampled datapoints of the dataset with $\mathcal{D}_{\text{pend}}$, using the Adam optimizer until convergence at $15000$ epochs. For a fair comparison and for stable predictions over longer horizons, we adjusted the loss on the latent Koopman predictions, referred to as $\mathcal L_{\text{koop}}$ in~\citep{wrongkoopman2024}, to sum over a training horizon of $H_{\train}=12$ timesteps, similar to our latent loss term $\ell_{\text{HNN},d}$ in~\eqref{eq:loss_RO_HNN}.

\subsection{Particle Vortex of Section \ref{sec4:sub3:pv_sys}}
\label{appendix2:pv_sys}

\subsubsection{Dataset}
\label{appendix2:pv_sys:sub:data_generation}

\textbf{System.}
In Sec.~\ref{sec4:sub3:pv_sys}, we learn the dynamics of an $n = 90$-dimensional particle vortex, consisting of $j=\left\{1,...,N\right\}$ particles with phase-space coordinates $\bm x_j =\left(q_j, p_j\right)^\intercal$ and uniform interaction strengths $\Gamma_j = 1$. The particle vortex dynamics are governed by the Hamiltonian 
\begin{equation}
\hamiltonian(\bm q, \bm p) = -\sum_{j<k}\log|\bm x_j - \bm x_k|,
    \label{appendix2:pv_sys:hamiltonian}
\end{equation}
that models the interaction between each $j\neq k$ pair of particles~\citep{Xiong2020:NonsepSymplNN}. Note that, as the particle vortex dynamics are purely determined via the logarithmic interaction,  its Hamiltonian function does not separate into kinetic and potential energy, in contrast to mechanical systems such as the pendulum and the cloth.

\textbf{Data generation.} 
We generate a training dataset $\mathcal{D}_{\text{pv}} = \{\{\bmq_{n,k}, \bmp_{n,k}\}_{k=1}^{K}\}_{n=1}^{N}$ by simulating $N=20$ trajectories of the conservative system over the time interval $\mathcal{I}=[0,10.0]\SI{}{\second}$ with timestep $\Delta t = 10^{-3}\SI{}{\second}$ and Strang-symplectic solver with weight parameter $w=0.1$, resulting in $K=10000$ steps per trajectory.
For each trajectory, initial conditions are randomly sampled to mimic clustered vortex distributions. The particles are evenly split among $j=\{1,2,3\}$ clusters. For each cluster, we randomly sample a center $\bm{c}_j$ within a radius of $R = \SI{6}{\meter}$ form the origin. Then, a cluster radius is sampled uniformly from $r_j\in[0.1, 2]\SI{}{\meter}$, and particles within a cluster are positioned following a Gaussian distribution $\mathcal N\sim(\bm{c}_j,r_j^2\mathbf{I})$ around the center $\bm{c}_j$. 
For the testing dataset, we generate $N=10$ trajectories via the same distribution of initial conditions, but simulating the system over a time interval of $\mathcal{I}=[0,15.0]\SI{}{\second}$.

\subsubsection{Model Training}
\label{appendix2:pv_sys:sub:model_training}
The results presented in Sec.~\ref{sec4:sub3:pv_sys} are obtained via \acp{rohnn} composed of a geometrically-constrained symplectic \ac{ae} and a latent geometric \ac{hnn}. We conduct experiments with two \ac{rohnn} with latent space dimensions $d=6$ and $d=10$.
The constrained \ac{ae} is composed of $l=\{1,2,3,4\}$ pairwise biorthogonal encoder and decoder layers of sizes $n_l = \{32, 48, 64, 90\}$.
The biorthogonal weight matrices are initialized by sampling a random orthogonal matrix $\bm O \in \mathbb{R}^{n_l\times n_l}$ from the Haar distribution and setting $\bm\Phi=\bm\Psi=\bm O_{[:,:n_{l-1}]}$, where $\bm O_{[:,:n_{l-1}]}$ are the first $n_{l-1}$ column entries of $\bm O$. Bias vectors are initialized as $\bm{b}_{l}=\bm{0}$.
The latent Hamiltonian network $\check V_{\bm\theta_{\check{V}}}$ is parametrized by $2$ hidden Euclidean layers of $32$ neurons with SoftPlus activation functions. All weights are initialized by sampling from a Xavier normal distribution with gain $\sqrt{2}$, and all bias vector entries are initialized to $1$.

We train the model on the joint loss~\eqref{eq:loss_RO_HNN} on $3000$ random samples from the dataset $\mathcal{D}$ with Strang-symplectic integration over a training horizon $H_{\train}=8$ timesteps. For better convergence, we scale the loss term $\ell_{\text{HNN,d}}$ via a scalar factor $\lambda = 10^{3}$. The parameters are optimized via Riemannian Adam~\citep{becigneul2018riemannianoptimization} until convergence at $3000$ epochs with a learning rate of $1.5\times 10^{-2}$ for the \ac{ae} parameters and $7\times 10^{-4}$ for the \ac{hnn} parameters.

In Sec.~\ref{sec4:sub3:pv_sys}, we consider a comparison with a \ac{rohnn} with a latent black-box \ac{hnn} $\check {\mathcal H}_{\bm \theta}$ composed of $2$ layers of $64$ neurons. We set the learning rate to $2\times 10^{-3}$. The remaining of the \ac{rohnn} architecture and training pipeline are unchanged.

\textbf{Non-symplectic \ac{ae} baseline.}
In Table~\ref{tab:pv_dyns:h_vs_spd_table}, we compare the \ac{rohnn} with variants that use a non-symplectic \ac{ae} instead of the proposed geometrically-constrained symplectic \ac{ae}.
We consider black-box and geometric \acp{hnn} in latent spaces of dimensions $d=\left\{6, 10\right\}$ learned by a non-symplectic projection-constrained \ac{ae}. We use a single vanilla constrained \ac{ae} with latent space dimension $n_0=\{12, 20\}$ with $4$ pairwise biorthogonal encoder-decoder layers of sizes $n_l = \{64, 96, 128, 180\}$. We assume the first $\{6, 10\}$ output dimensions of the latent space to correspond to reduced position $\check{\bm q}$ and the last $\{6, 10\}$ to correspond to the reduced momentum $\check{\bm p}$, which are used as inputs for the latent dissipative \ac{hnn}.
The training procedure remains the same as for the other \acp{rohnn} described above.

\textbf{\ac{hnko} baseline.}
For the results in Table~\ref{tab:pv_dyns:h_vs_spd_table}, the \ac{hnko} first maps the $2n=180$-dimensional particle observations into a $d=800$-dimensional latent space via a fully-connected neural network of $6$ hidden layers with Tanh activations.
The Hamiltonian latent dynamics are modeled on the $400$-dimensional sphere. The resulting state predictions are reconstructed by a fully-connected neural network with $3$ hidden layers with Tanh activations. 
The overall model is trained on $3000$ randomly sampled datapoints, using the Adam optimizer until convergence at $15000$ epochs. We train on a horizon of $H_{\train}=8$ timesteps, as outlined in~\ref{appendix2:pendulum:sub:model_training}.

\subsection{Cloth of Section \ref{sec4:sub2:cloth}}
\label{appendix2:cloth}

\subsubsection{Dataset}
\label{appendix2:cloth:sub:data_generation}
\textbf{System.}
Our last set of experiments is conducted on a deformable thin cloth modeled in \textsc{Mujoco} as a flexible composite object with $i=\{1,...,200\}$ masses $m_i = \SI{0.1}{\kg}$, equally spaced over a width of $\SI{0.1}{\m}$ and length of $\SI{0.2}{\m}$. Generalized coordinates are given by the Cartesian positions $\bm{q}_i = (x_i, y_i, z_i)^\intercal$ of each mass' center of mass in the world frame. The viscous damping coefficient is uniformly set to $d_i=0.01\SI{}{\newton\second\per\meter}$.

\textbf{Data generation.}
Each trajectory captures the cloth falling on a sphere from a height of $\SI{0.12}{\m}$ in the center above the origin of the sphere. To vary scenarios, the radius of the sphere is randomly-sampled from $r \in \left[0.02, 0.12\right]\SI{}{\m}$. The state evolution is simulated with timestep $\Delta t = \SI{e-4}{\s}$ over a time interval $\mathcal{I} = [0, 0.3] \SI{}{\s}$, resulting in $K=3000$ samples per trajectory.
We generate $N=20$ trajectories for a training dataset $\mathcal{D}_{\text{cloth}} = \{\{\bmq_{n,k}, \bmp_{n,k}, \bmtau_{n,k}\}_{k=1}^{K}\}_{n=1}^{N}$, and $N=10$ testing trajectories over a longer time interval $\mathcal{I} = [0, 0.5] \SI{}{\s}$. 
When learning the damping force via a dissipative \ac{hnn}, the generalized force vector consists of external constraint forces, i.e., $\bmtau = \bmtau_{\text{c}}$.
The ablation of Sec.~\ref{sec4:sub2:cloth} compares the dissipative geometric \ac{hnn} against a conservative \ac{hnn} for which all external forces are provided. In this case, the training dataset is composed of generalized force vector $\bmtau = \bmtau_{\text{d}} + \bmtau_{\text{c}}$ that contains both the damping forces $\bmtau_{\text{d}}$ and the constraint forces $\bmtau_{\text{c}}$.

\subsubsection{Model Training}
\label{appendix2:cloth:sub:model_training}

For the RO-HNN experiments in Sec.~\ref{sec4:sub2:cloth}, we train a \ac{rohnn} composed of a geometrically-constrained symplectic \ac{ae} and a latent dissipative geometric \ac{hnn}. The underlying constrained \ac{ae} has $l=\{1,2,3,4\}$ pairwise biorthogonal encoder and decoder layers of sizes $n_l = \{32, 64, 128, 600\}$ with latent space dimension $n_0=6$ or $n_0=10$. 
The biorthogonal weight matrices are initialized by sampling a random orthogonal matrix $\bm O \in \mathbb{R}^{n_l\times n_l}$ from the Haar distribution and setting $\bm\Phi_l=\bm\Psi_l=\bm O_{[:,:n_{l-1}]}$, where $\bm O_{[:,:n_{l-1}]}$ are the first $n_{l-1}$ column entries of $\bm O$. Bias vectors are initialized as $\bm{b}_{l}=\bm{0}$.

The latent potential energy network $\check V_{\bm\theta_{\check{V}}}$ is parametrized with $L_{\check{V}}=L_{\check T, \mathbb R}=2$ hidden Euclidean layers of $32$ neurons. The Euclidean part  $g_{\mathbb{R}}$ of the inverse mass-inertia network $\check{\bmM}^{-1}_{\bm\theta_{\check{T}}}$ and of the damping-matrix network $\check{\bm D}_{\bm\theta_{\check{D}}}$ are composed of $2$ hidden layers with $32$ neurons. For both networks, we fix the basepoint of the exponential map layer $g_{\text{Exp}}$ to the origin $\bm P=\mathbf I$.
All activation functions are SoftPlus, all weights are initialized by sampling from a Xavier normal distribution with gain $\sqrt{2}$, and all bias vector entries are initialized to $1$.

We train the model on the joint loss~\eqref{eq:loss_RO_HNN} on $3000$ samples from the dataset $\mathcal{D}$ with Strang-symplectic integration over a training horizon $H_{\train}=8$ timesteps.  The scaling constant on the latent loss term $\ell_{\text{HNN,d}}$ is set to $\lambda = 10^{4}$. We train the \ac{rohnn} with Riemannian Adam~\citep{becigneul2018riemannianoptimization} until convergence at $3000$ epochs with a learning rate of $1.5\times 10^{-2}$ for the \ac{ae} parameters and $7\times 10^{-4}$ for the \ac{hnn} parameters.

In Fig.~\ref{fig:cloth:fulltraj_abl_lnn_arch_10d_cloth_medians_positions}, we compare the geometric \ac{rohnn} with a black-box variant where the latent \ac{hnn} is encoded as a single black-box network ${\mathcal{H}}_{\bm \theta}$ corresponding to a fully-connected \ac{mlp} of $2$ hidden layers with a width of $256$ neurons.
The \ac{hnn} weights are initialized by sampling from a Xavier normal distribution with gain $\sqrt{2}$, and the bias vector entries are initialized to $1$. This black-box \ac{rohnn} is training with the same parameters as the geometric \ac{rohnn}, except for the learning rate of the \ac{hnn} parameters, which we set as $2\times 10^{-3}$.

\textbf{Sequentially-trained baseline.}
To assess the effectiveness of the proposed joint training procedure, we compare the jointly-trained \ac{rohnn} with a variant that sequentially trains first the geometrically-constrained \ac{ae}, and second the latent \ac{hnn}. As convergence is difficult to achieve when training only the latent \ac{hnn} on a fully-trained representation of the \ac{ae}, we first train only the \ac{ae} by optimizing $\ell_{\text{AE}}$ for $3000$ epochs with a learning rate of $1.5\times 10^{-2}$. Subsequently, we jointly optimize the \ac{ae} and latent loss~\eqref{eq:loss_RO_HNN}. We train the networks jointly within the \ac{rohnn} with Riemannian Adam~\citep{becigneul2018riemannianoptimization} until convergence at learning rates $1.5\times 10^{-2}$ for the \ac{ae} parameters and $7\times 10^{-4}$ for the \ac{hnn} parameters.

\textbf{Projection and \ac{ae} baselines.} In App.~\ref{appendix3:experiments_cloth} (see Fig.~\ref{fig:cloth:over_latdim_k_steps_arch_errors}), we compare the ability of a latent HNN to learn accurate dynamics using different reduction methods to obtain the symplectic embedding $\varphi$ and corresponding reduction $\rho$. We compare the RO-HNN with geometrically-constrained symplectic \ac{ae} with linear and quadratic symplectic manifold Galerkin (SMG) projections~\citep{Peng2015:SympMOR,Sharma2023symplMOR}, and a weakly-symplectic \ac{ae}~\citep{Buchfink23:SymplecticMOR}.
We compute the linear and quadratic SMG projections onto latent spaces of symplectic submanifolds of three different dimensionalities $d=\{2,6,10\}$, following~\citep{Sharma2023symplMOR}, via $3000$ training datapoints. In both cases, we then train a latent \ac{hnn} on the terms $\ell_{\text{HNN}, n}$ and $\ell_{\text{HNN}, d}$ of the joint loss equation~\eqref{eq:loss_RO_HNN} on $3000$ samples from the dataset $\mathcal{D}$ with Strang-symplectic integration over a training horizon $H_{\train}=8$ timesteps. The model is trained with Riemannian Adam~\citep{becigneul2018riemannianoptimization} until convergence at $3000$ epochs with a learning rate of $7\times 10^{-4}$. Note that this essentially corresponds to a scenario with pre-trained symplectic submanifolds, as the parameter optimization for the linear and quadratic embedding maps happens once in the beginning.

The weakly-symplectic AE consists of two independent constrained AEs for position and momentum. We use $4$ layers of size $n_l = \{32, 64, 128, 600\}$ with varying latent space dimension. We train the network jointly on the sum of the losses~\eqref{eq:loss_RO_HNN} and \eqref{eq:symplecticity_loss} via Riemannian Adam~\citep{becigneul2018riemannianoptimization} until convergence at $3000$ epochs with a learning rate of $1.5\times 10^{-2}$ for the \ac{ae} parameters and $7\times 10^{-4}$ for the \ac{hnn} parameters.

\textbf{Non-symplectic \ac{ae} baseline.} In App.~\ref{appendix3:experiments_cloth} (see Fig.~\ref{fig:cloth:symplectic_architecture_ablation}), we ablate geometrically-constrained symplectic \ac{ae} of the \ac{rohnn} in a dissipative scenario. To do so, we train a \ac{rohnn} that utilizes a non-symplectic projection-constrained \ac{ae} instead of geometrically-constrained symplectic \ac{ae}. Specifically, we use a single vanilla constrained \ac{ae} with latent space dimension $n_0=20$ with $4$ pairwise biorthogonal encoder-decoder layers of sizes $n_l = \{64, 128, 256, 1200\}$. Note that we assume the first $10$ output dimensions of the latent space to correspond to reduced position $\check{\bm q}$ and the last $10$ to correspond to the reduced momentum $\check{\bm p}$, which are used as inputs for the latent dissipative \ac{hnn}.

\textbf{Comparison against \ac{rolnn}.}
In Sec.~\ref{sec4:sub2:cloth} and App.~\ref{appendix3:experiments_cloth}, we compare the \acp{rohnn} against \acp{rolnn}~\citep{friedl2025reduced}. 
The \acp{rolnn} are trained on the dataset $\mathcal{D}_{\text{cloth, vel}} = \{\{\bmq_{n,k}, \dbmq_{n,k}, \bmtau_{n,k}\}_{k=1}^{K}\}_{n=1}^{N}$ obtained from the dataset described in App.~\ref{appendix2:cloth:sub:data_generation} by transforming the momentum data into velocities via $\dbmq = \bmM(\bmq)^{-1}\bmp$.

We construct the \ac{rolnn} following the procedure described in~\citep{friedl2025reduced} and use the same architecture for the constrained \ac{ae} as for the \ac{rohnn}. Specifically, we consider a latent space of dimension $n_0=10$ and use $l=\{1,2,3,4\}$ pairwise biorthogonal encoder and decoder layers of sizes $n_l = \{32, 64, 128, 600\}$.
The kinetic and potential energy networks of the latent geometric \ac{lnn} consist of $2$ hidden Euclidean layers with $64$ neurons and SoftPlus activation functions, initialized as for the \ac{rohnn}. Notice that, for the \ac{rolnn}, the dissipation forces $\bmtau_{\text{d}}$ are not learned but provided as ground truth in the external input $\bmtau\! =\! \bmtau_{\text{c}}+ \bmtau_{\text{d}}$.

The \ac{rolnn} is trained on $3000$ samples from $\mathcal{D}_{\text{cloth, vel}}$ with a Runge-Kutta integrator of order $4$ over a training horizon $H_{\train}=8$ timesteps. We train the \ac{rohnn} with Riemannian Adam~\citep{becigneul2018riemannianoptimization} until convergence at $3000$ epochs with a learning rate of $5\times 10^{-2}$ for the \ac{ae} parameters, $2\times 10^{-4}$ for the \ac{hnn} parameters, and a regularization $\gamma=2\times10^{-5}$ for $3000$ epochs.

In our comparison, we also consider a black-box version of the \ac{rolnn}, hereinafter referred to as black-box \ac{rolnn}, where the latent \ac{lnn} is encoded as a single black-box network ${\mathcal{L}}_{\bm \theta}$ representing the Lagrangian function, which we model via a single fully-connected \ac{mlp} of $2$ hidden layers with a width of $256$ neurons.
The network weights are initialized by sampling from a Xavier normal distribution with gain $\sqrt{2}$, and the bias vector entries are initialized to $1$. The black-box \ac{rolnn} is trained with the same parameters as the original \ac{rolnn}, except for the learning rate for the \ac{lnn} parameters, which is set as of $2\times 10^{-3}$.

\textbf{Comparison against \ac{hnko}.} For the prediction errors reported in Tab.~\ref{tab:cloth:long_traj_rel_errors_different_damping_architectures}, we train an \ac{hnko} baseline in a $d=80$-dimensional latent space using a fully connected neural network with $6$ hidden layers and Tanh activations.
The Hamiltonian latent dynamics are modeled on a $40$-dimensional sphere. State reconstruction is performed by a fully-connected neural network with $3$ hidden layers and Tanh activations. 
We train the network on $3000$ training samples from $\mathcal{D}_{\text{cloth}}$, using Adam for $15000$ epochs on a horizon of $H_{\train}=8$ timesteps, adjusting the original loss to multi-step prediction as described in~\ref{appendix2:pendulum:sub:model_training}.

%% file: texfiles/experiments_in_appendix.tex
\section{Additional Experimental Results}
\label{appendix3:experiments_in_appendix}

This section presents additional results, complementing those presented in Sec.~\ref{sec4:results}.

\subsection{Coupled Pendulum of Section \ref{sec4:sub1:bd_coupled_pendulum}}
\label{appendix3:experiments_pendulum}

This section presents additional results on learning the Hamiltonian dynamics of a $15$-\ac{dof} coupled pendulum.

\textbf{Learning high-dimensional dynamics. }
Fig.~\ref{fig:bd_pend:full_traj_plot_pendulum_3dofs_over_time} complements Fig.~\ref{fig:bd_pend:combination_figure_traj_errors} by depicting the predicted long-term ($5$s) positions and momenta. For the ease of visualization, we change the prediction coordinates and plot the first $3$-\ac{dof} corresponding to the latent pendulum. We observe that \ac{rohnn} leads to accurate long-term predictions similar to those of the $3$-\ac{dof} \ac{hnn}. Unlike the $15$-dimensional full-order \ac{hnn}, the \ac{hnko} yields stable, but inaccurate long-term predictions, exhibiting significantly higher deviation from the ground truth trajectory than the predictions of the \ac{rohnn}.

Fig.~\ref{fig:bd_pend:full_traj_latent_space_rohnn_vs_3dpend} depicts the original $15$-\ac{dof} trajectories projected into the $3$-DoF latent space learned by the \ac{rohnn}, along with its latent dynamic predictions over the full prediction horizon of $5$s. We observe that the latent predictions are accurate and match the projected original trajectories. As expected, the learned latent space does not coincide with the phase-space of the original pendulum due to the nonlinear dimensionality reduction conducted via the \ac{ae}, but displays comparable frequencies and amplitudes.

\begin{figure}
    \centering
    \adjustbox{trim=0cm 0.2cm 0cm 0cm}{
            \includesvg[width=.6\linewidth]{figures/bd_pend/fulltraj_3dofs_bd_rep_traj1_rebuttal.svg}  
        }
    \caption{Reconstructed trajectories of the \ac{rohnn} (\solidblueoneline), $3$-\ac{dof} \ac{hnn} (\solidorangetwoline), and \ac{hnko}(\solidgreenthreeline) compared to ground truth (\blackline). The $15$-\ac{dof} \ac{hnn} leads to unstable long-term predictions and is not depicted.}
    \label{fig:bd_pend:full_traj_plot_pendulum_3dofs_over_time}
\end{figure}

\begin{figure}
    \centering
    \adjustbox{trim=0cm 0.2cm 0cm 0cm}{
            \includesvg[width=.6\linewidth]{figures/bd_pend/fulltraj_latent_space_rohnn_vs_3dpend.svg}  
        }
    \caption{Trajectories of the original $15$-\ac{dof} pendulum projected into the latent space of the \ac{rohnn} (\solidgreyline), and corresponding dynamic predictions obtained via the latent \ac{hnn} (\dashedblueoneline). As expected, they does not coincide directly with the trajectories of the underlying $3$-DoF pendulum representation (\blackline).}
\label{fig:bd_pend:full_traj_latent_space_rohnn_vs_3dpend}
\end{figure}

\textbf{\ac{ae} architecture.} Figure~\ref{fig:bd_pend:intrusive_k_pred_comparison_boxplots} accompanies and validates the results of Table~\ref{tab:bd_pend:ae_rec_and_proj_based_pred_errors} by displaying the median and quartiles of the prediction errors obtained by different symplectic dimensionality reduction methods in the intrusive \ac{mor} scenario of Sec.~\ref{sec4:sub1:bd_coupled_pendulum}.

\begin{figure}
    \centering
    \adjustbox{trim=0cm 0.0cm 0cm 0cm}{
            \includesvg[width=.7\linewidth]{figures/bd_pend/intrusive_k_pred_comparison_boxplots.svg}  
        }
    \caption{Prediction errors ($\downarrow$) of intrusive symplectic dimensionality reduction approaches over $10$ test pendulum trajectories.}
\label{fig:bd_pend:intrusive_k_pred_comparison_boxplots}
\end{figure}

\textbf{Latent \ac{hnn} architecture.} 
Here we further evaluate the impact of \ac{hnn} architecture. 
We compare the performance of our geometric \ac{hnn} to learn the low-dimensional dynamics of the latent $3$-\ac{dof} pendulum against \emph{(1)} a non-geometric variant that parametrizes the inverse mass-inertia matrix via a Cholesky network, and \emph{(2)} two \acp{hnn} encoded as a single black-box network ${\hamiltonian}_{\bm{\theta}}$, where we consider two \acp{mlp} of $64$- and $256$-neurons width. 
Compared to Sec.~\ref{sec4:sub1:bd_coupled_pendulum}, we consider a doubled amount of training datapoints with $6000$ random samples.
As shown in Fig.~\ref{fig:3d_pend:full_traj_arch_abl_at_D6000}-\emph{left}, the geometric \ac{hnn} still achieves the lowest reconstruction error, with differences compared to the black-box \ac{hnn} increased compared to the smaller dataset of Fig.~\ref{fig:3d_pend:full_traj_arch__and_solver_ablations_at_D3000}-\emph{left}. This showcases the importance of considering both the quadratic energy structure of mechanical systems, and the geometry of their mass-inertia matrices, for both enhanced performance and data efficiency.

\begin{figure}
    \centering
    \adjustbox{trim=0cm 0cm 0cm 0cm}{
            \includesvg[width=.295\linewidth]{figures/pendulum_3d/full_traj_arch_abl_at_D6000.svg}  
        }
    \adjustbox{trim=0cm 0cm 0cm 0cm}{
            \includesvg[width=.68\linewidth]{figures/pendulum_3d/k_steps_solver_abl_at_D3000.svg}}  
    \caption{\emph{Left}: Ablation of the latent \ac{hnn} architecture on a doubled training set size $| \mathcal D|=6000$ compared to Fig.~\ref{fig:3d_pend:full_traj_arch_abl_at_D6000}-\emph{left}. 
    \emph{Middle, right}:
    Ablation of the latent integrator of the geometric \ac{rohnn} at $|\mathcal D|=3000$ for learning the dynamics of a $15$-\ac{dof} pendulum. Errors are obtained via short-term prediction horizons $H\Delta t=\SI{0.25}{\second}$.}
    \label{fig:3d_pend:full_traj_arch_abl_at_D6000}
\end{figure}

\textbf{Latent integrator.}
We compare the Strang symplectic integrator against \emph{(1)} a symplectic leapfrog integrator that disregards that the Hamiltonian is non-separable, \emph{(2)} a Runge-Kutta integrator of order $4$ that overlooks its symplectic structure, and \emph{(3)} an explicit Euler integrator that also overlooks the symplectic structure. Compared to Sec.~\ref{sec4:sub1:bd_coupled_pendulum} (see Fig.~\ref{fig:3d_pend:full_traj_arch__and_solver_ablations_at_D3000}-middle,right), we consider shorter prediction horizons, feeding the model with ground truth initial conditions every $H\Delta t=\SI{0.25}{\second}$, since the explicit Euler integrator did not lead to stable long-term predictions for $H\Delta t=\SI{5}{\second}$.
Figs.~\ref{fig:3d_pend:full_traj_arch_abl_at_D6000}-\emph{middle}, \emph{right} show that the networks trained via the Strang-symplectic integrator achieve the lowest reconstruction error and conserves energy best during integration, showcasing the importance of considering the symplectic structure of the system during numerical integration for stable predictions on short- and long-term time horizons.

\textbf{Training under noisy observations.} 
To assess the robustness of the \acp{rohnn}, we evaluate its performance under noisy observations and compare it against the \acp{hnko} baseline, which is reported to be robust to noise in high-dimensional systems. 

We generate noisy training data $\left\{\bm q_{i} + \bm{\epsilon}_{\bm{q},i}, \bm p_{i}  + \bm{\epsilon}_{\bm{p},i}\right\}$ corrupted with zero-mean Gaussian noise $\bm{\epsilon}_{\bm{q},i} \sim \mathcal{N}(\bm{0}, \sigma_{\text{q}}^2\bm{I})$ and $\bm{\epsilon}_{\bm{p},i} \sim \mathcal{N}(\bm{0}, \sigma_{\bm{p}}^2\bm{I})$. The noise level $c_{\text{noise}}$ is determines the standard deviations, which is also proportional to the maximum entry of the position and momentum, i.e., $\sigma_{\bm{q}} = c_{\text{noise}}\max_{j,k}| q_{j,k}|$ and $\sigma_{\bm{p}} = c_{\text{noise}}\max_{j,k}| p_{j,k}|$.

Table~\ref{tab:app_bd_pend:noisy_prediciton_results} reports the prediction errors on a testing dataset of $10$ noise-free trajectories over time horizons $H\Delta t=\{0.25,5\}$s for three noise level $c_{\text{noise}}=\{0,0.05,0.1\}$. The noiseless results are repeated from Table~\ref{tab:bd_pend:pred_errors} for completeness.
As expected, the performance of both models decreases with increasing noise magnitude. In each scenario, the \ac{rohnn} outperforms the \ac{hnko} baseline. Note that the \ac{rohnn} trained at $c_{\text{noise}}=0.1$ outperforms the \ac{hnko} trained without noise, demonstrating the enhanced accuracy and robustness of the \ac{rohnn} to noisy observations.

\begin{table}[tbp]
\centering
\caption{Mean and standard deviation of prediction errors ($\downarrow$) over $N=10$ noise-free test pendulum trajectories, comparing the performances of the \acp{rohnn} and \acp{hnko} trained on noisy observations.}
\label{tab:app_bd_pend:noisy_prediciton_results}
\resizebox{\linewidth}{!}{
            \begin{tabular}{cc cc|cc|cc}
 \multicolumn{2}{c}{} & \multicolumn{2}{c}{$c_{\text{noise}} = 0$} & \multicolumn{2}{c}{$c_{\text{noise}} =  5\%$} & \multicolumn{2}{c}{$c_{\text{noise}} = 10\%$} \\
 & $H\Delta t$ (s) & \textcolor{blue01}{\ac{rohnn}} & \textcolor{green03}{\ac{hnko}} 
  & \textcolor{blue01}{\ac{rohnn}} & \textcolor{green03}{\ac{hnko}}
   & \textcolor{blue01}{\ac{rohnn}} & \textcolor{green03}{\ac{hnko}}
\\
\midrule

\multirow{2}{*}{$\frac{\|\tilde{\bmq}_{\text{p}} - \bmq\| }{ \|\bmq\|}$} & $0.25$ &  $(1.66\pm 1.38)\times 10^{-1}$ & $(5.64\pm 4.41)\times 10^{-1}$ 
& $(2.44\pm 1.94)\times 10^{-1}$ & $(9.76\pm 10.48)\times 10^{-1}$ 
& $(3.23\pm 2.86)\times 10^{-1}$ & $(1.02\pm 0.78)\times 10^{0}$
\\ 

& $5$ & $(7.08\pm 7.56)\times 10^{-1}$  & $(1.32\pm 0.94)\times 10^{0}$
& $(8.40\pm 8.17)\times 10^{-1}$ & $(1.98\pm 1.84)\times 10^{0}$ 
& $(9.14\pm 10.02)\times 10^{-1}$ & $(4.06\pm 3.47)\times 10^{0}$
\\  

\midrule

\multirow{2}{*}{$\frac{\|\tilde{\bmp}_{\text{p}} - \bmp\| }{ \|\bmp\|}$}  & $0.25$ & $(5.33\pm 5.23)\times 10^{-2}$  & $(5.93\pm 10.73)\times 10^{-1}$ &
$(7.02\pm 7.17)\times 10^{-2}$ & $(4.56\pm 8.34)\times 10^{-1}$ 
& $(9.64\pm 10.30)\times 10^{-2}$ & $(9.60\pm 17.23)\times 10^{-1}$
\\ 

& $5$ & $(1.98\pm 2.67)\times 10^{-1}$ & $(1.23\pm 2.08)\times 10^{0}$  
& $(3.34\pm 4.97)\times 10^{-1}$ & $(1.62\pm 2.97)\times 10^{0}$ 
& $(3.61\pm 4.94)\times 10^{-1}$ & $(3.13\pm 5.58)\times 10^{0}$
\\  
\end{tabular}
}
\end{table}

\textbf{Number of training samples.}
Table~\ref{tab:app_bd_pend:training_points_scaling} assesses the influence of the number of the number of available training samples on the \ac{rohnn} convergence. It reports the mean and standard deviation of the loss~\eqref{eq:loss_RO_HNN} across $N=10$ testing trajectories and $10$ individual \acp{rohnn} trained from different random seeds.
We observe decreasing prediction errors with increasing number of training samples, before a stabilization around a high number of training data.

\begin{table}[tbp]
\centering
\caption{Mean and standard deviation of testing loss ($\downarrow$) over different training set size $\mathcal D$. Errors are computed over $N=10$ test trajectories and $10$ different \acp{rohnn} that started training from different random seeds.}
\label{tab:app_bd_pend:training_points_scaling}
\resizebox{0.7\linewidth}{!}{
            \begin{tabular}{cccc}
$|\mathcal D|$ & $1000$ & $3000$ & $6000$
\\
\midrule
$\ell_{\text{RO-HNN}}$ &$(9.84\pm 4.64)\times 10^{-3}$ & $(1.73\pm 0.33)\times 10^{-3}$ & $(1.67\pm 0.21)\times 10^{-3}$ \\
\end{tabular}
}
\end{table}

\subsection{Particle Vortex ($90$-DoF) of Section~\ref{sec4:sub3:pv_sys}}
\label{appendix3:experiments_pv}
This section presents additional results on learning the Hamiltonian dynamics of a $90$-\ac{dof} particle vortex.

Fig.~\ref{fig:pv:particlevortex_dims_over_time} depicts the predicted positions and momenta of the particles along with the ground truth in the high-dimensional state space for \acp{rohnn} with latent dimension $d=\{6,10\}$.
Fig.~\ref{fig:pv:latent_particlevortex_dims_over_time} depicts the predicted positions and momenta of the particles in the reduced phase space of the \ac{ae} along with the projected ground truth. We observe that both models accurately predict the particle vortex dynamics, with the $d=6$-dimensional model slightly outperforming the $10$-dimensional one (see also Table~\ref{tab:pv_dyns:h_vs_spd_table} and Fig.~\ref{fig:pv:k_step_pos_mom_pred_box_err_over_latdim_and_architectures}).
This shows that the choice of latent dimension is a trade off between the latent space expressivity and the limitations of \acp{hnn} in higher dimensions.
In general, we observed that errors initially decrease as the latent dimension increases, suggesting that higher-dimensional latent spaces better capture the original high-dimensional dynamics. The errors then increase beyond a certain latent dimension, indicating that the latent \ac{hnn} becomes harder to train. 

Fig.~\ref{fig:pv:k_step_pos_mom_pred_box_err_over_latdim_and_architectures} accompanies and validates the results of Table~\ref{tab:pv_dyns:h_vs_spd_table} by displaying the median and quartiles of the prediction errors obtained by different latent \acp{hnn} in combination with the symplectic \ac{ae}.

\begin{figure}[tbp]
    \centering
    \adjustbox{trim=0cm 0cm 0cm 0cm}{
            \includesvg[width=\linewidth]{figures/pv_dyns/6and10pv_dyns_over_time_and_dims.svg}  
        }
    \caption{Predicted (\orangecircle, \darkorangecircle, \marooncircle) vs ground truth (\greycircle, \darkgreycircle, \blackcircle) positions of the particle vortex. The dynamics are learned with \ac{rohnn} with $d=6$ and $d=10$. Times beyond $10$s are out of the training data distribution.}
    \label{fig:pv:particlevortex_dims_over_time}
\end{figure}
\begin{figure}[tbp]
    \centering
    \adjustbox{trim=0cm 0cm 0cm 0cm}{
            \includesvg[width=\linewidth]{figures/pv_dyns/6and10latent_pv_dyns_over_time_and_dims.svg}  
        }
    \caption{Predicted (\orangecircle, \darkorangecircle, \marooncircle) vs ground truth (\greycircle, \darkgreycircle, \blackcircle) reduced positions of the particle vortex in the latent space of the $\ac{rohnn}$ with $d=6$ and $d=10$. Times beyond $10$s are out of the training data distribution.}
    \label{fig:pv:latent_particlevortex_dims_over_time}
\end{figure}

\begin{figure}
\vspace{-0.3cm}
    \centering
    \adjustbox{trim=0cm 0.0cm 0cm 0.0cm}{
            \includesvg[width=.6\linewidth]{figures/pv_dyns/k_step_pos_mom_pred_box_err_over_latdim_and_architectures.svg}  
        }
    \caption{\ac{rohnn} prediction errors ($\downarrow$) for black-box and geometric \acp{hnn} in latent spaces learned by a symplectic \ac{ae} with latent dimensions $d=6$ (\powdermagentabox) and $d=10$ (\olivegreenbox) over $10$ particle vortex trajectories.
    }
    \label{fig:pv:k_step_pos_mom_pred_box_err_over_latdim_and_architectures}
    \vspace{-0.3cm}
\end{figure}

\subsection{Cloth ($600$-\ac{dof}) of Section~\ref{sec4:sub2:cloth}}
\label{appendix3:experiments_cloth}
This section presents additional results on learning the Hamiltonian dynamics of a $600$-\ac{dof} thin cloth falling on a sphere.

\textbf{Learning high-dimensional dynamics with dissipation.}
As shown in Table~\ref{tab:cloth:long_traj_rel_errors_different_damping_architectures}, the $10-$dimensional model slightly outperforms the $6$-dimensional one, modeling more details of the cloth, as observed in Fig.~\ref{fig:cloth:mj_vis_damped_traj_pred_10D_6D_rolnn}.

Fig.~\ref{fig:cloth:fulltraj_10d_arch_abl_median} accompanies Table~\ref{tab:cloth:long_traj_rel_errors_different_damping_architectures} by visualizing the median and quartiles of the \ac{rohnn} ($d=10$) reconstructed and latent prediction errors over time for two different parametrization of the reduced dissipation matrix $\check{\bm{D}}$. Both dissipative \ac{rohnn} perform similarly to the conservative \ac{rohnn}, showing that the \ac{rohnn} can successfully predict dissipative dynamics in a stable manner, including beyond the training time horizon.

\begin{figure}
    \centering
    \adjustbox{trim=0cm 0cm 0cm 0cm}{
            \includesvg[width=.99\linewidth]{figures/cloth/fulltraj_abl_arch_10d_cloth_medians.svg}  
        }
    \caption{Median and quartiles of the latent and reconstructed prediction errors of $10$-dimensional \acp{rohnn} with latent dissipation matrix parametrized with a \ac{spd} network (\solidlightpinkline), a Cholesky network (\soliddeepochreline), and ground truth values (\solidlighterfoamblueline). The gray-shaded area indicates the time horizon beyond the training data support.}
    \label{fig:cloth:fulltraj_10d_arch_abl_median}
\end{figure}

Fig.~\ref{fig:cloth:prediction_of_dofs} shows the predictions of the \acp{rohnn} with different parametrizations of the dissipation matrix $\check{\bm{D}}$ for selected dimensions of a test trajectory. This shows that the dissipative \acp{rohnn} successfully learn the dissipation forces, achieving similar prediction errors as the conservative models
Fig.~\ref{fig:cloth:energy_pred_and_err} displays the predicted latent energy to be compared with the ground-truth energy projected in the symplectic latent space. Overall, our results demonstrate the ability of the \ac{rohnn} to infer long-term predictions of dissipative systems.

\begin{figure}
    \centering
    \adjustbox{trim=0cm 0cm 0cm 0cm}{
            \includesvg[width=.8\linewidth]{figures/cloth/long_traj_dofs.svg}  
        }
    \caption{Predicted cloth positions and momenta for $6$-dimensional \acp{rohnn} with latent dissipation matrix parametrized with a \ac{spd} network (\soliddarkpinkline), a Cholesky network (\soliddeepochreline), and ground truth values (\dasheddarkblueline), and $10$-dimensional \acp{rohnn} with latent dissipation matrix parametrized with a \ac{spd} network (\solidlightpinkline), a Cholesky network (\solidlighterdeepyellowline), and ground truth values (\dashedlighterfoamline). The grey-shaded areas indicates interval beyond the data support.}
    \label{fig:cloth:prediction_of_dofs}
\end{figure}

\begin{figure}
    \centering
    \adjustbox{trim=0cm 0cm 0cm 0cm}{
            \includesvg[width=.4\linewidth]{figures/cloth/long_traj_energy_pred_and_err.svg}  
        }
    \caption{\emph{Top}: Ground truth (\blackline) and predicted latent energies for $6$-dimensional \acp{rohnn} with latent dissipation matrix parametrized with a \ac{spd} network (\soliddarkpinkline), a Cholesky network (\soliddeepochreline), and ground truth values (\dasheddarkblueline), and $10$-dimensional \acp{rohnn} with latent dissipation matrix parametrized with a \ac{spd} network (\solidlightpinkline), a Cholesky network (\solidlighterdeepyellowline), and ground truth values (\dashedlighterfoamline). \emph{Bottom}: Energy errors for the same models. The grey-shaded areas indicate intervals beyond the data support, for which the ground truth is extrapolated from the last observation.}
    \label{fig:cloth:energy_pred_and_err}
\end{figure}

\textbf{Comparison against \ac{rolnn}.}
Fig.~\ref{fig:cloth:fulltraj_abl_lnn_arch_10d_cloth_medians_positions} compares the latent prediction and reconstructed prediction errors of the geometric and black-box \ac{rohnn} and \ac{rolnn} over time. 

Our results show that the geometric \ac{rohnn} outperforms the \ac{rolnn}, leading to more accurate predictions even as the \ac{rohnn} learns the dissipation forces via the latent damping matrix, whereas the \ac{rolnn} receives their ground truth measurements. 
Moreover, Fig.~\ref{fig:cloth:fulltraj_abl_lnn_arch_10d_cloth_medians_positions} shows that the  geometric \ac{rohnn} and \ac{rolnn} featuring geometric latent \ac{hnn} and 
\ac{lnn} outperform their black-box counterparts, showcasing the importance of considering the quadratic energy structure of mechanical systems in both network types. 

We hypothesize that the improved accuracy of the \ac{rohnn} compared to the \ac{rolnn} can be attributed to \emph{(1)} the first-order dynamic formulation stemming from Hamiltonian mechanics, which is easier to learn and optimize than the second-order Lagrangian formulation, \emph{(2)} the Strang-symplectic integrator which is specifically designed for Hamiltonian systems, in contrast to the Runge Kutta integrators typically used in the case of continuous-time Lagrangians. This aligns with the discussions in~\citep{Liu24:PINN}, which showed that, by using position and momentum observations, \acp{hnn} learn mass-inertia matrices that are close to the physical solutions, while \acp{lnn} only learn one of the solutions satisfying the Euler-Lagrange equations.

\begin{figure}
    \centering
    \adjustbox{trim=0cm 0cm 0cm 0cm}{
            \includesvg[width=.5\linewidth]{figures/cloth/fulltraj_vs_lnn_10d_cloth_medians_recpos_latpos.svg}  
        }
    \caption{Median and quartiles of the latent and reconstructed prediction errors of $10$-dimensional geometric \ac{rohnn} (\solidlighterfoamblueline), geometric \ac{rolnn} (\solidryellowline), a \ac{rohnn} with latent black-box \ac{hnn} (\dashedlighterfoamline), and a \ac{rolnn} with latent black-box \ac{lnn} (\dashedryellowline). }
    \label{fig:cloth:fulltraj_abl_lnn_arch_10d_cloth_medians_positions}
\end{figure}

\textbf{Latent dimension and training ablation.}
We compare the performance of our dissipative \ac{rohnn} across several latent dimensions $d=\{2,6,10\}$ with jointly-trained geometrically-constrained symplectic \ac{ae} and latent geometric \ac{hnn} against sequentially-trained architectures. Specifically, we consider \emph{(1)} linear and \emph{(2)} quadratic symplectic manifold Galerkin (SMG) projections~\citep{Peng2015:SympMOR,Sharma2023symplMOR}, \emph{(3)} a weakly-symplectic \ac{ae} trained jointly with a latent geometric \ac{hnn}, and \emph{(4)} a \ac{rohnn} with pretrained geometrically-constrained \ac{ae}.
Fig.~\ref{fig:cloth:over_latdim_k_steps_arch_errors} shows that our jointly-trained \ac{rohnn} significantly outperforms all baselines for all dimensions, leading to reduced relative reconstruction, latent prediction, and reconstructed prediction errors. This showcases \emph{(1)} the higher expressivity of the \acp{ae} compared to linear and quadratic projection methods, \emph{(2)} the importance of structurally-embedding the symplecticity condition, unlike the weakly-symplectic \ac{ae}, and \emph{(3)} the importance of joint training, allowing the \ac{rohnn} to jointly learn a symplectic submanifold and the associated dynamics.

\begin{figure}
    \centering
    \adjustbox{trim=0cm 0cm 0cm 0cm}{
            \includesvg[width=.85\linewidth]{figures/cloth/errors_over_latdim/ae_arch_over_err_and_dim.svg}  
        }
    \caption{Mean and standard deviation of the relative reconstruction (\emph{left}), latent prediction (\emph{middle}), and reconstructed prediction (\emph{right}) errors over $10$ cloth trajectories with $H\Delta t = \SI{0.0025}{\second}$. Our \ac{rohnn} with geometrically-constrained symplectic \ac{ae} (\mutedbluesolidcrossline) is compared against linear SMG reduction (\mutedlimesolidcrossline), quadratic SMG reduction (\mutedgreygreensolidcrossline), a weakly symplectic \ac{ae} (\mutedpurplesolidcrossline), and a sequentially-trained \ac{rohnn} with pretrained geometrically-constrained symplectic \ac{ae} (\rustysolidcrossline). The pretrained \ac{ae} (\rustydashedcrossline) is depicted for completeness. Notice that the linear SMG and quadratic SMG projections led to diverging dynamics for $d>2$ and $d>6$, respectively, for which results are not depicted.} \label{fig:cloth:over_latdim_k_steps_arch_errors}
    \vspace{-0.5cm}
\end{figure}

Finally, we compare the performance of the dissipative \ac{rohnn} against \emph{(1)} a conservative \ac{rohnn}, where the dissipation forces $\bmtau_{\text{d}}$ are not learned but provided as ground truth in the external input $\bmtau\! =\! \bmtau_{\text{c}}+ \bmtau_{\text{d}}$, and \emph{(2)} a dissipative \ac{rohnn} where the dissipation matrix is parametrized via Cholesky decomposition for latent dimensions $d=\{2,6,10\}$. The mass-inertia matrix is parametrized via \ac{spd} networks in all cases. Fig.~\ref{fig:cloth:over_latdim_full_traj_damping_archs_errors} shows the obtained latent prediction and reconstructed prediction errors. Both dissipative \acp{hnn} achieve errors close to the conservative \ac{hnn} where the ground truth dissipative forces are provided, with the geometric \ac{hnn} slightly outperforming its Cholesky counterpart. However, the effect is less pronounced as when learning the inverse mass-inertia matrix, which we attribute to the reduced influence of damping compared to inertia in the overall dynamics.

\begin{figure}
    \centering
    \adjustbox{trim=0cm 0cm 0cm 0cm}{
            \includesvg[width=.6\linewidth]{figures/cloth/errors_over_latdim/damping_arch_over_err_and_dim.svg}  
        }
    \caption{Mean and standard deviation of the latent prediction (\emph{left}) and reconstructed prediction (\emph{right}) errors for different parametrization of the latent dissipation matrix $\check{\bm{D}}$ over $10$ test cloth trajectories. We compare our \ac{spd} network (\solidpastelpurpleline) against a Cholesky network (\soliduglyyellowline), and the ground truth parametrization (\solidblueoneline). } 
    \label{fig:cloth:over_latdim_full_traj_damping_archs_errors}
\end{figure}

\textbf{Ablation of the symplectic architecture to learn dissipative dynamics.}
As discussed in Sec.~\ref{sec3:sub1:hamiltonian_rom}, the dissipative dynamics do not preserve a symplectic structure. Our proposed \ac{rohnn} features two main symplecticity-preserving components, namely the geometrically-constrained symplectic \ac{ae} and the Strang-symplectic integrator, which we ablate here. We consider two variations of each components, i.e., (1) our geometrically-constrained symplectic \ac{ae} and a vanilla projection-constrained \ac{ae}, and (2) the Strang-symplectic integrator and a Runge-Kutta integrator of order $4$. Throughout all experiments, we set the latent dimension $d=10$. Fig.~\ref{fig:cloth:symplectic_architecture_ablation} shows the obtained reconstructed prediction errors. We observe that our geometrically-constrained symplectic \ac{ae} leads to significantly lower median errors than vanilla projection-constrained \ac{ae} independently of the choice of integrator, showcasing the benefit of preserving the structure of \ac{fom} vector field in the \ac{rom} (see Proposition~\ref{prop:reduced_dissipative_vector_field}). Moreover, despite the dissipative structure, the \ac{rohnn} obtained with the Strang-symplectic integrator outperforms the non-symplectic Runge-Kutta integrator. We hypothesize that this is due to the fact that the evolution of this dissipative system is mostly governed by its Hamiltonian function, especially over the short timesteps taken by the integrators.

\begin{figure}
    \centering
    \adjustbox{trim=0cm 0cm 0cm 0cm}{
            \includesvg[width=.6\linewidth]{figures/cloth/fulltraj_symplectic_ablation_cloth.svg}  
        }
    \caption{Prediction errors ($\downarrow$) of \acp{rohnn} with geometrically-constrained symplectic or vanilla constrained \ac{ae} with Strang symplectic integrator (\bluebox) or Runge-Kutta integrator of order $4$ (\lightbluebox) over $10$ testing cloth trajectories.}
    \label{fig:cloth:symplectic_architecture_ablation}
\end{figure}

\section{Runtimes}
\label{appendix3:runtimes}
This section compares the runtimes of different models. All experiments were performed locally on a MacBook Pro with M3 CPU.

\textbf{Speedup of simulation via the \ac{rohnn}.} 
We aim at providing an idea of the computational effort of the \ac{rohnn} compared to the evaluation of the \acp{fom}. To do so, we symbolically derive the Hamiltonian equations of motion with known physical quantities of the $15$-\ac{dof} coupled pendulum of Sec.~\ref{sec4:sub1:bd_coupled_pendulum} and obtain the equations of motion of the $600$-\ac{dof} cloth from Sec.~\ref{sec4:sub2:cloth} from Mujoco. We compare the wall-clock time of the evaluation of these two \acp{fom} against the respective \ac{rohnn} averaged over $10$ 
single trajectory roll-outs from the same initial conditions as in the testing dataset. We consider two different integrators, namely an Euler forward and the Strang symplectic integrator, and different step size $\Delta t$ on a time-horizon of $H\Delta t=5$s for the pendulum and $H\Delta t=0.5$s for the cloth.
The evaluation times for the coupled pendulum and the cloth are given in Table~\ref{tab:speedup}, with the corresponding relative position prediction errors depicted in Fig.~\ref{fig:speedup}.
We observe a prominent reduction of the evaluation time for the \acp{rohnn} compared to the \acp{fom}. This reduction is exacerbated for higher-dimensional systems, e.g., the cloth, where the evaluation time of the \ac{rohnn} remains significantly lower than that of the \ac{fom} evaluated with the computationally-cheaper forward Euler integrator or increased step size. Moreover, as shown in Fig.~\ref{fig:speedup}, the \acp{rohnn} exhibit lower prediction errors than their \ac{fom} counterparts in addition to a reduced computational complexity.
This showcases that the \ac{rohnn} not only enable accurate learning of unknown high-dimensional Hamiltonian dynamics, but also the computationally-efficient and accurate evaluation of known systems via surrogate dynamics.

\begin{table}[tbp]
\centering
\caption{Evaluation wall clock times for different ODE-solvers on analytic \ac{fom} compared to \ac{rohnn}. Runtimes are averaged over $10$ forward passes and given in $\SI{}{\second}$.}
\label{tab:speedup}
\resizebox{0.6\linewidth}{!}{
\begin{tabular}{c|ccc|ccc}
& \multicolumn{3}{c|}{\textbf{pendulum} ($n=15, d=3$)}  & \multicolumn{3}{c}{\textbf{cloth} ($n=600, d=10$)} \\
& \multicolumn{2}{c}{$\Delta t = 10^{-2}$s} & $\Delta t = 10^{-1}$s & \multicolumn{2}{c}{$\Delta t = 10^{-4}$s} & $\Delta t = 10^{-3}$s \\
& Strang & Euler & Strang & Strang & Euler & Strang \\
\midrule

\ac{fom}, $n-$\ac{dof} & $1.29$& \textcolor{rbrown}{$0.75$} & \textcolor{rpurple}{$0.18$} &  $255.24$ & \textcolor{rbrown}{$77.62$} & \textcolor{rpurple}{$109.74$}\\

\ac{rohnn}, $d-$\ac{dof} & \textcolor{blue01}{$0.79$} & --- & --- & \textcolor{blue01}{$16.01$} & --- & --- \\
\end{tabular}
}
\end{table}

\begin{figure}
    \centering
    \adjustbox{trim=0cm 0cm 0cm 0cm}{
            \includesvg[width=.5\linewidth]{figures/speedup_ablation_pend_and_cloth.svg}  
        }
    \caption{Position prediction errors introduced by various simulation speedup methods: Euler-forward integration (\solidrbrownline), Strang integration (\solidrpurpleline) at larger stepsize, and \ac{rohnn} (\solidblueoneline).}
    \label{fig:speedup}
    \vspace{-0.3cm}
\end{figure}

\textbf{Comparison of runtimes of \acp{hnn} and \acp{rohnn}. }
\label{appendix3:runtimes_rohnn}
Table~\ref{tab:app_bd_pend:runtimes_forward_pass} reports the averaged runtimes for the forward pass of the differently-sized network architectures considered in Sec.~\ref{sec4:sub1:bd_coupled_pendulum}. 
The reported times correspond to the wall clock time of one forward pass of a batch of $10$ initial conditions, predicted over ${H=10}$ timesteps with the Strang-symplectic integrator. We observe that the \ac{rohnn} speeds up the forward dynamics computation compared to the \ac{hnn}, highlighting the computational advantages of \acp{rom} compared to \acp{fom}. Moreover, the black-box \ac{hnn} is computationally more efficient than the geometric \ac{hnn} at the expense of prediction accuracy.

\begin{table}[tbp]
\centering
\caption{Evaluation wall clock times for different network architectures on the $15$-\ac{dof} pendulum. Runtimes are averaged over $10$ forward passes and given in $\SI{}{\milli\second}$.}
\label{tab:app_bd_pend:runtimes_forward_pass}
\resizebox{0.5\linewidth}{!}{
\begin{tabular}{cc|ccc}
\multicolumn{2}{c|}{$15$-\ac{dof}} & \multicolumn{2}{c}{$3$-\ac{dof}} \\
Geometric \ac{hnn} & Geometric \ac{rohnn}  & Geometric \ac{hnn} & Black-box \ac{hnn} \\
  \midrule
  $100.25$ & $26.34$ & $18.10$ & $8.04$ \\
\end{tabular}
}
\bigskip
\caption{Evaluation wall clock times of different variants for the \ac{ae} on the $600$-DoF cloth dataset. Runtimes are averaged over $10$ forward passes and given in $\SI{}{\milli\second}$.}
\label{tab:app_cloth:runtimes_ae_forward_pass}
\resizebox{.7\linewidth}{!}{
\begin{tabular}{c|ccc}
Position-level \ac{ae} & Geometric \ac{ae} with analytic lift & Geometric \ac{ae} with \texttt{autodiff.vjp} lift & \ac{ae} with naive lift\\
$\varphiq\circ\rhoq(\bm q)$ & $\varphi\circ\rho(\bm q, \bm p)$  & $\varphi\circ\rho(\bm q, \bm p)$ & $\varphiq\circ\rhoq(\bm q, \bm p)$\\
  \midrule
  $16.92$ & $20.93$ & $54.24$ & $19.03$
\end{tabular}
}
\end{table}

\textbf{Runtimes of the lifted \ac{ae}.}
\label{appendix3:runtimes_ae}
Table~\ref{tab:app_cloth:runtimes_ae_forward_pass} reports the average wall-clock times for one forward pass of a batch of $100$ states for several projection scenarios on the $600$-\ac{dof} cloth dataset. We consider reduction to a latent space of dimension $d\!=\!10$ via $4$ layers of size $n_l = \{32, 64, 128, 600\}$. We consider (1) a position-level constrained \ac{ae}, where only position projections $\tilde{\bmq}=\varphiq \circ \rhoq(\bmq)$ are computed via the encoder and decoder layers, (2) a geometric symplectic \ac{ae}, whose lifted mappings~\eqref{eq:cotangent_lifted_maps} are computed analytically as a composition of layer derivatives (see App.~\ref{appendix1:constrained_ae}) to project both positions and momenta via $(\tilde{\bmq}, \tilde{\bmp}) = \varphi \circ \rho(\bmq, \bmp)$, (3) a geometric symplectic \ac{ae} whose lifted maps are computed via automatic differentiation using Pytorch's \texttt{autograd.vjp} function, and (4) a naive constrained \ac{ae} that jointly projects the positions and momenta in $2d\!=\!20$-dimensional latent space via doubled layer dimensions, i.e., $n_l = \{64, 128, 256, 1200\}$.  
The first two variants were evaluated under \texttt{torch.no\_grad()}, reflecting a realistic scenario for evaluation of forward dynamics in the \ac{rohnn}. 

The runtimes reported in Table~\ref{tab:app_cloth:runtimes_ae_forward_pass} show that the analytic computation of the lifted mappings is significantly faster than the automatic-differentiation-based implementation. This is expected, as our analytic implementation avoids the construction of a backward graph. 
It is worth emphasizing that the geometric \ac{ae} with analytic lifts requires significantly less than twice the runtime of the position-level constrained \ac{ae}.
Therefore, using a single cotangent-lifted \ac{ae} that jointly projects positions and momenta is computationally more advantageous than training two separate \acp{ae} for separate projections.